\pdfoutput=1
\documentclass[10pt,logo,copyright]{nvidiatechreport}
\usepackage[authoryear,sort&compress,round]{natbib}

\usepackage[utf8]{inputenc} 
\usepackage[T1]{fontenc}    

\usepackage{parskip}        
\usepackage{xurl}           
\usepackage{hyperref}       
\usepackage{booktabs}       
\usepackage{amsfonts}       
\usepackage{nicefrac}       
\usepackage{microtype}      
\usepackage{xcolor}         
\usepackage[dvipsnames]{xcolor} 
\usepackage{fontawesome5}    
\renewcommand{\sfdefault}{\reportfontfamily} 
\usepackage{accsupp}         
\definecolor{diagramInk}{HTML}{243449}
\definecolor{diagramMuted}{HTML}{65758A}
\definecolor{diagramPlan}{HTML}{167F91}
\definecolor{diagramRaster}{HTML}{76B900}
\definecolor{diagramCode}{HTML}{3A6CCB}
\definecolor{diagramNative}{HTML}{8055B7}
\definecolor{diagramReview}{HTML}{D79532}
\DeclareRobustCommand{\diagramicon}[1]{\makebox[1.1em][c]{\BeginAccSupp{method=escape,ActualText={}}\faIcon{#1}\EndAccSupp{}}\hspace{0.3em}}

\usepackage{graphicx}
\usepackage{animate}        
\usepackage{subcaption}
\usepackage{tabularx}
\usepackage{makecell}
\usepackage{adjustbox}
\usepackage{setspace}
\newcolumntype{M}[1]{>{\centering\arraybackslash}m{#1}}
\usepackage{float}
\usepackage{placeins}
\usepackage[most]{tcolorbox}
\usepackage{enumitem}
\usepackage{tikz}
\usetikzlibrary{positioning,shapes,arrows}
\usepackage{amsmath,amsfonts,bm, bbm,leftindex}
\usepackage{multirow}
\usepackage{comment}
\usepackage{gensymb}
\usepackage{lipsum}
\usetikzlibrary{arrows.meta, positioning, fit}
\usepackage[para]{threeparttable}
\usepackage{tikz}
\usetikzlibrary{tikzmark}

\def\eqref#1{equation~\ref{#1}}

\def\1{\bm{1}}

\def\x{\mathbf{x}}

\DeclareMathAlphabet{\mathsfit}{\encodingdefault}{\sfdefault}{m}{sl}
\SetMathAlphabet{\mathsfit}{bold}{\encodingdefault}{\sfdefault}{bx}{n}

\makeatletter
\let\save@mathaccent\mathaccent
\newcommand*\if@single[3]{%
  \setbox0\hbox{${\mathaccent"0362{#1}}^H$}%
  \setbox2\hbox{${\mathaccent"0362{\kern0pt#1}}^H$}%
  \ifdim\ht0=\ht2 #3\else #2\fi
  }
\newcommand*\rel@kern[1]{\kern#1\dimexpr\macc@kerna}
\newcommand*\widebar[1]{\@ifnextchar^{{\wide@bar{#1}{0}}}{\wide@bar{#1}{1}}}
\newcommand*\wide@bar[2]{\if@single{#1}{\wide@bar@{#1}{#2}{1}}{\wide@bar@{#1}{#2}{2}}}
\newcommand*\wide@bar@[3]{%
  \begingroup
  \def\mathaccent##1##2{%
    \let\mathaccent\save@mathaccent
    \if#32 \let\macc@nucleus\first@char \fi
    \setbox\z@\hbox{$\macc@style{\macc@nucleus}_{}$}%
    \setbox\tw@\hbox{$\macc@style{\macc@nucleus}{}_{}$}%
    \dimen@\wd\tw@
    \advance\dimen@-\wd\z@
    \divide\dimen@ 3
    \@tempdima\wd\tw@
    \advance\@tempdima-\scriptspace
    \divide\@tempdima 10
    \advance\dimen@-\@tempdima
    \ifdim\dimen@>\z@ \dimen@0pt\fi
    \rel@kern{0.6}\kern-\dimen@
    \if#31
      \overline{\rel@kern{-0.6}\kern\dimen@\macc@nucleus\rel@kern{0.4}\kern\dimen@}%
      \advance\dimen@0.4\dimexpr\macc@kerna
      \let\final@kern#2%
      \ifdim\dimen@<\z@ \let\final@kern1\fi
      \if\final@kern1 \kern-\dimen@\fi
    \else
      \overline{\rel@kern{-0.6}\kern\dimen@#1}%
    \fi
  }%
  \macc@depth\@ne
  \let\math@bgroup\@empty \let\math@egroup\macc@set@skewchar
  \mathsurround\z@ \frozen@everymath{\mathgroup\macc@group\relax}%
  \macc@set@skewchar\relax
  \let\mathaccentV\macc@nested@a
  \if#31
    \macc@nested@a\relax111{#1}%
  \else
    \def\gobble@till@marker##1\endmarker{}%
    \futurelet\first@char\gobble@till@marker#1\endmarker
    \ifcat\noexpand\first@char A\else
      \def\first@char{}%
    \fi
    \macc@nested@a\relax111{\first@char}%
  \fi
  \endgroup
}
\makeatother

\newcommand{\frameworkname}{Nemotron-Labs-VisualCreation}
\newcommand{\rastermodel}{Nemotron-Labs-Image-30B-A3B}
\newcommand{\tokenizermodel}{Nemotron-Labs-Image-Tokenizer}
\newcommand{\codebasemodel}{Nemotron-3-Nano-Omni-30B-A3B}
\newcommand{\supervlmodel}{Nemotron-3.5-Super-VL}
\newcommand{\codefamily}{Nemotron-Labs-VisionCode}
\newcommand{\codemodel}{Nemotron-Labs-VisionCode-30B-A3B}

\definecolor{darkred}{rgb}{0.7, 0.0, 0.0}

\usepackage{pifont}
\usepackage{wrapfig}

\usepackage{fancyhdr} 

\fancypagestyle{authorsfooter}{%
  \fancyfoot{}%
  \fancyfoot[C]{%
    \begin{minipage}[b]{\textwidth}
      \normalfont\footnotesize
      \hrule height 0.4pt
      \vspace{6pt}

      \noindent
      \textsuperscript{*}Equal contribution; listed alphabetically by surname.
      \textsuperscript{$\sharp$}Project lead.
      \textsuperscript{$\dagger$}Work done during internships at NVIDIA.

      \smallskip
      \noindent
      \textbf{Affiliations:}
      \textsuperscript{$\diamond$}NVIDIA.

      \smallskip
      \noindent
      \textbf{Additional affiliations:}
      \textsuperscript{$\ddagger$}University of North Carolina at Chapel Hill;
      \textsuperscript{$\S$}University of California, Santa Cruz;
      \textsuperscript{$\P$}University of Maryland;
      \textsuperscript{$\Vert$}Nanjing University.

      \smallskip
      \noindent
      \textcopyright{} 2026 NVIDIA. All rights reserved.

      \smallskip
      \raggedleft\thepage
    \end{minipage}%
  }%
}

\usepackage[nameinlink]{cleveref}
\crefname{equation}{Eq.}{Eqs.}
\crefname{figure}{Fig.}{Figs.}
\crefname{section}{Sec.}{Sec.}
\crefname{appendix}{App.}{App.}
\crefname{table}{Tab.}{Tabs.}
\crefname{algorithm}{Algo}{Algo}
\crefname{thm}{Thm}{Thm}
\Crefname{thm}{Thm}{Thm}
\crefname{prop}{Prop}{Prop}
\usepackage{longtable}
\usepackage{graphicx}
\usepackage{booktabs}
\usepackage{makecell}
\usepackage{xcolor}
\usepackage{xcolor}
\usepackage{graphicx}
\usepackage{subcaption}
\usepackage{xcolor}
\usepackage{xcolor}
\usepackage{hyperref}
\definecolor{repairred}{RGB}{211,47,47}
\definecolor{repairgreen}{RGB}{20,128,68}

\newcommand{\crefnames}[3]{%
  \@for\next:=#1\do{%
    \expandafter\crefname\expandafter{\next}{#2}{#3}%
  }%
}

\newcommand{\papertitle}{Nemotron-Labs-VisualCreation: Pixel as Code for Agentic Visual Creation}
\renewcommand{\reportdate}{2026-11-20}
\title{\papertitle}
\author{{\large NVIDIA}}

\hypersetup{
  pdftitle={\papertitle},
  pdfauthor={NVIDIA},
  pdfsubject={Technical report on agentic visual creation through discrete visual tokens, structured code, and rendered feedback},
  pdfkeywords={Pixel as Code, agentic visual creation, text-to-image generation, multimodal code generation, image tokenization, discrete visual tokens, SVG generation, visual feedback},
  pdflang={en-US},
  pdfdisplaydoctitle=true,
  bookmarksnumbered=true
}

\begin{abstract}
Visual creation demands expressive imagery together with precise control over text, layout, and structure. We introduce \frameworkname{}, an agentic framework that connects these capabilities through \emph{Pixel as Code}. Raster images are represented by learned discrete visual tokens, while structured artifacts use explicit programs. A shared generation-and-rendering interface connects these representations to visual feedback across images, vector graphics, webpages, and documents. A Nemotron-native harness coordinates planning and raster candidate selection, with targeted repair for editable structured and hybrid artifacts. To support this interface, we develop \tokenizermodel{}, the autoregressive raster generator \rastermodel{}, and the structured-code model family \codefamily{}. On approximately 178K images spanning six visual domains, our tokenizer improves reconstruction PSNR from 26.08 to 29.57\,dB over Emu3.5 IBQ. It also provides $7.13\times$ faster encoding and $10.93\times$ faster decoding at batch size one on A100. In standalone evaluation, one \codefamily{} variant surpasses Kimi K3.1 Pro on the aggregate across Chart2Code, MMSVG, Design2Code, and Image2Struct. At the system level, the Nemotron-native configuration raises GenEval2 Soft-TIFA GM from 31.71 to 52.56 using the same raster generator. With our best setting, \frameworkname{} achieves performance competitive with GPT Image 2 and Nano Banana Pro across most benchmarks with available comparisons.

\par\smallskip
\noindent
{\normalfont\normalsize\mdseries
\href{https://huggingface.co/nvidia/Nemotron-Labs-Image-30B-A3B}{%
  \raisebox{-0.28em}{\includegraphics[height=1.45em]{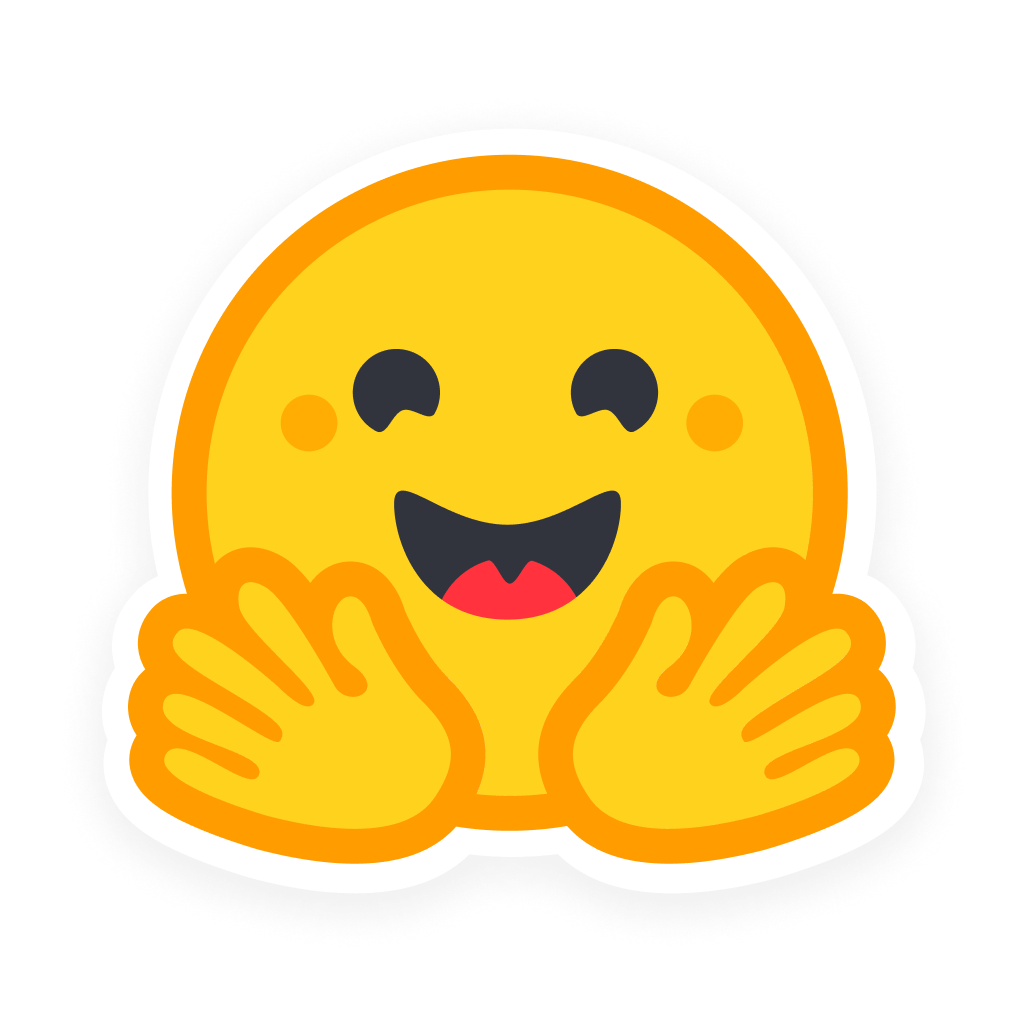}}%
  \hspace{0.65em}%
  \textcolor[rgb]{0,0,0.55}{\texttt{https://huggingface.co/nvidia/Nemotron-Labs-Image-30B-A3B}}%
}
}
\end{abstract}

\begin{document}

\maketitle

\abscontent

\begin{figure}[H]
    \centering
    \begingroup
    \sffamily
\definecolor{ink}{HTML}{243449}
\definecolor{muted}{HTML}{65758A}
\definecolor{raster}{HTML}{628E14}
\definecolor{code}{HTML}{3768B0}
\definecolor{review}{HTML}{AF7922}
    \resizebox{\linewidth}{!}{\input{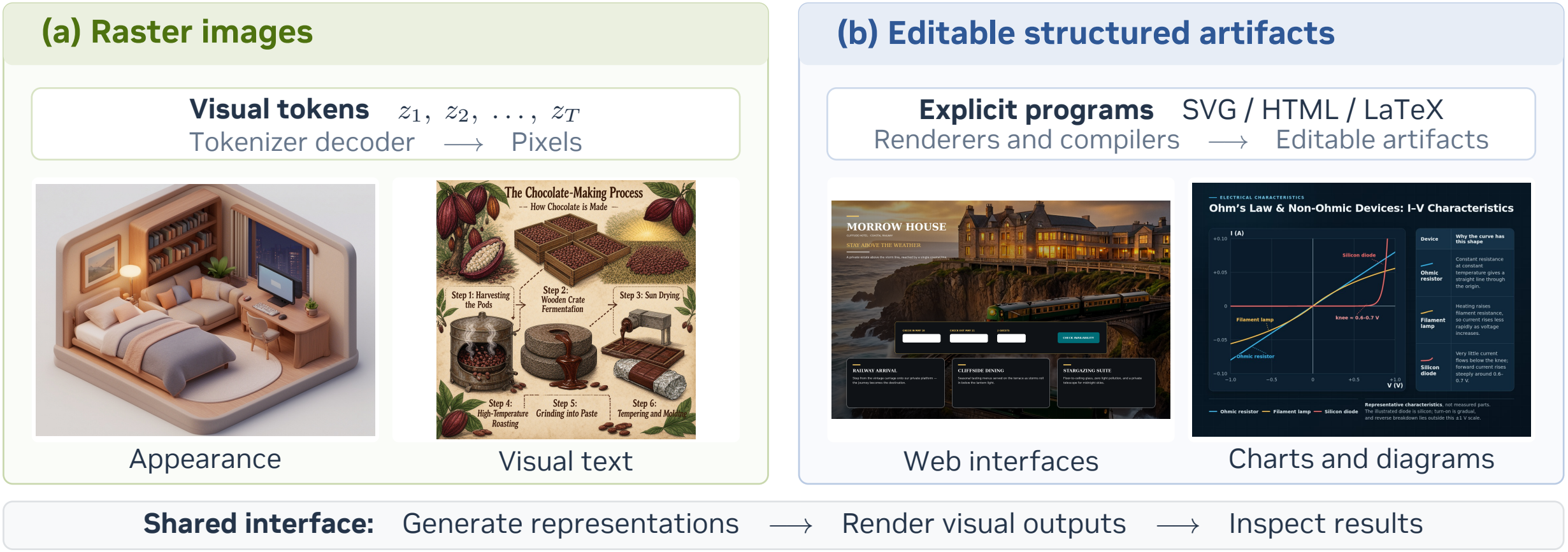}}
    \endgroup
    \caption{\textbf{Pixel as Code at a glance.} Learned visual tokens and explicit programs provide complementary representations for raster images and editable structured artifacts. Their renderers connect both representations to visual outputs that agents can inspect. Hybrid artifacts combine raster assets with structured layouts. Representations are schematic; thumbnails show representative system outputs. Figure~\ref{fig:overall_framework} details the route-specific generation, selection, and repair workflows.}
    \label{fig:teaser}
\end{figure}

\begin{figure}[p]
    \centering
    \includegraphics[pagebox=cropbox,width=\linewidth,height=0.90\textheight,keepaspectratio]{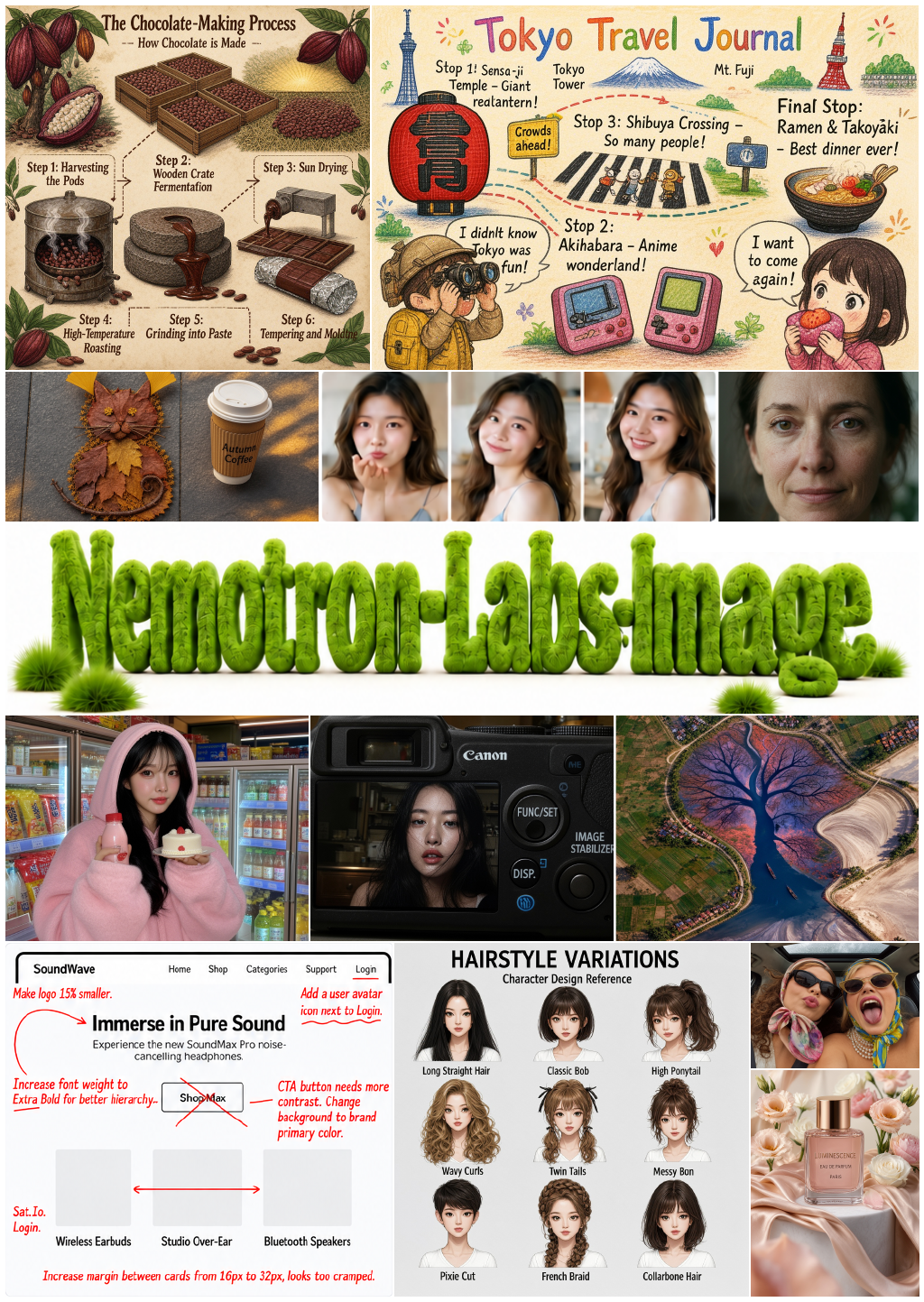}
    \caption{Raster image generation examples from \frameworkname{}. The examples span illustrated infographics, text-rich compositions, portraits, stylized typography, and product imagery, illustrating diverse visual styles and content requirements.}
    \label{fig:image-generation}
\end{figure}

\begin{figure}[p]
    \centering
    \includegraphics[pagebox=cropbox,width=\linewidth,height=0.92\textheight,keepaspectratio]{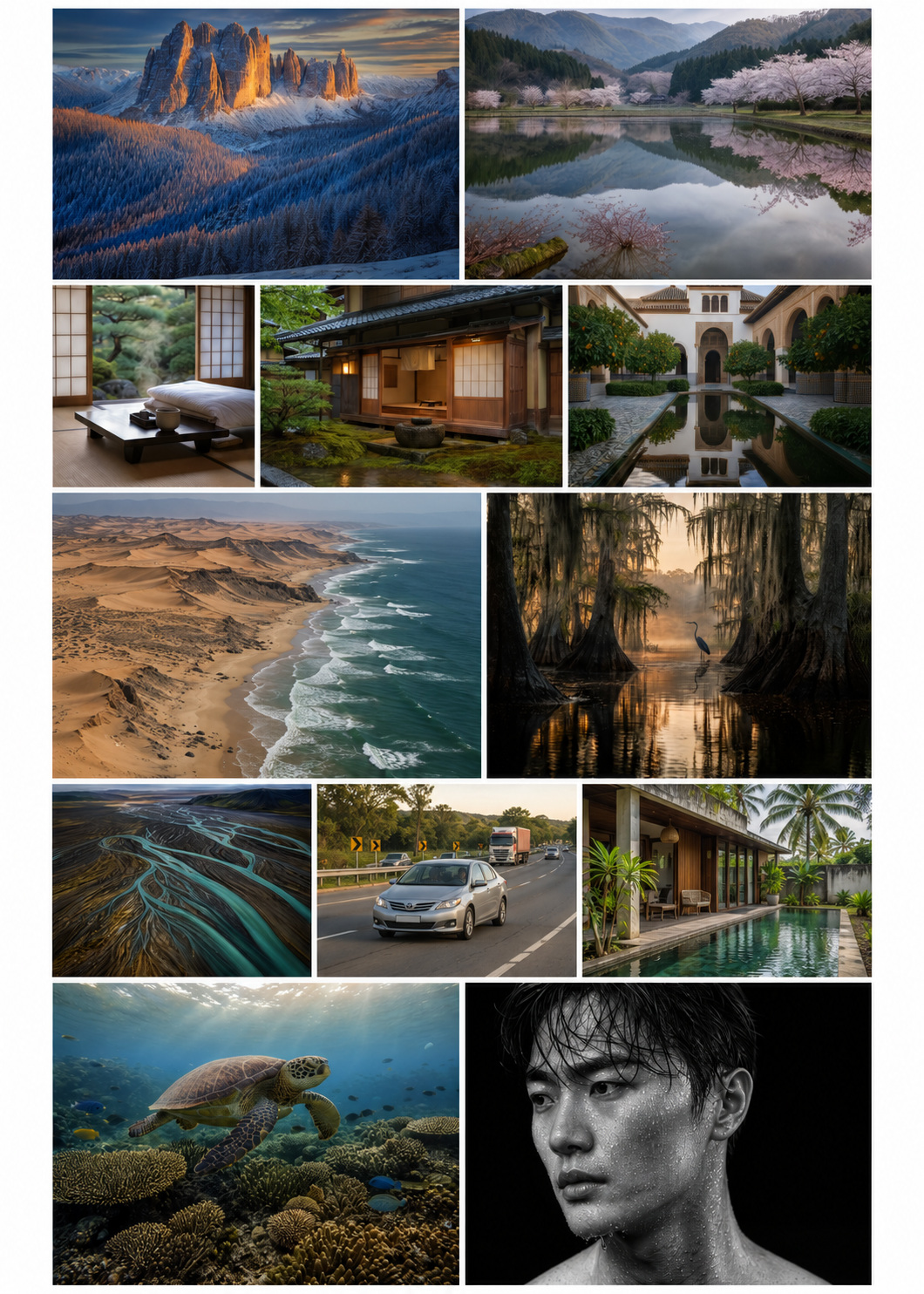}
    \caption{Raster image generation examples from \frameworkname{}, spanning natural landscapes, wildlife, and portraiture. The examples showcase fine visual detail, diverse lighting conditions, and a range of textures and compositions.}
    \label{fig:image-natural}
\end{figure}

\begin{figure}[p]
    \centering
    \includegraphics[pagebox=cropbox,width=\linewidth,height=0.90\textheight,keepaspectratio]{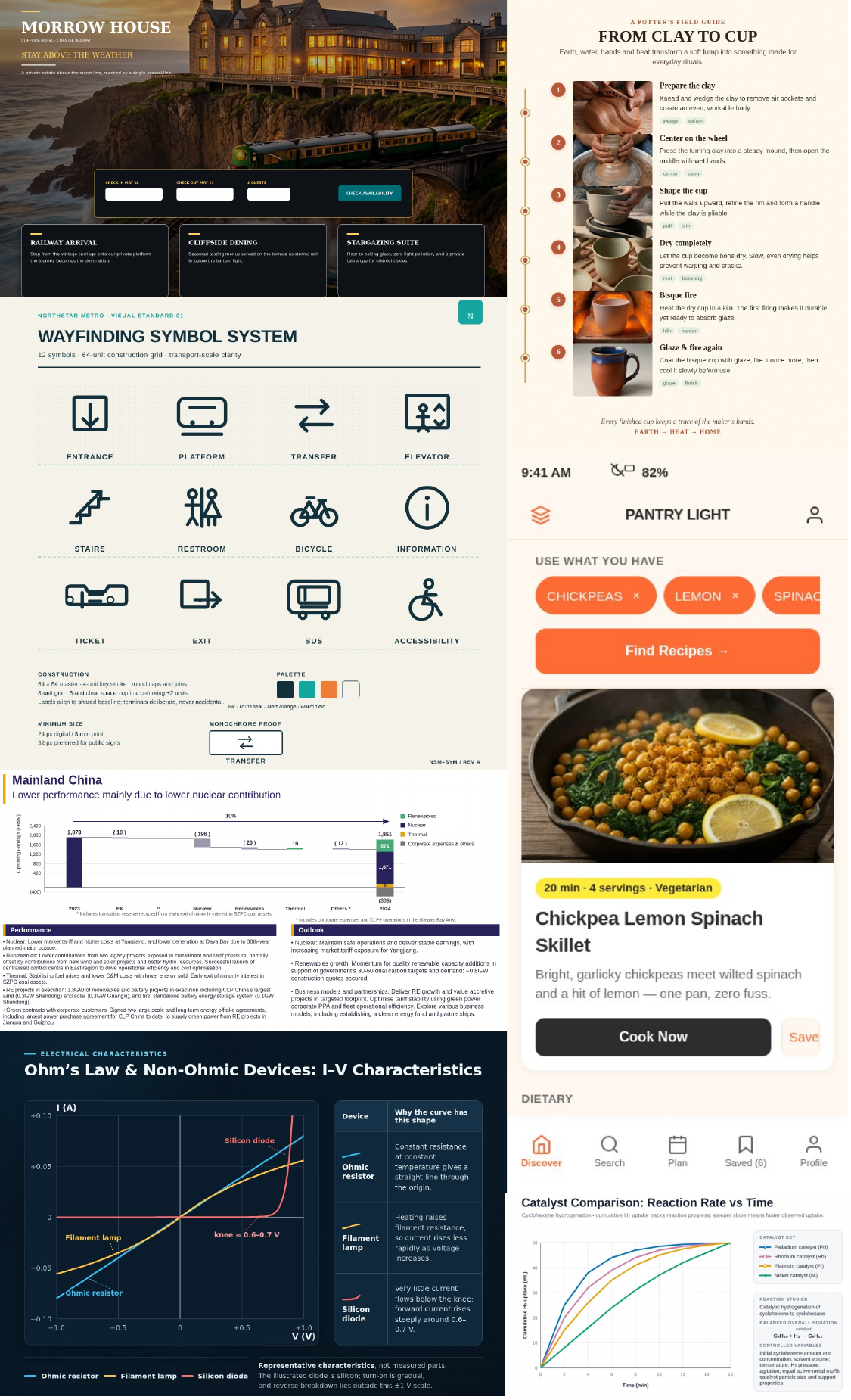}
    \caption{Editable structured artifacts created by \frameworkname{}, including webpages, mobile interfaces, vector icon systems, presentation slides, scientific figures, and charts. The framework combines generated visual assets with explicit representations to compose text, graphics, and layouts that can be rendered and further edited.}
    \label{fig:visual-design}
\end{figure}

\clearpage

\section{Introduction}
\label{sec:intro}

Visual creation spans photographs, illustrations, charts, webpages, posters, scientific figures, and slides. A single request may demand expressive imagery, exact wording, factual accuracy, spatial relationships, and an editable deliverable. Meeting these needs calls for both expressive synthesis and explicit control over content and structure, coordinated through representation, rendering, and visual inspection.

Recent work provides complementary foundations. Edify Image~\citep{nvidia2024edify2d}, Sana~\citep{xie2024sana}, and PixelDiT2~\citep{yu2026pixeldit2} explore representations and architectures for image synthesis, while Cosmos 3~\citep{nvidia2026cosmos3} integrates multimodal understanding and generation. Agentic image creation~\citep{li2026bizgeneval} and scene-generation systems~\citep{xia2026sage,ling2026scenethesis} investigate planning, tool use, and visual assessment. These directions motivate a common interface for coordinating learned visual generation and explicit artifact structure.

We introduce \textbf{Pixel as Code}, a shared interface in which generators predict discrete representations and renderers produce visual outputs for inspection (Figure~\ref{fig:teaser}). Learned visual tokens capture raster appearance and are decoded into pixels. Explicit programs---including SVG, HTML/CSS, plotting code, and document markup---expose content, geometry, and layout for direct editing through their respective rendering backends. Hybrid artifacts combine these complementary representations, for example by embedding a generated image within an editable webpage.

Building on this interface, we present \textbf{\frameworkname{}}. Its harness coordinates semantic planning, raster generation, and visual selection. Structured and hybrid requests additionally establish a layout contract and produce editable representations. For browser-based artifacts, visual audits assign failures to causal owners for targeted repair and checkpoint selection; native formats use their corresponding generation and review workflows. Figures~\ref{fig:image-generation} and~\ref{fig:visual-design} illustrate representative outputs.

Our contributions span the creation interface, its learned components, and their evaluation:

\textbf{A shared interface for agentic visual creation.} We formulate Pixel as Code around a common \emph{generate representation $\rightarrow$ render $\rightarrow$ inspect} interface and describe how a Nemotron-native harness coordinates planning, generation, visual selection, and route-specific refinement. The browser-based workflow makes rendered feedback actionable through a resolved layout contract, explicit repair ownership, and checkpoint retention (Section~\ref{sec:framework}).

\textbf{Complementary models for raster and structured generation.} We develop \tokenizermodel{} as the discrete raster interface, adapt Nemotron 3 Nano into the autoregressive generator \rastermodel{} through multistage mid-training and reinforcement learning, and develop \codefamily{} as a family of specialized models for explicit programs and markup. Sections~\ref{sec:data} and~\ref{sec:train} describe the data and training, with coding variants introduced in Section~\ref{sec:structured_code_training}.

\textbf{Evaluation from reconstruction to complete creation systems.} We assess tokenizer reconstruction across six visual domains, raster generation across five capability categories, and structured code generation across five tasks. One \codefamily{} variant leads the task-balanced aggregate among models evaluated on all five tasks: Chart2Code, MMSVG image-to-SVG and text-to-SVG, Design2Code, and Image2Struct. It surpasses Kimi K3.1 Pro~\citep{moonshot2026kimik31} with fewer total and active parameters (Section~\ref{sec:mm_coding_benchmarks}). At the system level, the Nemotron-native configuration raises GenEval2 Soft-TIFA GM from 31.71 to 52.56 and WISE Verified from 46.30 to 88.60 with the same raster generator. With our best setting, the framework is competitive with GPT Image 2 and Nano Banana Pro across most benchmarks with available comparisons.

\section{Overall Framework}
\label{sec:framework}
\label{sec:harness}

\begin{figure}[!htb]
    \centering
    \begin{adjustbox}{max width=\linewidth}
    \begin{tikzpicture}[
        model/.style={draw=diagramCode!75!black, fill=diagramCode!14,
            text=diagramInk, line width=0.65pt, rounded corners=3pt,
            text width=3.0cm, minimum height=8.5mm,
            align=center, inner sep=4pt, font=\footnotesize\sffamily},
        planning/.style={model, draw=diagramPlan!80!black, fill=diagramPlan!17},
        rastermodel/.style={model, draw=diagramRaster!75!black, fill=diagramRaster!22},
        nativemodel/.style={model, draw=diagramNative!80!black, fill=diagramNative!17},
        review/.style={model, draw=diagramReview!75!black, fill=diagramReview!22},
        service/.style={model, draw=diagramMuted!65, fill=white, dashed},
        output/.style={model, draw=diagramCode!85!black, fill=diagramCode!85!black, text=white},
        heading/.style={font=\small\sffamily\bfseries, text=white, anchor=west},
        flow/.style={-{Latex[length=1.8mm]}, line width=0.65pt, draw=diagramMuted},
        repairflow/.style={flow, draw=diagramReview!80!black, line width=0.9pt},
        panel/.style={rounded corners=4pt, line width=0.5pt}
    ]
        \draw[panel, draw=diagramPlan!35, fill=diagramPlan!5] (-1.8,1.35) rectangle (1.8,-7.85);
        \draw[panel, draw=diagramRaster!45, fill=diagramRaster!6] (2.3,1.35) rectangle (13.85,-0.45);
        \draw[panel, draw=diagramCode!35, fill=diagramCode!5] (2.3,-1.0) rectangle (13.85,-5.65);
        \draw[panel, draw=diagramNative!35, fill=diagramNative!6] (2.3,-6.15) rectangle (13.85,-7.85);
        \fill[rounded corners=3pt, diagramPlan!85!black] (-1.8,1.35) rectangle (1.8,0.66);
        \fill[rounded corners=3pt, diagramRaster!72!black] (2.3,1.35) rectangle (13.85,0.66);
        \fill[rounded corners=3pt, diagramCode!85!black] (2.3,-1.0) rectangle (13.85,-1.64);
        \fill[rounded corners=3pt, diagramNative!85!black] (2.3,-6.15) rectangle (13.85,-6.79);
        \node[heading] at (-1.65,1.0) {\diagramicon{route}Route and plan};
        \node[heading] at (2.5,1.0) {\diagramicon{image}Direct raster};
        \node[heading] at (2.5,-1.35) {\diagramicon{laptop-code}Browser-based structured / hybrid artifacts};
        \node[heading] at (2.5,-6.48) {\diagramicon{file-alt}Native presentation / document / SVG};

        \node[planning] (route) at (0,0.05)
            {\diagramicon{project-diagram}\textbf{Artifact routing}\\Super-VL};
        \node[service] (retrieve) at (0,-1.35)
            {\diagramicon{search}Conditional search\\and reference\\retrieval};
        \node[planning] (plan) at (0,-2.85)
            {\diagramicon{tasks}\textbf{Semantic planning}\\Ultra};
        \node[font=\footnotesize\sffamily, text=diagramMuted, align=center, text width=3.2cm] at (0,-4.4)
            {Choose artifact family,\\representation, renderer,\\and asset policy};
        \node[font=\footnotesize\sffamily, text=diagramInk, align=center] (input) at (0,1.95)
            {User request and references};
        \draw[flow] (input.west) -- (-2.05,1.95) |- (route.west);
        \draw[flow] (route) -- (retrieve);
        \draw[flow] (retrieve) -- (plan);

        \node[rastermodel] (images) at (4.1,0.05)
            {\diagramicon{images}\textbf{Raster candidates}\\Image};
        \node[review] (judge) at (8.1,0.05)
            {\diagramicon{eye}\textbf{Visual selection}\\Super-VL};
        \node[output, draw=diagramRaster!62!black, fill=diagramRaster!62!black] (raster) at (12.1,0.05)
            {\diagramicon{check-circle}\textbf{Selected raster image}};
        \draw[flow] (images) -- (judge);
        \draw[flow] (judge) -- (raster);

        \node[planning, text width=3.2cm, inner sep=3pt, font=\scriptsize\sffamily] (layout) at (4.1,-2.35)
            {\diagramicon{columns}\textbf{Layout and assets}\\Frozen plan + manifest\\Ultra; Image as needed};
        \node[font=\footnotesize\sffamily, text=diagramMuted, align=left, text width=6.8cm] at (10.0,-2.35)
            {Freeze request checklist and layout contract.\\Both coders receive the same plan and assets.};
        \node[model] (coder) at (4.1,-3.55)
            {\diagramicon{code}\textbf{Dual Coder}\\Ultra + Super-VL};
        \node[review] (audit) at (8.1,-3.55)
            {\diagramicon{desktop}\textbf{Render and audit}\\Renderer checks + Super-VL};
        \node[review, text width=3.2cm] (select) at (12.1,-3.55)
            {\diagramicon{clipboard-check}\textbf{Guard \& compare}\\Pairwise judge: Super-VL};
        \node[review] (repair) at (8.1,-5.0)
            {\diagramicon{tools}\textbf{Validate and repair}\\Assign causal owner};
        \node[output, text width=3.2cm] (best) at (12.1,-5.0)
            {\diagramicon{check-circle}\textbf{Best editable}\\\textbf{artifact}\\Retained checkpoint};
        \draw[flow] (layout) -- (coder);
        \draw[flow] (coder) -- (audit);
        \draw[flow] (audit) -- (select);
        \draw[flow] (select) -- (best);
        \draw[repairflow] (audit) -- (repair);
        \draw[repairflow] (repair.west) -- (4.1,-5.0) -- (coder.south);

        \node[nativemodel] (nativegen) at (4.1,-7.3)
            {\diagramicon{file-code}\textbf{Format-specific generation}};
        \node[service, draw=diagramNative!65, fill=diagramNative!10] (nativerender) at (8.1,-7.3)
            {\diagramicon{eye}Render, inspect,\\and judge by format};
        \node[output, draw=diagramNative!85!black, fill=diagramNative!85!black] (nativeout) at (12.1,-7.3)
            {\diagramicon{check-circle}\textbf{Native artifact}};
        \draw[flow] (nativegen) -- (nativerender);
        \draw[flow] (nativerender) -- (nativeout);

        \draw[semithick, draw=diagramMuted] (plan.east) -- (2.02,-2.85);
        \draw[flow] (2.02,-2.85) |- (images.west);
        \draw[flow] (2.02,-2.85) |- (layout.west);
        \draw[flow] (2.02,-2.85) |- (nativegen.west);
    \end{tikzpicture}
    \end{adjustbox}
    \caption{Route-specific control flow in the Nemotron-native creation harness. Ultra denotes Nemotron 3 Ultra, Super-VL denotes \supervlmodel{}, and Image denotes \rastermodel{}. Direct raster outputs use visual candidate selection. Browser artifacts use the Dual Coder and a bounded repair loop; repairs may update assets, the plan, or code according to the diagnosed cause. Native formats use their own generation and review workflows. \codefamily{} is evaluated separately in Section~\ref{sec:mm_coding_benchmarks}.}
    \label{fig:overall_framework}
\end{figure}
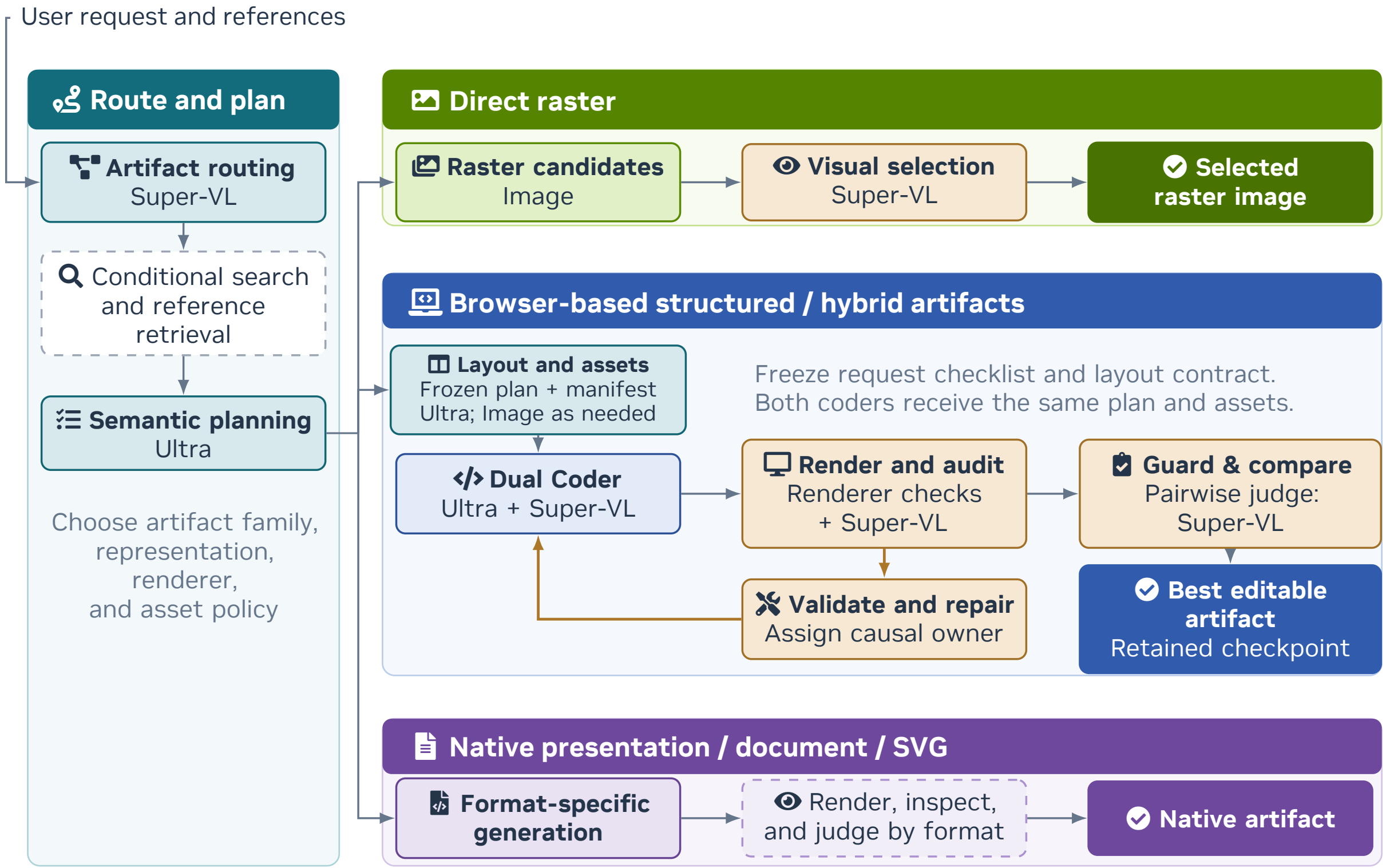

\frameworkname{} turns a user request and optional references into raster images or editable structured artifacts. Routing selects the artifact family, target surface, representation, and visual-asset policy. Direct raster requests use candidate generation and visual selection. Structured and hybrid requests additionally establish a layout contract, generate executable representations, and support targeted repair before checkpoint selection. The shared \textbf{Pixel as Code} interface connects these routes through generation, rendering, and visual inspection. Figure~\ref{fig:overall_framework} shows their control flow and model assignments; Section~\ref{sec:components} describes the components and rendering backends.

\subsection{Pixel as Code: Discrete Representations and Renderers}
\label{sec:framework_task}

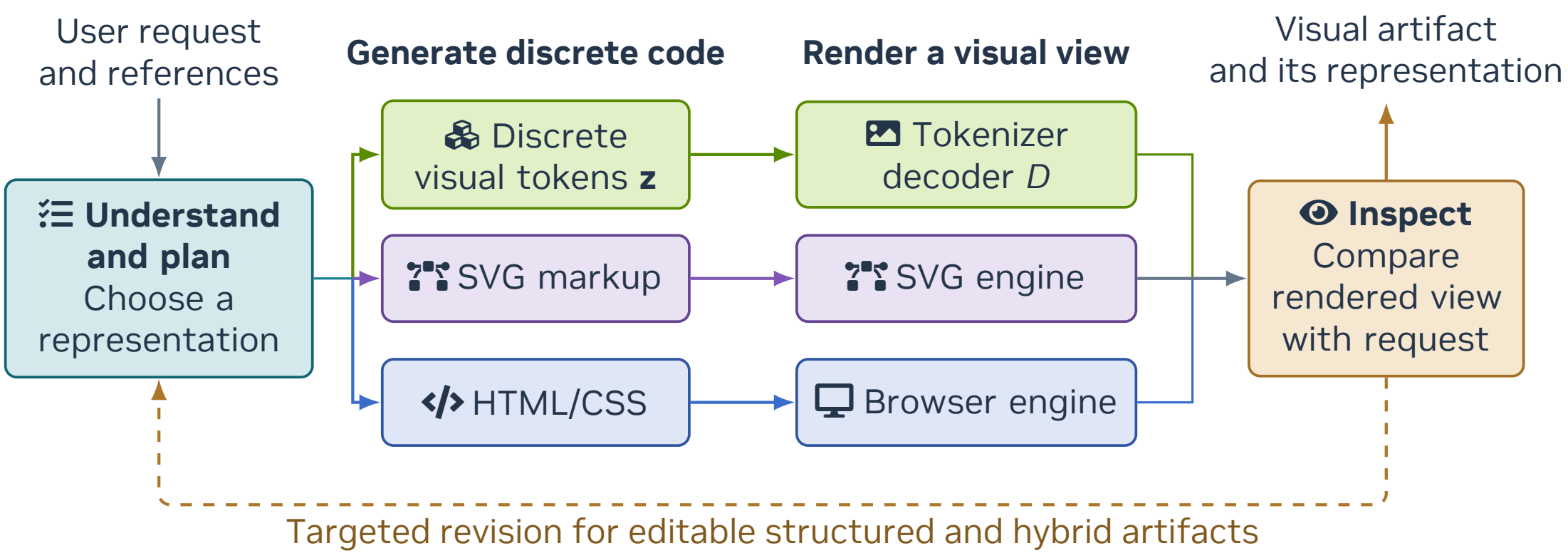
\begin{figure}[!htb]
    \centering
    \begin{adjustbox}{max width=\linewidth}
    \begin{tikzpicture}[
        stage/.style={draw=diagramMuted, fill=white, text=diagramInk,
            rounded corners=3pt, line width=0.7pt,
            align=center, inner sep=5pt, font=\small\sffamily},
        raster/.style={draw=diagramRaster!75!black, fill=diagramRaster!22},
        vector/.style={draw=diagramNative!80!black, fill=diagramNative!17},
        web/.style={draw=diagramCode!80!black, fill=diagramCode!16},
        code/.style={stage, text width=2.5cm, minimum height=8mm},
        renderer/.style={stage, text width=2.8cm, minimum height=8mm},
        flow/.style={-{Latex[length=2mm]}, line width=0.8pt, draw=diagramMuted}
    ]
        \node[stage, draw=diagramPlan!80!black, fill=diagramPlan!18, text width=2.5cm, minimum height=16mm] (plan) at (0,0)
            {\diagramicon{tasks}\textbf{Understand}\\\textbf{and plan}\\Choose a representation};
        \node[code, raster] (bitmap) at (3.5,1.15) {\diagramicon{cubes}Discrete\\visual tokens \textbf{z}};
        \node[code, vector] (svg) at (3.5,0) {\diagramicon{bezier-curve}SVG markup};
        \node[code, web] (html) at (3.5,-1.15) {\diagramicon{code}HTML/CSS};
        \node[renderer, raster] (decoder) at (7.5,1.15) {\diagramicon{image}Tokenizer\\decoder \textit{D}};
        \node[renderer, vector] (svgrender) at (7.5,0) {\diagramicon{bezier-curve}SVG engine};
        \node[renderer, web] (browser) at (7.5,-1.15) {\diagramicon{desktop}Browser engine};
        \node[stage, draw=diagramReview!75!black, fill=diagramReview!23, text width=2.2cm, minimum height=16mm] (inspect) at (11.4,0)
            {\diagramicon{eye}\textbf{Inspect}\\Compare rendered view with request};
        \node[font=\small\sffamily\bfseries, text=diagramInk] at (3.5,2.1) {Generate discrete code};
        \node[font=\small\sffamily\bfseries, text=diagramInk] at (7.5,2.1) {Render a visual view};
        \node[font=\small\sffamily, text=diagramInk, align=center] (request) at (0,2.1) {User request\\and references};
        \node[font=\small\sffamily, text=diagramInk, align=center] (output) at (11.4,2.1) {Visual artifact\\and its representation};
        \draw[flow] (request) -- (plan);
        \draw[semithick, draw=diagramPlan] (plan.east) -- (1.8,0);
        \draw[flow, draw=diagramRaster!75!black] (1.8,0) |- (bitmap.west);
        \draw[flow, draw=diagramNative] (1.8,0) -- (svg.west);
        \draw[flow, draw=diagramCode] (1.8,0) |- (html.west);
        \draw[flow, draw=diagramRaster!75!black] (bitmap) -- (decoder);
        \draw[flow, draw=diagramNative] (svg) -- (svgrender);
        \draw[flow, draw=diagramCode] (html) -- (browser);
        \draw[semithick, draw=diagramRaster!75!black] (decoder.east) -- (9.6,1.15) -- (9.6,0);
        \draw[semithick, draw=diagramCode] (browser.east) -- (9.6,-1.15) -- (9.6,0);
        \draw[flow] (svgrender.east) -- (inspect.west);
        \draw[flow, draw=diagramReview!80!black] (inspect) -- (output);
        \draw[flow, dashed, rounded corners=3pt, draw=diagramReview!80!black] (inspect.south) -- (11.4,-2.1) -- (0,-2.1) -- (plan.south);
        \node[font=\small\sffamily, text=diagramReview!60!black, below] at (5.7,-2.1) {Targeted revision for editable structured and hybrid artifacts};
    \end{tikzpicture}
    \end{adjustbox}
    \caption{The representation--renderer interface underlying the creation harness. Learned visual tokens, SVG, and HTML/CSS are discrete representations with corresponding rendering backends. All routes support visual inspection; the dashed return path denotes revision of editable structured and hybrid artifacts. Direct raster outputs use candidate selection, as shown in Figure~\ref{fig:overall_framework}.}
    \label{fig:representation_renderers}
\end{figure}

\begin{samepage}
Let $q$ denote the user request, $v$ optional reference material, and $r$ the chosen representation. Visual creation produces a discrete representation $c_r$ and a rendered view $\hat{I}$:
\begin{equation}
    c_r \sim G_r(\,\cdot \mid q,v),
    \qquad
    \hat{I}=R_r(c_r),
    \label{eq:code_render_interface}
\end{equation}
where $G_r$ generates the representation and $R_r$ renders it (Figure~\ref{fig:representation_renderers}). Here, code refers to either learned discrete visual tokens or an explicit artifact representation. This abstraction does not require all representations to share a vocabulary or renderer. Instead, it exposes a common interface through which an artifact can be executed, rendered, visually inspected, and, when its representation is editable, revised.
\end{samepage}

For the \textbf{raster image route}, the image tokenizer provides both the discrete representation and the rendering interface. Its encoder $E$ maps an image $I$ to discrete visual tokens $\mathbf{z}=E(I)$, while its decoder $D$ reconstructs a raster image $\hat{I}=D(\mathbf{z})$. This separation between discrete representation learning and generation builds on vector-quantized autoencoders~\citep{oord2017vqvae,esser2021taming}; HART~\citep{tang2025hart} further studies hybrid discrete and continuous visual representations. \rastermodel{} serves as the \textbf{raster image generator}: it autoregressively predicts discrete visual tokens, and the tokenizer decoder acts as their learned renderer. Encoding supplies representations for reference images and training targets, whereas decoding converts generated codes into visible pixels. For a direct raster request, the rendered candidates are final artifact candidates.

For \textbf{structured routes}, the generated representation explicitly exposes content, geometry, and layout through formats such as SVG, HTML/CSS/JavaScript, presentation objects, document structures, or plotting programs. Corresponding SVG engines, browsers, office renderers, plotting runtimes, and document compilers convert these representations into visual evidence. Because the source representation remains accessible, localized edits can be applied after visual inspection. A \textbf{hybrid route} additionally embeds one or more generated raster assets within a structured artifact while preserving text, shapes, layout, and interactions in an editable representation. For browser-based artifacts, the Nemotron-native configuration assigns code generation to a Dual Coder composed of Nemotron 3 Ultra~\citep{blakeman2026nemotron} and \supervlmodel{}. Native presentation, document, and SVG outputs use format-specific generators and renderers. We separately train \codefamily{} for this interface (Section~\ref{sec:structured_code_training}) and evaluate the models in standalone tests (Section~\ref{sec:mm_coding_benchmarks}).

\FloatBarrier
\subsection{Agentic Creation Loop}
\label{sec:framework_loop}

Prior agentic scene-generation systems combine planning with visual feedback: SAGE~\citep{xia2026sage} coordinates generation tools and critics, and Scenethesis~\citep{ling2026scenethesis} integrates language-based planning, visual layout refinement, and scene judging. Our harness makes this feedback operational through a resolved layout contract, renderer-grounded evidence, a root-cause report, and explicit repair ownership (Figure~\ref{fig:overall_framework}).

\textbf{Routing and structured planning.}
\supervlmodel{} classifies each request as a natural image, structured canvas, interactive interface, multi-page document, or visual series. It jointly determines the target surface, output representation, canvas count, rendering backend, and whether generated raster assets are absent, optional, or required. Search and reference retrieval are invoked conditionally, only when external factual or visual grounding materially affects the request. Nemotron 3 Ultra then produces a semantic artifact plan. For structured and hybrid routes, this plan is resolved into a layout contract that specifies the layout mode, ordered regions, normalized bounds, component ownership, and rendering constraints. 

\textbf{Direct raster creation.}
Natural-image requests bypass structured code generation. \rastermodel{} produces multiple raster candidates from the planned visual brief, optional references, and a supported aspect ratio. Acting as the multimodal judge, \supervlmodel{} compares the rendered candidates against the original request and retains the strongest image. This route therefore uses semantic planning and visual selection, but does not invoke the structured artifact repair loop described below.

\textbf{Structured and hybrid artifact creation.}
Structured artifacts use explicit code or native document representations, while hybrid artifacts additionally request raster assets from \rastermodel{}. In the browser-based Dual Coder configuration, Nemotron 3 Ultra and \supervlmodel{} independently instantiate the same frozen plan and asset manifest. Native presentation, document, and SVG routes use their corresponding format-specific generation workflows. Each candidate is executed or rendered and visually inspected before an eligible candidate is retained as the initial artifact checkpoint $C_0$. Raster generation supplies requested image content, whereas the structured representation determines how assets, text, shapes, and interactions are composed into an editable artifact.

\begin{figure}[t]
    \centering

    \begin{subfigure}{0.96\textwidth}
        \centering
        \includegraphics[width=\linewidth]
        {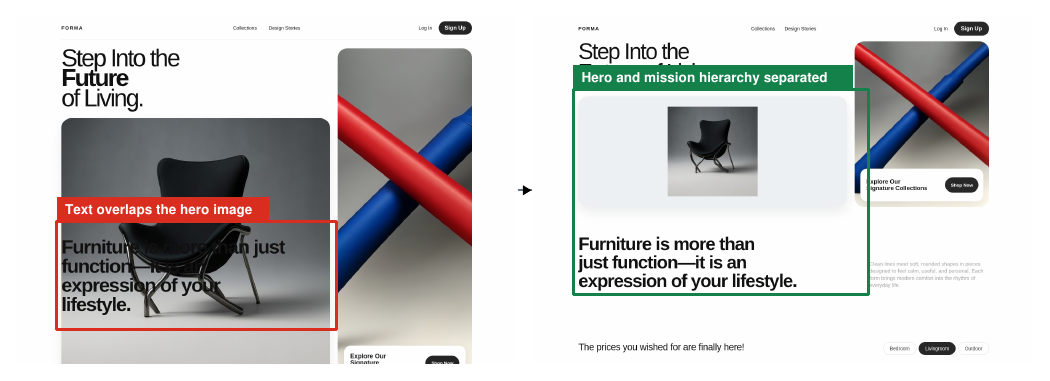}
        \caption{Layout repair resolves text--image overlap and restores the visual hierarchy.}
        \label{fig:repair-furniture}
    \end{subfigure}

    \vspace{2mm}

    \begin{subfigure}{0.96\textwidth}
        \centering
        \includegraphics[width=\linewidth]
        {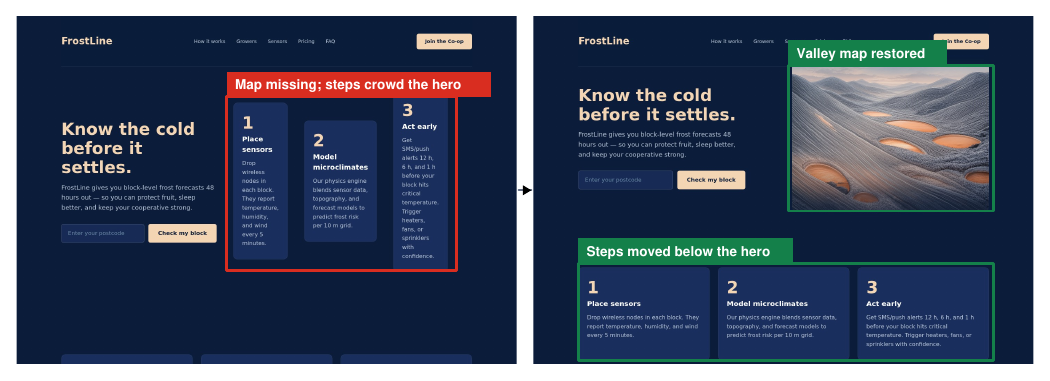}
        \caption{Asset and layout repair restores the missing visual and relocates the workflow steps.}
        \label{fig:repair-frostline}
    \end{subfigure}

    \vspace{2mm}

    \begin{subfigure}{0.96\textwidth}
        \centering
        \includegraphics[width=\linewidth]
        {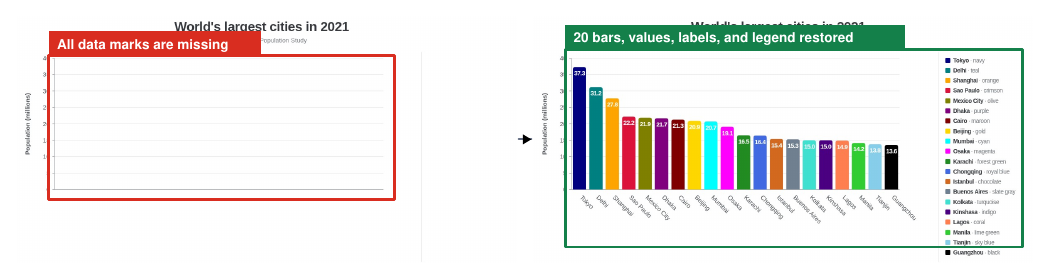}
        \caption{Code and content repair restores the chart marks, labels, values, and legend.}
        \label{fig:repair-chart}
    \end{subfigure}

    \caption{
Representative examples of self-correction.
\textbf{\textcolor{repairred}{Red}} annotations identify issues in the
initial renderings, while
\textbf{\textcolor{repairgreen}{green}} annotations highlight the
corresponding repairs.
}
    \label{fig:self-correction-examples}
\end{figure}

\textbf{Evidence-grounded visual audit and targeted repair.}
For browser-based artifacts, the audit combines deterministic renderer evidence with a multimodal review against the frozen request and acceptance checklist. Renderer evidence identifies failures such as execution errors, missing regions, clipping, overlap, out-of-bounds geometry, and broken interactions. \supervlmodel{} evaluates the visible artifact and proposes a diagnosis code, causal evidence, a confidence score, and an owner within the rendered structure. Before automatic repair is applied, the harness validates this proposal against stable artifact selectors. Native formats use their own rendering, inspection, and review workflows.

The browser-based repair workflow distinguishes the visible \emph{diagnosis} from the underlying \emph{repair target}. Diagnosis codes cover component rendering, collision, containment, spatial density, typography, color contrast, layout structure, content fidelity, visual-asset quality, and interaction behavior. Repair targets identify the responsible layer: \emph{visual plan, asset, layout, code, interaction, or content}. Findings associated with the same causal owner are grouped into a root-cause tree, allowing the harness to repair a shared cause while freezing unaffected pages or regions. Asset-related findings can trigger raster regeneration; planning or layout findings can trigger a localized replan; and code, interaction, or content findings are assigned to the structured generation path. Every resulting candidate is executed, rendered, audited, and compared with the retained checkpoint. The numbers of accepted repair rounds and total attempted repair cycles are bounded separately.

\textbf{Checkpoint retention and output selection.}
For exposition, let $C_t=(P_t,A_t,c_t,\hat{I}_t)$ denote the artifact state after round $t$, comprising its plan, assets, explicit representation, and rendered view. This tuple is a conceptual summary; the implementation additionally retains the frozen checklist, deterministic inspection results, multimodal review, and repair-validation records. Best-candidate retention can be expressed as
\begin{equation}
    C^{\star}_0=C_0,\qquad
    C^{\star}_t=\operatorname{Select}\!\left(q,C^{\star}_{t-1},C_t\right).
    \label{eq:artifact_checkpoint_selection}
\end{equation}
For ordinary browser-based requests, the selection operator first applies deterministic fatal-render guards and then performs a blind pairwise comparison between an eligible candidate and the incumbent using the same frozen request checklist. A tie or an insufficient-confidence decision retains the earlier incumbent. Native-format routes retain the strongest rendered candidate according to their format-specific inspection and judging criteria. The loop terminates when the audit passes, no eligible improvement is found, or the configured repair budget is exhausted. The retained checkpoint provides the editable artifact together with its rendered preview.

\textbf{Implementation details.}
Before generation, the browser-based workflow freezes the request checklist and resolved layout contract. A candidate passes only when both deterministic rendering checks and the multimodal visual audit pass. The default harness configuration generates up to four candidates for each raster-generation job. The browser-based repair loop permits up to three accepted repairs within eight attempted repair cycles and retains the incumbent when pairwise selection returns a tie or confidence below $0.75$. Model and renderer assignments are summarized in Section~\ref{sec:components}. The routing and review prompt contracts, retrieval policies, machine-readable
schemas, and harness control limits are documented in
Appendix~\ref{app:harness_details}.

\textbf{Qualitative self-correction examples.}
Figure~\ref{fig:self-correction-examples} shows three representative
before-and-after repairs. In the furniture landing page, the agent resolves
text--image overlap and restores the content hierarchy. In the FrostLine
landing page, it restores the missing visual asset and reorganizes the hero
and workflow sections. In the city-population chart, it recovers all 20 bars
together with their values, city labels, and legend entries. These examples
show how the agent translates rendered feedback into targeted edits while
preserving the unaffected parts of the design.

\FloatBarrier

\section{System Components}
\label{sec:components}

The framework combines planning, discrete representation generation, rendering, and visual inspection. We describe these component responsibilities below. Section~\ref{sec:data} presents data for both generation routes, and Section~\ref{sec:train} describes training for the image tokenizer, raster image generator, and structured code generator.

\subsection{Planning and Tool Coordination}
\label{sec:planner}

\textbf{\supervlmodel{}}~\citep{nvidia2026nemotron35} first performs artifact routing, and \textbf{Nemotron 3 Ultra}~\citep{blakeman2026nemotron} serves as the planner. Ultra turns the request and any retrieved context into a semantic plan. Structured and hybrid routes additionally resolve a layout contract with ordered regions, normalized bounds, component ownership, and rendering constraints. During repair, Ultra receives issues assigned to plan revision and updates the affected specification. Harness services supply search and reference retrieval when external grounding is needed. The shared interface in Equation~\ref{eq:code_render_interface} lets planning coordinate raster assets and explicit code within the same artifact.

\subsection{Discrete Representation Generators}

\subsubsection{Raster Image Generator}
\label{sec:imgmodel}

\textbf{\rastermodel{}} is an autoregressive raster image generator. Given text and optional source or reference images, it predicts a discrete visual-token sequence that the tokenizer decoder renders into a raster image. Training includes text-to-image generation and text-and-image-conditioned generation (TI2I) with flexible aspect ratios. TI2I covers image editing, multi-turn generation, and synthesis from one or more visual references.

The generator is initialized from Nemotron 3 Nano 30B-A3B and retains its hybrid Mamba2--Transformer MoE architecture and attention configuration~\citep{nvidia2025nemotron3nano,dao2024mamba2}. Text prompts, reference images, and target images are represented as discrete tokens within a unified causal sequence. Section~\ref{sec:bitmap_training} describes its multistage mid-training and reinforcement learning.

\subsubsection{Structured Code Generator}
\label{sec:mm_coding}

The multimodal-coding component generates representations with explicit content and structure: plotting programs for charts, SVG from images or text, HTML/CSS from visual designs, edits to existing webpage code, and document markup. These outputs are interpreted by the corresponding rendering backends described below.

For browser-based artifacts, the \textbf{Dual Coder} combines \textbf{Nemotron 3 Ultra} and \textbf{\supervlmodel{}} for code generation and code or layout repair. The two models independently materialize the same frozen plan and asset manifest, then revise the code when visual diagnosis assigns an issue to this component. Native presentation, document, and SVG routes use format-specific generation workflows.

\textbf{\codefamily{}} is the model family we train and evaluate for the same structured-code interface. Its multimodal-coding supervision covers the explicit representations and editing operations needed by these tasks (Section~\ref{sec:structured_coding_data}). Section~\ref{sec:structured_code_training} introduces the model variants and their base models. Section~\ref{sec:mm_coding_benchmarks} reports their standalone execution and quality results against reference models. These measurements evaluate the trained coding models; the Dual Coder is a separate harness-level model assignment.

\subsection{Rendering Backends}

\subsubsection{Learned Image Renderer: Tokenizer Decoder}
\label{sec:bitmap_renderer}

The dedicated image tokenizer, \textbf{\tokenizermodel{}}, has approximately 642M parameters and uses a fixed Leech-lattice codebook~\citep{zhao2024bsq,zhao2025leech}. Its encoder discretizes a raster image into a compact visual-token sequence; its decoder is the learned image renderer that converts predicted tokens back to pixels. This separates prediction of the discrete representation from its visual realization.

The encoder and decoder follow a vision-transformer design~\citep{dosovitskiy2021image}, each with 24 layers, a hidden width of 1,024, 16 attention heads, and an MLP ratio of 4. The tokenizer uses $16\times16$ patches and 24-dimensional quantized embeddings with L2 normalization. Its codebook contains 196,560 normalized Leech-lattice vectors and remains fixed during training. Rotary position embeddings~\citep{su2021roformer} use a base of 10,000. The decoder uses an output FFN with a configured dimension of 3,072 and clamps reconstructed values to $[-1,1]$. Table~\ref{tab:tokenizer_architecture} summarizes the architecture and image interface; Section~\ref{sec:tokenizer_training} describes the training objectives and optimization schedule.

\begin{table}[!htbp]
    \centering
    \caption{\tokenizermodel{} architecture. The encoder produces discrete visual codes and the decoder renders them into raster images.}
    \label{tab:tokenizer_architecture}
    \begingroup
    \small
    \setlength{\tabcolsep}{6pt}
    \renewcommand{\arraystretch}{1.10}
    \begin{tabularx}{\linewidth}{@{}lX@{}}
        \toprule
        Hyperparameter & Value \\
        \midrule
        \multicolumn{2}{@{}l}{\textbf{Transformer architecture}} \\
        Implementation class & \texttt{VITVQModel} \\
        Parameters & Approximately 642M \\
        Encoder / decoder layers & 24 / 24 \\
        Hidden width / attention heads & 1,024 / 16 \\
        MLP ratio & 4 \\
        Decoder output FFN & Enabled; output FFN dimension 3,072 \\
        RoPE $\theta$ / dropout & 10,000 / 0 \\
        \midrule
        \multicolumn{2}{@{}l}{\textbf{Discrete representation}} \\
        Patch size & $16\times16$ \\
        Quantized embedding dimension & 24 \\
        Codebook & Normalized Leech lattice; 196,560 entries \\
        Codebook weights / L2 normalization & Frozen / enabled \\
        \midrule
        \multicolumn{2}{@{}l}{\textbf{Image interface}} \\
        Encode / decode resize & None / none \\
        Zero-mean input & Enabled \\
        Output clamp & $[-1,1]$ \\
        \bottomrule
    \end{tabularx}
    \endgroup
\end{table}

Section~\ref{sec:tokenizer_evaluation} evaluates the reconstruction quality, single-image latency, and encoding/decoding throughput of this learned raster interface.

\subsubsection{Structured Rendering Backends}

SVG engines render vector markup~\citep{w3c2018svg2}, browser engines render HTML/CSS/JavaScript~\citep{whatwg2026html}, and office renderers, plotting runtimes, or document compilers render their respective representations. These backends play the same interface role as the tokenizer decoder: they map generated discrete representations to visual outputs that can be inspected. Their source representations also expose elements such as shapes, text, and layout for targeted edits. The harness executes and renders revised representations after repair and preserves artifact checkpoints for subsequent comparison.

\subsection{Multimodal Perception and Evaluation}
\label{sec:multimodal_evaluator}

\textbf{\supervlmodel{}} provides visual assessment and candidate selection in addition to its routing and coding roles. Direct raster requests use it to select an image. In the browser-based repair loop, its audit evaluates the rendered artifact against the frozen request checklist; the root-cause tree groups issues by causal owner for assignment to the responsible component; and pairwise selection identifies the artifact state to retain. Deterministic rendering checks and validation of artifact selectors supplement the multimodal review. Native formats use their own inspection and judging workflows. These responsibilities share rendered visual evidence but follow the route-specific control flow in Section~\ref{sec:framework_loop}.

\section{Training Data}
\label{sec:data}

The data pipeline follows the three trainable components of the framework: the image tokenizer learns the discrete raster representation and its renderer; \rastermodel{} learns to predict visual tokens from conditioning inputs; and \codefamily{} learns to generate explicit programs and markup. We describe the data for each component in this order, following the representation--renderer interface of Section~\ref{sec:framework}.

\subsection{Image Tokenizer Data}
\label{sec:tokenizer_data}

We curate an approximately \textbf{200M-image pool} from the NVCLIP image corpus~\citep{nvidianvclip} through clustering, deduplication, and filtering using SigLIP~2~\citep{tschannen2025siglip2}, DINOv2~\citep{oquab2023dinov2}, and SSCD~\citep{pizzi2022sscd}. We train the tokenizer on approximately \textbf{153M images} from this pool.

Images provide reconstruction targets for learning the encoder and decoder described in Section~\ref{sec:bitmap_renderer}. The resulting encoder supplies discrete visual tokens for raster image generation, while the decoder renders predicted tokens back into images. Section~\ref{sec:tokenizer_training} details the reconstruction, perceptual, distillation, and adversarial objectives.

\subsection{Raster Image Generation Data}
\label{sec:raster_data}

The raster image data pipeline combines text-to-image generation (T2I) with text-and-image-conditioned generation (TI2I). T2I conditions on a textual description; TI2I additionally conditions on one or more source or reference images. TI2I covers image editing, generation from visual references, and multi-turn generation using preceding visual context. Both task families use the image tokenizer to convert target images into the discrete visual codes predicted by the generator introduced in Section~\ref{sec:imgmodel}.

\subsubsection{Text-to-Image Data}

\textbf{Data Collection.}
The web-scale portion of our text-to-image collection contains approximately \textbf{326M image--text pairs}. It combines GPIC~\citep{chandrasegaran2026gpic}, Conceptual 12M (CC12M)~\citep{changpinyo2021cc12m}, CommonCatalog CC-BY and CommonCatalog CC-BY-SA~\citep{gokaslan2023commoncanvas}, Megalith~\citep{ollinmegalith}, and WIT~\citep{srinivasan2021wit}. Together, these collections provide broad coverage of visual concepts, web photographs, encyclopedic imagery, and visual content embedded in documents.

We supplement these sources with approximately \textbf{48M synthetic images} generated with Z-Image~\citep{team2025zimage}, bringing the text-to-image collection to approximately \textbf{374M examples} for mid-training.

\textbf{Data Annotation and Filtering.}
We annotate images using Qwen3.5-27B, Qwen3.5-397B-A17B, and Qwen3.5-4B~\citep{qwen2026qwen35}. We remove watermarked and low-quality images, and use Qwen3.5-397B-A17B to filter out captions that do not align with the image content.

\subsubsection{Data Across Training Stages}
The raster generator uses arbitrary-resolution mid-training with progressively increasing maximum image areas of $256^2$, $512^2$, and $1024^2$ pixels, allowing variable aspect ratios within each stage (Section~\ref{sec:multistage_training}). Reinforcement learning uses a prompt collection that is refreshed to emphasize harder semantic constraints and a broader distribution of resolutions and aspect ratios, as described in Section~\ref{sec:reinforcement_learning}.

\subsection{Structured Code Generation Data}
\label{sec:structured_coding_data}

The multimodal-coding mixture contains approximately \textbf{4.7M effective training examples}, including configured sampling repetitions. It covers visual reconstruction, text-conditioned program generation, editing, and code understanding across several artifact types.

Vector-graphics and visual-program examples support SVG generation, editing, and understanding~\citep{wang2025internsvg,zhao2025vincicoder,rodriguez2025starvector}. Web examples connect rendered pages to their source code and provide generation and editing supervision~\citep{yun2024web2code,sun2025januscoder}. Chart and visualization examples cover plotting code, structured chart representations, and program understanding~\citep{xu2025chartmoe,ni2025viscoder,ni2025viscoder2}.

The mixture also includes formula transcription~\citep{blecher2025latexocr}, table-to-\LaTeX{} conversion, and CAD-related supervision~\citep{niu2025cmecad}, supplemented by teacher-generated chain-of-thought examples and direct ground-truth anchors. Together, these task groups connect visual appearance and user intent to the explicit representations predicted by \codefamily{}; Section~\ref{sec:structured_code_training} describes their use in supervised adaptation.

\FloatBarrier

\section{Training}
\label{sec:train}

Training is organized around three components of the Pixel as Code framework. The \textbf{image tokenizer} learns a discrete visual representation and its decoder. The \textbf{raster image generator}, \rastermodel{}, learns to predict visual tokens, and the \textbf{structured code generator}, \codefamily{}, learns to produce explicit programs and markup from the multimodal-coding mixture in Section~\ref{sec:structured_coding_data}. We describe their training separately in Sections~\ref{sec:tokenizer_training}, \ref{sec:bitmap_training}, and~\ref{sec:structured_code_training}, following the component definitions in Section~\ref{sec:components}.

\subsection{Image Tokenizer Training}
\label{sec:tokenizer_training}

The image tokenizer learns a discrete visual representation and its rendering interface from the NVCLIP images described in Section~\ref{sec:tokenizer_data}, using the architecture in Table~\ref{tab:tokenizer_architecture}. Training spans 670K steps and combines pixel reconstruction, perceptual supervision, feature distillation, and adversarial learning. The codebook remains fixed, while the encoder is frozen after 480K steps. The decoder continues to learn to reconstruct images from the resulting discrete codes.

Visual compression determines both reconstruction fidelity and the sequence budget for generation. Cosmos~\citep{nvidia2025cosmos} develops visual tokenizers for world modeling, DC-AE~\citep{chen2024dcae} studies deep compression for high-resolution diffusion, and DC-AR~\citep{wu2025dcar} uses a deep-compression hybrid tokenizer for masked autoregressive generation. In our tokenizer, the $16\times16$ patch size represents a $256\times256$ image with 256 tokens, compared with 1,024 tokens for $8\times8$ patches. In our tokenizer study, this setting enables $4\times$ more samples per node. This trade-off connects the fidelity of the learned renderer to the sequence budget of the raster image generator.

\textbf{Reconstruction.}
We use an L1 reconstruction loss with weight 1.0 and an LPIPS perceptual loss~\citep{zhang2018lpips} with weight 0.1.

\textbf{Decoder feature distillation.}
We use DINOv2 ViT-L~\citep{oquab2023dinov2} as the distillation teacher. Student features are taken from the third decoder block, projected through an MLP, and aligned to the teacher's spatial grid using two-dimensional adaptive average pooling. Both student and teacher features are L2-normalized independently at each token. For the resulting $N$ aligned token pairs $(s_i,t_i)$, the distillation loss is the negative mean cosine similarity,
\begin{equation}
    \mathcal{L}_{\mathrm{distill}}
    = -\frac{1}{N}\sum_{i=1}^{N}\cos(s_i,t_i),
    \qquad \lambda_{\mathrm{distill}}=0.1.
    \label{eq:tokenizer_distillation_loss}
\end{equation}
This auxiliary supervision acts on the learned decoder and supports arbitrary image resolutions through spatial feature alignment.

\textbf{Adversarial training.}
An unconditional PatchGAN discriminator~\citep{isola2017pix2pix} operates on RGB images with three layers, a base width of 64, $4\times4$ kernels, and batch normalization~\citep{ioffe2015batchnorm}. It uses the \texttt{pad\_max} mode for arbitrary resolutions, with blur downsampling and ActNorm disabled. The GAN objective is softplus, with a configured discriminator weight of 0.5 and adaptive discriminator weighting disabled. Adversarial training starts at step 0. R1 regularization~\citep{mescheder2018gan} uses $\gamma=10$ and is evaluated once every 16 steps with lazy regularization~\citep{karras2020stylegan2}. On regularization steps, the R1 penalty is multiplied by $(\gamma/2)\times16=80$, giving a time-averaged coefficient of $\gamma/2=5$. Table~\ref{tab:tokenizer_training} collects these loss settings and the training configuration.

\textbf{Quantizer training.}
We keep the codebook fixed throughout training and use a straight-through estimator~\citep{bengio2013straightthrough,oord2017vqvae} to propagate gradients through discrete quantization. The training objective does not include a commitment loss.

\textbf{Variable-resolution token packing.}
Each image is represented as a sequence of $16\times16$ patches. We pack sequences of different lengths into a batch while preserving image boundaries, so images retain their individual spatial resolutions without padding every sequence to a common length. Replicate padding only aligns each image to the patch grid. Batches are formed under a per-rank pixel budget of approximately 2.62M; their image count therefore varies with resolution. Table~\ref{tab:tokenizer_training} summarizes the packing configuration.

\begingroup
    \small
    \setlength{\tabcolsep}{6pt}
    \renewcommand{\arraystretch}{1.10}
    \setlength{\LTleft}{0pt}
    \setlength{\LTright}{0pt plus 1fill}
    \setlength{\LTcapwidth}{\linewidth}
    \setlength{\LTpre}{8pt plus 2pt minus 2pt}
    \setlength{\LTpost}{8pt plus 2pt minus 2pt}
    \begin{longtable}{@{}>{\raggedright\arraybackslash}p{0.49\linewidth}>{\raggedright\arraybackslash}p{\dimexpr0.51\linewidth-12pt\relax}@{}}
        \caption{Image tokenizer training configuration through 670K steps. AdamW optimizes both the tokenizer autoencoder and the discriminator; Table~\ref{tab:tokenizer_lr_schedule} gives the effective learning rates.}
        \label{tab:tokenizer_training} \\
        \toprule
        Hyperparameter & Value \\
        \midrule
        \endfirsthead
        \multicolumn{2}{@{}l}{\textbf{Table \thetable{} (continued): tokenizer training configuration}} \\
        \toprule
        Hyperparameter & Value \\
        \midrule
        \endhead
        \midrule
        \multicolumn{2}{r@{}}{\emph{Continued on the next page}} \\
        \endfoot
        \bottomrule
        \endlastfoot
        \multicolumn{2}{@{}l}{\textbf{Reconstruction and distillation}} \\*
        Pixel reconstruction / weight & Mean absolute RGB error in $[-1,1]$ / 1.0 \\
        Perceptual loss / weight & LPIPS / 0.1 \\
        Distillation teacher / weight & DINOv2 ViT-L / 0.1 \\
        Distillation objective & Negative mean cosine similarity \\
        Distillation attachment & Decoder block 3, followed by an MLP \\
        Feature alignment & 2D adaptive average pooling to the teacher grid \\
        Feature normalization & Tokenwise L2 normalization on both sides \\
        Arbitrary-resolution distillation & Enabled \\
        Commitment loss & Not used \\
        Quantization gradients / codebook & Straight-through estimator / fixed \\
        \midrule
        \multicolumn{2}{@{}l}{\textbf{Adversarial training}} \\*
        Discriminator & Unconditional PatchGAN; 3 input channels \\
        Layers / base width / kernel size & 3 / 64 / $4\times4$ \\
        Normalization & Batch normalization; ActNorm disabled \\
        Blur downsampling & Disabled \\
        Arbitrary-resolution mode & \texttt{pad\_max} \\
        GAN objective / start step & Softplus / 0 \\
        Discriminator weight / adaptive weighting & 0.5 / disabled \\
        R1 $\gamma$ / frequency & 10 / every 16 steps (lazy regularization) \\
        R1 multiplier & 80 when evaluated; time-averaged coefficient 5 \\
        \midrule
        \multicolumn{2}{@{}l}{\textbf{Data and token packing}} \\*
        Training corpus & NVCLIP (Section~\ref{sec:tokenizer_data}) \\
        Curated image pool & Approximately 200M images \\
        Images used in training & Approximately 153M \\
        Clustering, deduplication, and filtering & SigLIP~2, DINOv2, and SSCD \\
        Batch construction & Variable-resolution token packing \\
        \texttt{min\_pixels} / \texttt{max\_pixels} & 256 / 1,638,400 \\
        Size divisor / padding mode & 16 / replicate \\
        Pixel budget per rank per batch & 2,621,440 \\
        \midrule
        \multicolumn{2}{@{}l}{\textbf{Optimization}} \\*
        Autoencoder / discriminator optimizer & AdamW / AdamW \\
        Base learning rate & $4\times10^{-7}$ \\
        World size / nominal batch size & 32 / 8 \\
        Scaled base learning rate & $1.024\times10^{-4}$ \\
        Weight decay & $10^{-4}$ \\
        AdamW $\beta_1,\beta_2$ / $\epsilon$ & $(0.9,0.99)$ / $10^{-8}$ \\
        Precision / automatic mixed precision & BF16 / enabled \\
        Gradient checkpointing & Enabled \\
        Encoder freezing & After 480K steps \\
        Training duration & 670K steps \\
    \end{longtable}
\endgroup

\textbf{Optimization and learning-rate schedule.}
Both the tokenizer autoencoder and the discriminator use AdamW~\citep{loshchilov2019adamw} with weight decay $10^{-4}$, $\beta_1=0.9$, $\beta_2=0.99$, and $\epsilon=10^{-8}$. Training uses BF16 automatic mixed precision and gradient checkpointing~\citep{chen2016checkpointing}. The configuration's base learning rate is $4\times10^{-7}$, scaled by a world size of 32 and a nominal batch size of 8:
\begin{equation}
    \eta(t)
    = \eta_{\mathrm{base}}\,W\,B_{\mathrm{nom}}\,m(t)
    = (4\times10^{-7})\times32\times8\times m(t)
    = 1.024\times10^{-4}\,m(t),
    \label{eq:tokenizer_lr}
\end{equation}
where $m(t)$ is the stage multiplier in Table~\ref{tab:tokenizer_lr_schedule}. The nominal batch size is a learning-rate scaling parameter, distinct from the number of images in each packed batch. The multiplier decreases from 1.0 to 0.6 at 300K steps and to 0.36 at 480K steps. The last stage freezes the encoder and uses a learning rate of $3.6864\times10^{-5}$ through 670K steps.

\begin{table}[!htbp]
    \centering
    \caption{Tokenizer learning-rate stages. The multiplier scales the base rate after world-size and nominal-batch-size scaling. The codebook is frozen throughout training.}
    \label{tab:tokenizer_lr_schedule}
    \begingroup
    \small
    \setlength{\tabcolsep}{6pt}
    \renewcommand{\arraystretch}{1.10}
    \begin{tabular*}{\linewidth}{@{\extracolsep{\fill}}lccc@{}}
        \toprule
        Training steps & Multiplier $m(t)$ & Learning rate $\eta(t)$ & Encoder \\
        \midrule
        0--300K & 1.0 & $1.024\times10^{-4}$ & Trainable \\
        300K--480K & 0.6 & $6.144\times10^{-5}$ & Trainable \\
        480K--670K & 0.36 & $3.6864\times10^{-5}$ & Frozen \\
        \bottomrule
    \end{tabular*}
    \endgroup
\end{table}

Section~\ref{sec:tokenizer_evaluation} evaluates reconstruction quality, single-image latency, and encoding/decoding throughput. Appendix~\ref{app:tokenizer_reconstruction} provides domain-level reconstruction results and an ImageNet any-resolution comparison with an encoding-scale ablation.

\subsection{Raster Image Generator Training}
\label{sec:bitmap_training}

The raster generator adapts a language-pretrained backbone through multistage autoregressive mid-training, followed by on-policy reinforcement learning. Mid-training teaches visual-token prediction with progressively increasing pixel budgets; reinforcement learning optimizes the rendered outputs for semantic alignment and perceptual quality. Table~\ref{tab:nemotron_lab_image_training} summarizes the stage-wise configuration. Sections~\ref{sec:multistage_training} and~\ref{sec:reinforcement_learning} describe these two parts of training.

\subsubsection{Multi-Stage Mid-Training}
\label{sec:multistage_training}

We continue training the language-pretrained backbone to autoregressively predict target-image tokens from text and optional source or reference images. All three stages use 90\% T2I and 10\% TI2I data (Section~\ref{sec:raster_data}), covering text-to-image generation, editing, and reference-conditioned synthesis. The maximum image area increases through $256^2$, $512^2$, and $1024^2$ pixels, with variable aspect ratios; the corresponding training durations are 35K, 90K, and 70K steps. Table~\ref{tab:nemotron_lab_image_training} lists the optimization settings.

\begin{table}[!htbp]
\centering
\renewcommand{\arraystretch}{1.15}
\setlength{\tabcolsep}{7pt}

\resizebox{\textwidth}{!}{%
\begin{tabular}{@{}lcccc@{}}
\toprule
\textbf{Configuration}
&
\textbf{Mid-Training Stage 1}
&
\textbf{Mid-Training Stage 2}
&
\textbf{Mid-Training Stage 3}
&
\textbf{Reinforcement Learning}
\\
Maximum image pixels
&
\textbf{$256^2$}
&
\textbf{$512^2$}
&
\textbf{$1024^2$}
&
--
\\
\midrule

\multicolumn{5}{@{}l}{\textbf{Optimization}}
\\
\addlinespace[2pt]

LR scheduler
& Constant
& Constant
& Constant
& Constant
\\

Learning rate (Muon)
& $1 \times 10^{-3}$
& $1 \times 10^{-3}$
& $1 \times 10^{-3}$
& --
\\

Learning rate (Adam)
& $1 \times 10^{-4}$
& $1 \times 10^{-4}$
& $1 \times 10^{-4}$
& $2 \times 10^{-6}$
\\

Optimizer (Muon)
& \multicolumn{3}{c}{
$\mu=0.95$, Polar Express, spectral-norm LR adjustment
}
& --
\\

Optimizer (Adam)
& \multicolumn{3}{c}{
$\beta_1=0.9,\ \beta_2=0.95,\ \epsilon=10^{-8}$
}
& $\beta_1=0.9,\ \beta_2=0.98,\ \epsilon=10^{-8}$
\\

Weight decay
& 0.01 & 0.01 & 0.01 & 0.0
\\

Gradient norm clip
& 1.0 & 1.0 & 1.0 & 1.0
\\

EMA ratio
& 0.9999
& 0.9999
& 0.9999
& --
\\

Training steps
& 35K
& 90K
& 70K
& 200 rollout iterations
\\

Warmup steps
& 500
& 500
& 500
& 0
\\

Sequence length
& 81920
& 81920
& 81920
& 8192
\\

Prompts per rollout
& -- & -- & -- & 48
\\

Candidates per prompt
& -- & -- & -- & 32
\\

Trajectories per update
& -- & -- & -- & 768
\\

\midrule
\multicolumn{5}{@{}l}{\textbf{Training data mixture}}
\\
\addlinespace[2pt]

 T2I data
& 0.9 & 0.9 & 0.9 & 1.0
\\

TI2I data
& 0.1 & 0.1 & 0.1 & 0.0
\\

\bottomrule
\end{tabular}%
}

\caption{
\textbf{Raster image generator training configurations.}
For Nemotron-Labs-Image-30B-A3B, we report three arbitrary-resolution mid-training
stages with increasing maximum image pixel counts and the mixed-resolution
reinforcement-learning stage. Aspect ratios vary within each mid-training stage.
T2I denotes text-to-image generation; TI2I denotes generation conditioned on both
text and source or reference images.
The optimizer configuration uses Muon~\citep{jordan2024muon} with
Polar Express~\citep{amsel2026polarexpress} and Adam~\citep{kingma2015adam}.
}

\label{tab:nemotron_lab_image_training}
\end{table}

\subsubsection{Reinforcement Learning}
\label{sec:reinforcement_learning}

Reinforcement learning has been studied for both discrete autoregressive image generation, as in X-Omni~\citep{geng2025xomni}, and diffusion models. Data-regularized RL~\citep{ye2025ddrl} combines reward optimization with diffusion-based data regularization. We describe our discrete-token policy and its reward-driven training procedure below.

\textbf{From supervised learning to preference optimization.}
Building on multistage mid-training, we introduce reinforcement learning once high-quality, prompt-faithful candidates consistently emerge under repeated sampling. This best-of-$N$ headroom provides a strong on-policy learning signal: repeated rollouts explore new compositions of the concepts and visual patterns acquired during mid-training, while reward-guided updates consolidate successful discoveries into reliable single-sample behavior. We adopt a two-stage RL scheme: broad-coverage optimization followed by learnability-driven data refresh to sustain informative learning signals as the policy improves.

\textbf{Stage I: broad-coverage RL.}
The first RL stage draws from a broad prompt distribution spanning object-centric generation, attribute binding, counting, spatial relations, interactions, complex scene composition, stylistic control, and text rendering. This diverse coverage encourages the policy to explore and combine its acquired visual knowledge across a wide range of generation requirements. For every prompt, the current policy generates a group of candidates and receives pointwise alignment and image-quality scores. Groups with sufficient reward variation yield informative relative advantages, providing a discriminative learning signal for group-relative policy optimization.

\textbf{Stage II: learnability-driven data refresh.}
As Stage~I proceeds, easy prompts increasingly produce uniformly strong candidates and their within-group alignment-score range contracts. We retain for GRPO only prompt groups with sufficient within-group reward variation; groups lacking a discriminative relative ordering are discarded before the policy update. The growing prevalence of low-range groups signals that the current prompt distribution no longer matches the policy's evolving learning frontier. We therefore refresh the prompt distribution while preserving the reward definition and optimizer state. The refreshed mixture retains structured compositional prompts, redistributes general prompts across multiple resolutions and aspect ratios according to the empirical prior of our mid-training corpus, and introduces longer, more challenging prompts with denser semantic and compositional constraints. Pairing these difficult cases with diverse image geometries restores informative reward variation without changing the optimization objective. The resulting loop---\emph{roll out, diagnose, recurate, and improve}---uses the model's evolving behavior to identify which examples can still induce informative on-policy comparisons. Our two-stage RL training instantiates one iteration of this broader self-improvement process. Figure~\ref{fig:training_progression} uses representative examples to illustrate the progression from mid-training to RL alignment and the full agentic workflow. Across these examples, RL improves legible text rendering, single-object structure and surface detail, and compliance with counting and spatial relations. The complete agentic workflow further improves exact text and count fidelity, coverage of densely specified multi-object scenes, fine-grained actions and material detail, and consistency under explicit artistic styles.

\newcommand{\trainingcomparisonheader}{%
    \begin{tabularx}{\linewidth}{@{}*{3}{>{\centering\arraybackslash}X}@{}}
        \textbf{Mid-trained generator} & \textbf{RL-aligned generator} & \textbf{Agentic workflow} \\
    \end{tabularx}\par\vspace{3pt}%
}
\newcommand{\trainingcomparisonrow}[4]{%
    {\fontsize{8}{9.4}\selectfont\raggedright\textbf{Prompt.} ``#1''\par}\vspace{3pt}%
    \noindent\makebox[0.325\linewidth][c]{\includegraphics[page=#2,pagebox=cropbox,width=0.300\linewidth]{figures/training_progression_panels.pdf}}\hfill%
    \makebox[0.325\linewidth][c]{\includegraphics[page=#3,pagebox=cropbox,width=0.300\linewidth]{figures/training_progression_panels.pdf}}\hfill%
    \makebox[0.325\linewidth][c]{\includegraphics[page=#4,pagebox=cropbox,width=0.300\linewidth]{figures/training_progression_panels.pdf}}\par\vspace{5pt}%
}

\begin{figure}[p]
    \centering
    \begingroup\sffamily\small\setstretch{1.0}
    \trainingcomparisonheader
    \trainingcomparisonrow{A warm lighting beautifully crafted sign in a charming neighborhood bookstore scene. The central wooden sign distinctly reads: ``Chapter One Books Discover Your Next Great Adventure!'' Directly beneath, in slightly smaller, welcoming text, appears ``Author signing tonight: Meet Emily Hartfield at 7pm!'' At the bottom, neatly handwritten script announces, ``Complimentary coffee and snacks inside''. Smaller adjoining signs set near displays feature engaging phrases such as ``New arrivals handpicked for you''. All textual elements on the signs are inviting, clearly readable, and thoughtfully designed to attract book lovers passing by.}{1}{2}{3}
    \trainingcomparisonrow{Pink lights illuminate a sleek motorcycle under smooth lighting. The scene is captured with incredible detail, emphasizing the intricate design and vivid colors of the bike. The setting is street-inspired, presenting a modern urban environment. The motorcycle glimmers with a dark white and light azure color scheme. Photorealistic elements bring forward an epic composition, revealing the high level of precision involved in the photography. Ray tracing reflections add depth and dynamic contrast, showcasing meticulous attention to shadows and highlights. Advanced techniques such as global illumination and volumetric lighting enhance the scene's realism, while chromatic aberration and diffraction gradation contribute subtle touches of artistic flair. The motorcycle and its immediate surroundings exhibit a rich interplay of light, focus, and texture, making it appear strikingly real.}{4}{5}{6}
    \trainingcomparisonrow{six flowers in front of a umbrella}{7}{8}{9}
    \endgroup
    \caption{Selected examples from the mid-trained generator, the RL-aligned policy, and the complete agentic creation system for the same user requests. These examples illustrate visual text rendering, surface detail, and counting. The columns compare complete configurations; the agentic workflow includes planning, external information, generation, and selection.}
    \label{fig:training_progression}
\end{figure}

\begin{figure}[p]
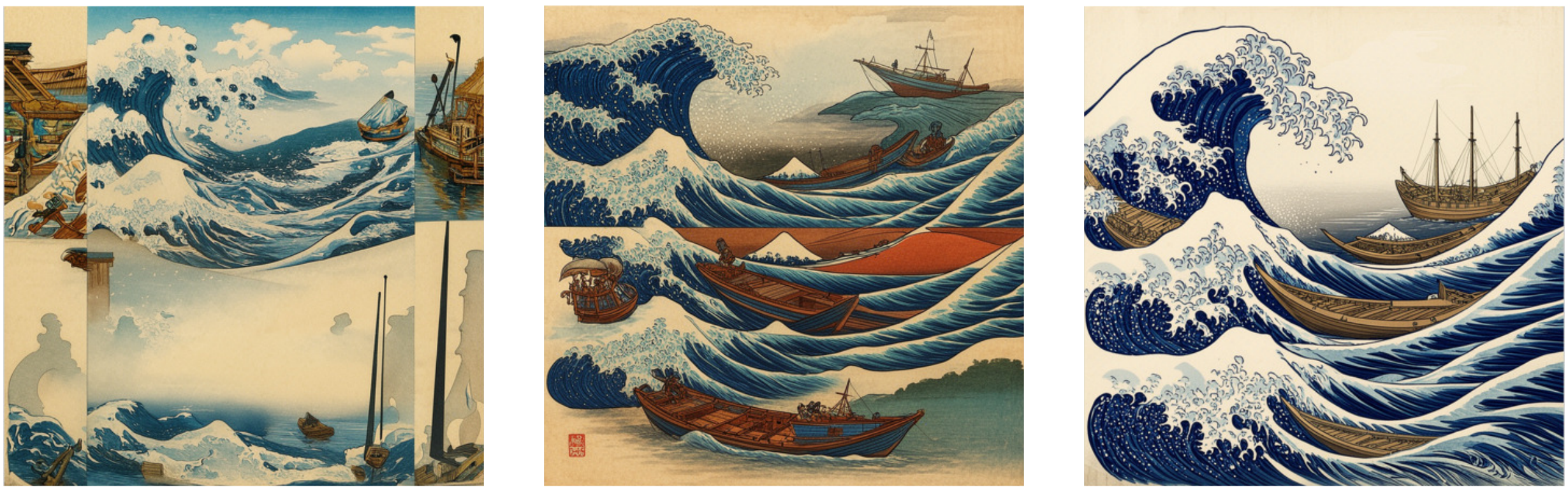

    \ContinuedFloat
    \centering
    \begingroup\sffamily\small\setstretch{1.0}
    \trainingcomparisonheader
    \trainingcomparisonrow{German culture theme, close-up of a wooden toy stall at a Christmas market, log-built booth decorated with pine garlands and string lights, tabletop displaying nutcrackers, wooden rocking horses, spinning wooden music boxes, handcrafted puppets, and small snowman figurines, handmade ornaments hanging from the beams, visible wood grain and carving marks, warm yellow bulbs illuminating the stall with a soft glow, background winter-clad crowd blurred into gentle color shapes, nighttime ambience, realistic photography look with intricate details}{10}{11}{12}
    \trainingcomparisonrow{An old craftsman sits at a workbench, gently blowing sawdust from a finished wood carving, eyes fixed on the details, gaze flickering, lips subtly tightened}{13}{14}{15}
    \trainingcomparisonrow{A scene of waves and fishing boats completely rendered in the ukiyo-e style, with lines, colors, and compositions strictly adhering to Hokusai's aesthetic norms.}{16}{17}{18}
    \endgroup
    \caption[]{Training progression (continued). These examples illustrate dense scene composition, fine-grained actions, and adherence to a requested artistic style. Prompts and generated outputs are retained from the supplied comparison images.}
    \label{fig:training_progression_more}
\end{figure}

We next detail the optimization procedure used to obtain the RL-aligned policy. We first formulate autoregressive image generation as a discrete-token policy, then define the pointwise reward and group-relative objective, and finally describe the rollout filtering and distributed execution used in training. We optimize the policy with Group Relative Policy Optimization (GRPO)~\citep{shao2024deepseekmath}. Because trajectories are regenerated from the continually updated policy rather than drawn from a fixed preference set, successive rollout iterations explore previously unrealized combinations of the model's acquired knowledge, while group-relative rewards determine which discoveries are reinforced by subsequent policies.

\textbf{Discrete-token policy.}
Let $\mathbf{x}$ denote the prompt tokens and $\mathbf{z}=(z_1,\ldots,z_T)$ the discrete visual-token sequence generated by $\pi_\theta$. The tokenizer decoder renders $\hat I=D(\mathbf{z})$, instantiating Equation~\ref{eq:code_render_interface}. The autoregressive policy factorizes as
\begin{equation}
    \pi_\theta(\mathbf{z}\mid\mathbf{x})
    = \prod_{t=1}^{T}
    \pi_\theta\!\left(z_t\mid\mathbf{x},\mathbf{z}_{<t}\right).
    \label{eq:ar_image_policy}
\end{equation}
This permits direct application of the token-level GRPO objective used for language models. Prompt and reference-image tokens provide context but are excluded from the action loss; the action mask selects only generated response tokens, including visual-token IDs.

\textbf{Composite pointwise reward.}
Each candidate code is rendered by the tokenizer decoder at its requested resolution and independently evaluated against the original prompt by a pointwise VLM judge. The judge scores six alignment dimensions---including counting, relations, actions, and text rendering---together with technical image quality. After aggregation and normalization to $[0,1]$, we define
\begin{equation}
    R(\mathbf{x},\mathbf{z}) = 0.7\,\bar r_{\mathrm{align}}(\mathbf{x},\mathbf{z})
    + 0.3\,\bar r_{\mathrm{quality}}(\mathbf{x},\mathbf{z}),
    \label{eq:composite_reward}
\end{equation}
One judge call produces both components under the same visual evidence and rubric. The complete instruction, structured output, and aggregation rule are given in Appendix~\ref{app:reward_judge}.

\textbf{Group-relative credit assignment.}
For each prompt, the rollout policy samples $\mathcal{G}(\mathbf{x})=\{\mathbf{z}^{(i)}\}_{i=1}^{G}$ and normalizes rewards within this group:
\begin{equation}
    A_i = \frac{R(\mathbf{x},\mathbf{z}^{(i)})-\mu_{\mathcal{G}}}
    {\sigma_{\mathcal{G}}+\epsilon},
    \qquad \mathbf{z}^{(i)}\in\mathcal{G}(\mathbf{x}),
    \label{eq:grpo_advantage}
\end{equation}
where $\mu_{\mathcal{G}}$ and $\sigma_{\mathcal{G}}$ are the group statistics. This removes prompt-specific score offsets, and the resulting sequence-level advantage $A_i$ is shared by all generated tokens in trajectory $i$. Let $m_{i,t}$ be the action mask and $\rho_{i,t}(\theta)$ the likelihood ratio between the current and rollout policies for token $z_t^{(i)}$. The PPO-style clipped objective~\citep{schulman2017ppo} is
\begin{equation}
\begin{split}
    \mathcal{L}_{\mathrm{GRPO}}(\theta)
    = -\mathbb{E}_{\substack{\mathbf{x}\sim\mathcal{D},\\
    \mathbf{z}^{(i)}\sim\pi_{\theta_{\mathrm{old}}}}}
    \Bigg[\frac{1}{\sum_{i,t}m_{i,t}}
    \sum_{i,t}m_{i,t}\min\Big(&\rho_{i,t}(\theta)A_i,\\
    &\operatorname{clip}(\rho_{i,t}(\theta),
    1-\epsilon_{\mathrm{lo}},1+\epsilon_{\mathrm{hi}})A_i\Big)\Bigg].
\end{split}
    \label{eq:ar_grpo_objective}
\end{equation}

\begin{figure}[!htbp]
    \centering
    \includegraphics[width=0.82\linewidth]{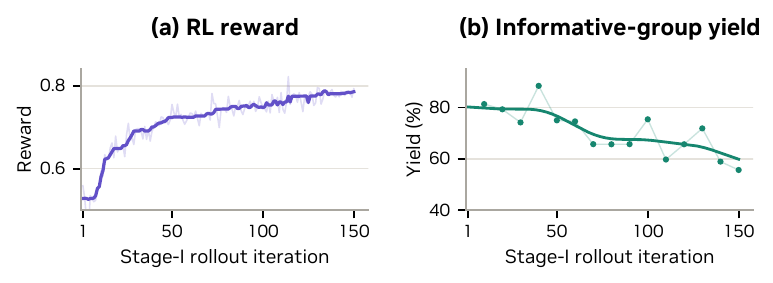}
    \caption{Stage~I saturation diagnostics. (a) RL reward over valid images; the light and dark curves show per-iteration values and a short-window median. (b) Fraction of fully valid groups satisfying Eq.~\ref{eq:alignment_range_filter}, audited every ten iterations; markers and line show the observations and their trend.}
    \label{fig:rl_saturation_diagnostics}
\end{figure}

We use $\epsilon_{\mathrm{lo}}=0.20$, $\epsilon_{\mathrm{hi}}=0.28$, a learning rate of $2\times10^{-6}$, and no explicit KL penalty; Table~\ref{tab:nemotron_lab_image_training} summarizes the remaining settings.

\textbf{Rollout and update schedule.}
Each rollout iteration samples $G=32$ candidates at temperature 1.0 for each of $B=48$ prompts, yielding 1,536 trajectories. Keeping prompt groups intact, we partition them into two disjoint batches of 768 trajectories and take one optimizer step per batch; every trajectory is therefore used once. Consequently, the reported 200 rollout iterations correspond to 400 optimizer steps. In pilot experiments, smaller groups produced noisy baselines and often failed to improve reward, whereas $G=32$ more reliably exposed meaningful within-prompt reward variation.

\textbf{Reward-range group filtering.}
Before optimization, we retain a prompt group only when its alignment-score range on the original 0--10 scale satisfies
\begin{equation}
    \max_{i\in\mathcal{G}} r_{\mathrm{align},i}
    - \min_{i\in\mathcal{G}} r_{\mathrm{align},i} \geq 2.
    \label{eq:alignment_range_filter}
\end{equation}
This removes both saturated groups whose candidates are uniformly strong and failed groups with no useful relative ordering. Rejection does not shrink the update batch: the sampler draws a new prompt and repeats generation and reward evaluation until all $B=48$ accepted groups have been collected. The accepted-group yield therefore measures how much of the current prompt distribution still provides informative within-group comparisons.

Figure~\ref{fig:rl_saturation_diagnostics} isolates the dynamics of Stage~I, before the prompt and resolution distribution is changed. Panel~(a) reports the mean RL reward over successfully decoded images at every rollout iteration. Panel~(b) audits the fraction of complete prompt groups for which all 32 candidates decode successfully and the alignment-score range satisfies Eq.~\ref{eq:alignment_range_filter}. As optimization proceeds, reward rises and gradually approaches a plateau, whereas the yield of informative groups decreases. This divergence indicates that the policy is improving on the current distribution while an increasing fraction of prompts no longer produces sufficiently discriminative within-group comparisons. We use this saturation signal to motivate the Stage~II data refresh described above, replacing increasingly exhausted prompts with harder semantic constraints and a broader resolution distribution.

\textbf{Rollout-policy consistency.}
Rollouts and gradient computation use separate inference and training implementations. To suppress bias from small numerical discrepancies over long visual-token sequences, we compute the geometric-mean likelihood ratio
\begin{equation}
    g_i=\exp\!\left(
    \frac{1}{\sum_t m_{i,t}}
    \sum_t m_{i,t}
    \log\frac{\pi_{\theta_{\mathrm{old}}}(z_t^{(i)}\mid\mathbf{x},\mathbf{z}_{<t}^{(i)})}
    {\pi_{\mathrm{roll}}(z_t^{(i)}\mid\mathbf{x},\mathbf{z}_{<t}^{(i)})}
    \right).
    \label{eq:rollout_actor_consistency}
\end{equation}
We retain sequences with $g_i\in[0.95,1.05]$ and apply token-level importance weights to the surviving losses.

\textbf{Scalable asynchronous execution.}
Following large-scale image RL systems~\citep{xu2026qwenimage2rl}, we decouple rollout generation, reward evaluation, and policy optimization across 256 GPUs. Disaggregated actors, inference engines, and decoding workers overlap generation with asynchronous VLM scoring. Failed scorer requests are retried; unrecoverable groups are discarded and backfilled rather than assigned artificial zero rewards. We checkpoint every 25 rollout iterations and retain complete rollout records for auditing.

\FloatBarrier
\subsection{Structured Code Generator Training}
\label{sec:structured_code_training}



We train two variants of \codefamily{}. \textbf{\codemodel{}} is initialized from the \codebasemodel{} GA model~\citep{nvidia2026nemotron3omni}\footnote{Official Hugging Face model: \url{https://huggingface.co/nvidia/Nemotron-3-Nano-Omni-30B-A3B-Reasoning-BF16}.}. \textbf{Nemotron-Labs-VisionCode-120B-A12B} is trained from \supervlmodel{}~\citep{nvidia2026nemotron35}. Both variants use the same training data and training procedure, including the supervised fine-tuning mixture and task-specific reinforcement learning described below.

We apply \textbf{multimodal-coding supervised fine-tuning} on approximately 4.7M effective training examples (Section~\ref{sec:structured_coding_data}). The mixture covers visual reconstruction, instruction-conditioned program generation, editing, and code understanding. For generation tasks, the targets are explicit programs and markup, providing supervision for the structured generation interface $G_r$ in Equation~\ref{eq:code_render_interface}.

Following supervised fine-tuning, we further optimize both models with Group Relative Policy Optimization~\citep{shao2024deepseekmath} using task-specific rewards. We build a custom sandbox that executes Python programs, renders SVG and HTML/CSS, and compiles LaTeX for reward evaluation. This infrastructure connects generated code to its executable or rendered outputs, providing feedback on both execution validity and task-specific quality. The reward design follows the requirements of each structured-code task. Execution checks assess whether generated programs run successfully, markup renders, edits can be parsed, and documents compile. Quality scores measure visual fidelity for reconstruction, text-image alignment and aesthetics for text-to-SVG generation, and edit accuracy for code editing. These outcome-based signals complement supervised fine-tuning by rewarding functional correctness and the quality of the resulting artifacts.

We evaluate both trained variants against their respective base models using execution, compilation, rendering, and task-specific quality metrics (Section~\ref{sec:mm_coding_benchmarks}).

\section{Experiments}
\label{sec:experiment}

Our evaluation follows the framework's representation--renderer interface. We first measure image tokenizer reconstruction quality and efficiency, then evaluate raster image generation and its agent-assisted configurations by visual capability, and finally assess structured code generation through execution, rendering, and task-specific quality metrics.

\subsection{Image Tokenizer Quality and Efficiency}
\label{sec:tokenizer_evaluation}
\label{sec:tokenizer_efficiency}

\textbf{Reconstruction quality.}
We evaluate the learned raster renderer on approximately 178K images spanning natural photographs, documents, charts, UI screenshots, faces, and generated or edited images. Appendix~\ref{app:tokenizer_evaluation_sources} lists the 35 source datasets by visual domain. Table~\ref{tab:tokenizer_reconstruction} compares \tokenizermodel{} with the Emu3.5 IBQ tokenizer~\citep{cui2025emu35}. We report PSNR, SSIM~\citep{wang2004ssim}, LPIPS~\citep{zhang2018lpips}, and GMSD~\citep{xue2013gmsd} for reconstruction fidelity, together with FID~\citep{heusel2017fid} and feature-distribution distance in DINOv2 space~\citep{oquab2023dinov2}. MUSIQ~\citep{ke2021musiq} and CLIP-IQA~\citep{wang2023clipiqa} assess reconstructed images without a reference.

\begin{table}[!htbp]
  \centering
  \caption{Overall tokenizer reconstruction quality on approximately 178K images across six visual domains. MUSIQ and CLIP-IQA score reconstructed images without a reference. The Laplacian-variance ratio has a target value of 1. Bold marks the better value for each metric.}
  \label{tab:tokenizer_reconstruction}
  \begingroup
  \small
  \setlength{\tabcolsep}{4pt}
  \renewcommand{\arraystretch}{1.12}
  \begin{tabular*}{\linewidth}{@{\extracolsep{\fill}}lrr@{}}
    \toprule
    Metric & Emu3.5 IBQ & \tokenizermodel{} (ours) \\
    \midrule
    \multicolumn{3}{l}{\textit{Reconstruction fidelity}} \\
    PSNR (dB) $\uparrow$ & 26.08 & \textbf{29.57} \\
    SSIM $\uparrow$ & 0.781 & \textbf{0.834} \\
    LPIPS $\downarrow$ & 0.0855 & \textbf{0.0713} \\
    GMSD $\downarrow$ & 0.0994 & \textbf{0.0675} \\
    \midrule
    \multicolumn{3}{l}{\textit{Feature-distribution distance}} \\
    FID $\downarrow$ & 36.36 & \textbf{35.06} \\
    FD-DINOv2 $\downarrow$ & 7.49 & \textbf{5.25} \\
    \midrule
    \multicolumn{3}{l}{\textit{Image quality and sharpness}} \\
    MUSIQ $\uparrow$ & \textbf{65.84} & 65.78 \\
    CLIP-IQA $\uparrow$ & \textbf{0.593} & 0.583 \\
    Lap-var ratio $\to 1$ & 0.892 & \textbf{0.982} \\
    \bottomrule
  \end{tabular*}
  \endgroup
\end{table}

On the six-domain evaluation, \tokenizermodel{} achieves 29.57\,dB PSNR versus 26.08\,dB for Emu3.5 IBQ, a 3.49\,dB improvement (Table~\ref{tab:tokenizer_reconstruction}). SSIM increases by 0.053, while LPIPS decreases by approximately 17\% and GMSD by 32\%. All six domains improve on these four fidelity metrics, with PSNR gains of 4.63--5.35\,dB on documents, charts, and UI screenshots. This is particularly relevant to a renderer used within visual artifacts containing text and structured layouts. Overall FID and FD-DINOv2 also decrease, although these gains do not hold in every domain. MUSIQ and CLIP-IQA are slightly lower overall. Appendix~\ref{app:tokenizer_reconstruction} reports the domain-level results and an ImageNet any-resolution comparison with an encoding-scale ablation.

\begin{figure}[!htbp]
    \centering
    \begingroup\sffamily
    \pgfmathsetlengthmacro{\tokpanelwidth}{0.244\linewidth}
    \pgfmathsetlengthmacro{\tokcontextwidth}{0.205\linewidth}
    \pgfmathsetlengthmacro{\tokpixel}{\tokpanelwidth/256}

    \small
    \begin{tabular}{@{}c@{\hspace{0.02\linewidth}}c@{\hspace{0.02\linewidth}}c@{\hspace{0.02\linewidth}}c@{}}
        \textbf{Input image} & \textbf{Original crop} & \textbf{Emu3.5 IBQ} & \parbox[c]{\tokpanelwidth}{\centering\small\bfseries \tokenizermodel{} (ours)} \\[4pt]
        \begin{tikzpicture}[baseline=(current bounding box.center)]
            \node[anchor=south west, inner sep=0] (context) at (0,0)
                {\includegraphics[width=\tokcontextwidth,height=\tokpanelwidth,keepaspectratio]{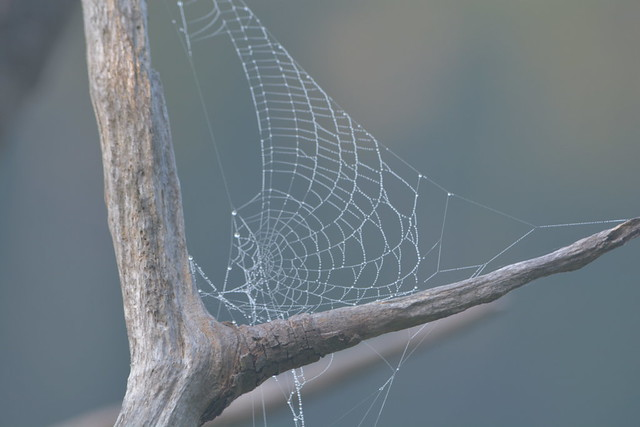}};
            \begin{scope}[x={(context.south east)},y={(context.north west)}]
                \draw[red!85!black,line width=0.7pt]
                    ({160/640},{1-30/427}) rectangle
                    ({(160+256)/640},{1-(30+256)/427});
            \end{scope}
        \end{tikzpicture} &
        \pgfmathsetlengthmacro{\tokimagewidth}{640*\tokpixel}%
        \begin{tikzpicture}[x=\tokpixel,y=\tokpixel,baseline=(current bounding box.center)]
            \path[use as bounding box] (0,0) rectangle (256,256);
            \begin{scope}[overlay]
                \clip (0,0) rectangle (256,256);
                \node[anchor=north west,inner sep=0] at (-160,{256+30})
                    {\includegraphics[width=\tokimagewidth]{figures/tokenizer_qualitative/original/01_s01_0021.jpg}};
            \end{scope}
        \end{tikzpicture} &
        \pgfmathsetlengthmacro{\tokimagewidth}{640*\tokpixel}%
        \begin{tikzpicture}[x=\tokpixel,y=\tokpixel,baseline=(current bounding box.center)]
            \path[use as bounding box] (0,0) rectangle (256,256);
            \begin{scope}[overlay]
                \clip (0,0) rectangle (256,256);
                \node[anchor=north west,inner sep=0] at (-160,{256+30})
                    {\includegraphics[width=\tokimagewidth]{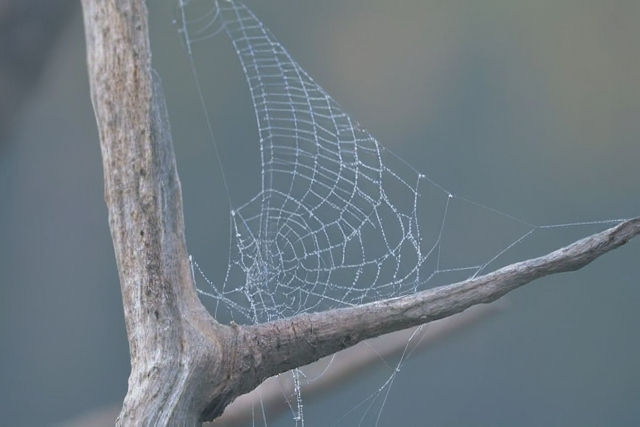}};
            \end{scope}
        \end{tikzpicture} &
        \pgfmathsetlengthmacro{\tokimagewidth}{640*\tokpixel}%
        \begin{tikzpicture}[x=\tokpixel,y=\tokpixel,baseline=(current bounding box.center)]
            \path[use as bounding box] (0,0) rectangle (256,256);
            \begin{scope}[overlay]
                \clip (0,0) rectangle (256,256);
                \node[anchor=north west,inner sep=0] at (-160,{256+30})
                    {\includegraphics[width=\tokimagewidth]{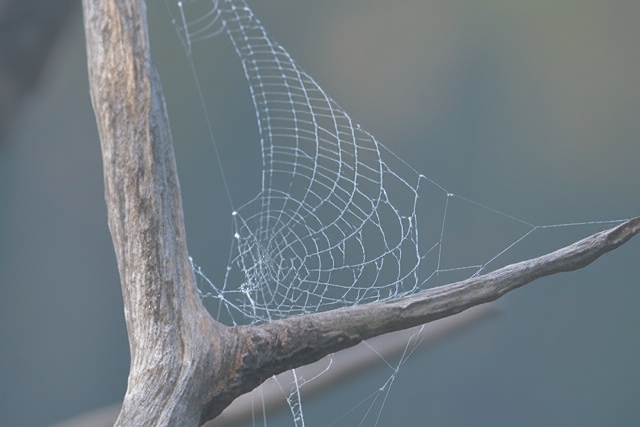}};
            \end{scope}
        \end{tikzpicture} \\[5pt]
        \begin{tikzpicture}[baseline=(current bounding box.center)]
            \node[anchor=south west, inner sep=0] (context) at (0,0)
                {\includegraphics[width=\tokcontextwidth,height=\tokpanelwidth,keepaspectratio]{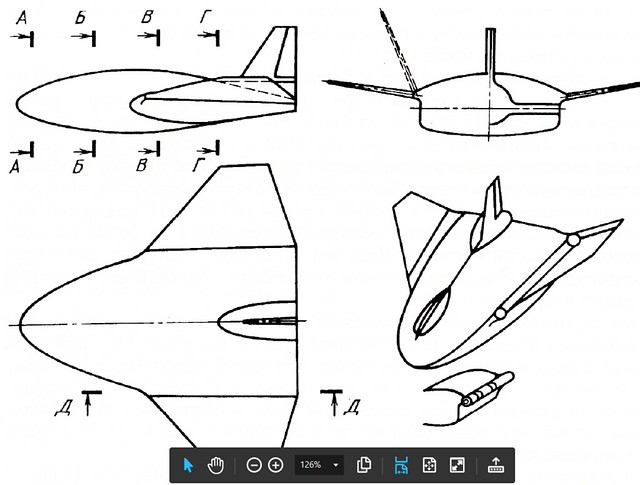}};
            \begin{scope}[x={(context.south east)},y={(context.north west)}]
                \draw[red!85!black,line width=0.7pt]
                    ({0/640},{1-120/485}) rectangle
                    ({(0+256)/640},{1-(120+256)/485});
            \end{scope}
        \end{tikzpicture} &
        \pgfmathsetlengthmacro{\tokimagewidth}{640*\tokpixel}%
        \begin{tikzpicture}[x=\tokpixel,y=\tokpixel,baseline=(current bounding box.center)]
            \path[use as bounding box] (0,0) rectangle (256,256);
            \begin{scope}[overlay]
                \clip (0,0) rectangle (256,256);
                \node[anchor=north west,inner sep=0] at (-0,{256+120})
                    {\includegraphics[width=\tokimagewidth]{figures/tokenizer_qualitative/original/02_s01_0028.jpg}};
            \end{scope}
        \end{tikzpicture} &
        \pgfmathsetlengthmacro{\tokimagewidth}{640*\tokpixel}%
        \begin{tikzpicture}[x=\tokpixel,y=\tokpixel,baseline=(current bounding box.center)]
            \path[use as bounding box] (0,0) rectangle (256,256);
            \begin{scope}[overlay]
                \clip (0,0) rectangle (256,256);
                \node[anchor=north west,inner sep=0] at (-0,{256+120})
                    {\includegraphics[width=\tokimagewidth]{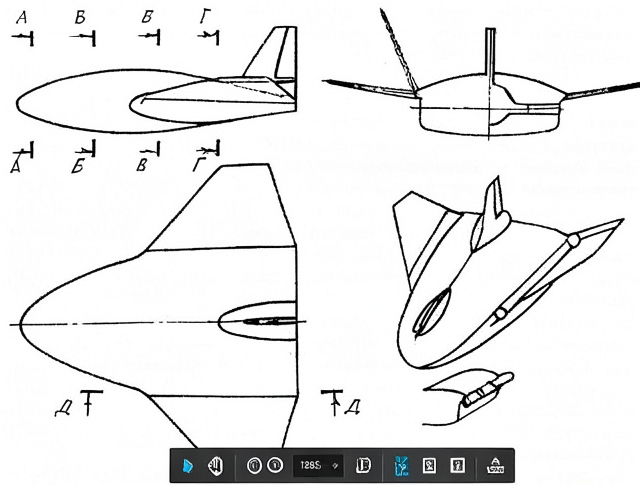}};
            \end{scope}
        \end{tikzpicture} &
        \pgfmathsetlengthmacro{\tokimagewidth}{640*\tokpixel}%
        \begin{tikzpicture}[x=\tokpixel,y=\tokpixel,baseline=(current bounding box.center)]
            \path[use as bounding box] (0,0) rectangle (256,256);
            \begin{scope}[overlay]
                \clip (0,0) rectangle (256,256);
                \node[anchor=north west,inner sep=0] at (-0,{256+120})
                    {\includegraphics[width=\tokimagewidth]{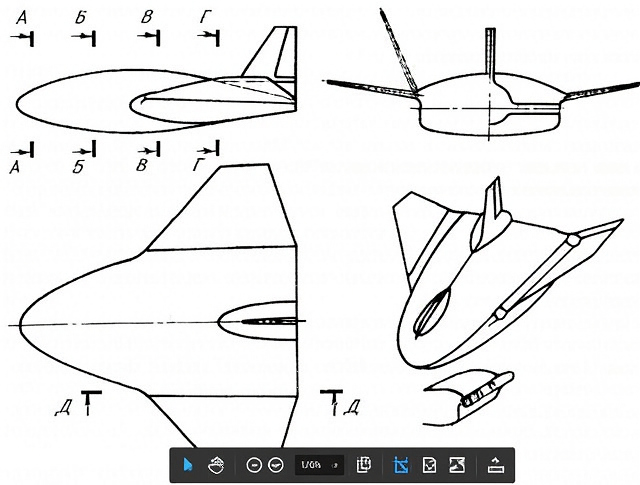}};
            \end{scope}
        \end{tikzpicture} \\[5pt]
        \begin{tikzpicture}[baseline=(current bounding box.center)]
            \node[anchor=south west, inner sep=0] (context) at (0,0)
                {\includegraphics[width=\tokcontextwidth,height=\tokpanelwidth,keepaspectratio]{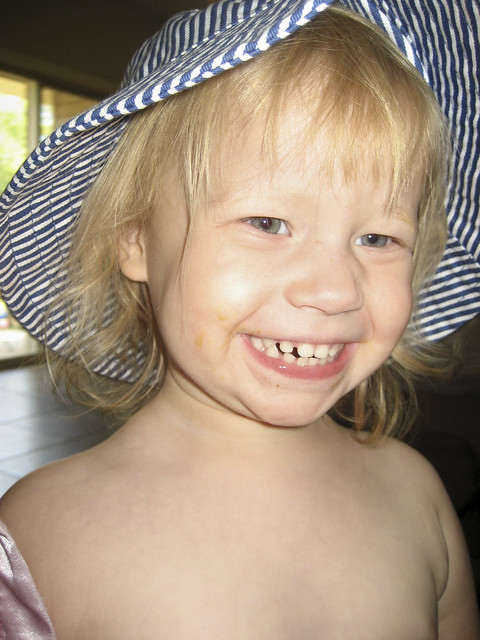}};
            \begin{scope}[x={(context.south east)},y={(context.north west)}]
                \draw[red!85!black,line width=0.7pt]
                    ({40/480},{1-20/640}) rectangle
                    ({(40+256)/480},{1-(20+256)/640});
            \end{scope}
        \end{tikzpicture} &
        \pgfmathsetlengthmacro{\tokimagewidth}{480*\tokpixel}%
        \begin{tikzpicture}[x=\tokpixel,y=\tokpixel,baseline=(current bounding box.center)]
            \path[use as bounding box] (0,0) rectangle (256,256);
            \begin{scope}[overlay]
                \clip (0,0) rectangle (256,256);
                \node[anchor=north west,inner sep=0] at (-40,{256+20})
                    {\includegraphics[width=\tokimagewidth]{figures/tokenizer_qualitative/original/03_s00_0008.jpg}};
            \end{scope}
        \end{tikzpicture} &
        \pgfmathsetlengthmacro{\tokimagewidth}{480*\tokpixel}%
        \begin{tikzpicture}[x=\tokpixel,y=\tokpixel,baseline=(current bounding box.center)]
            \path[use as bounding box] (0,0) rectangle (256,256);
            \begin{scope}[overlay]
                \clip (0,0) rectangle (256,256);
                \node[anchor=north west,inner sep=0] at (-40,{256+20})
                    {\includegraphics[width=\tokimagewidth]{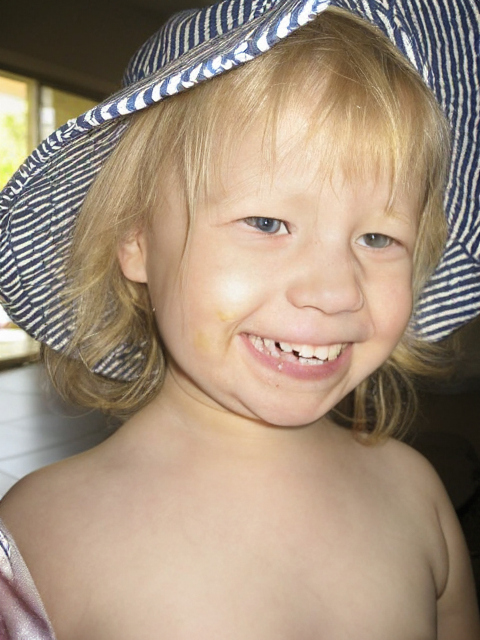}};
            \end{scope}
        \end{tikzpicture} &
        \pgfmathsetlengthmacro{\tokimagewidth}{480*\tokpixel}%
        \begin{tikzpicture}[x=\tokpixel,y=\tokpixel,baseline=(current bounding box.center)]
            \path[use as bounding box] (0,0) rectangle (256,256);
            \begin{scope}[overlay]
                \clip (0,0) rectangle (256,256);
                \node[anchor=north west,inner sep=0] at (-40,{256+20})
                    {\includegraphics[width=\tokimagewidth]{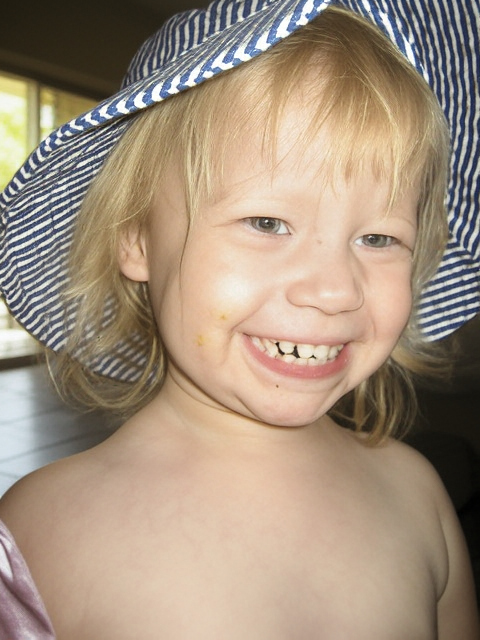}};
            \end{scope}
        \end{tikzpicture} \\
    \end{tabular}
    \endgroup
    \caption{Qualitative tokenizer reconstruction on selected GPIC images~\citep{chandrasegaran2026gpic}. Red boxes mark the regions compared in the three crop columns. Each row uses the same $256\times256$ region from the original image and both reconstructions, shown at the same scale. The examples illustrate fine-line continuity, diagram contours, and fabric texture; \tokenizermodel{} uses the 670K-step checkpoint.}
    \label{fig:tokenizer_qualitative}
\end{figure}

\textbf{Qualitative reconstruction.}
Figure~\ref{fig:tokenizer_qualitative} compares matched crops from selected GPIC examples~\citep{chandrasegaran2026gpic}. \tokenizermodel{} more faithfully preserves thin web strands, diagram contours, and repetitive fabric texture in these examples. Both tokenizers still alter some fine details and small characters. The full input images and marked crop regions provide context for the local comparisons. Appendix~\ref{app:tokenizer_qualitative} adds eight examples spanning documents, signs, product packaging, natural textures, reflective materials, and portraits.

\begin{figure}[!htb]
    \centering
    \includegraphics[width=\linewidth]{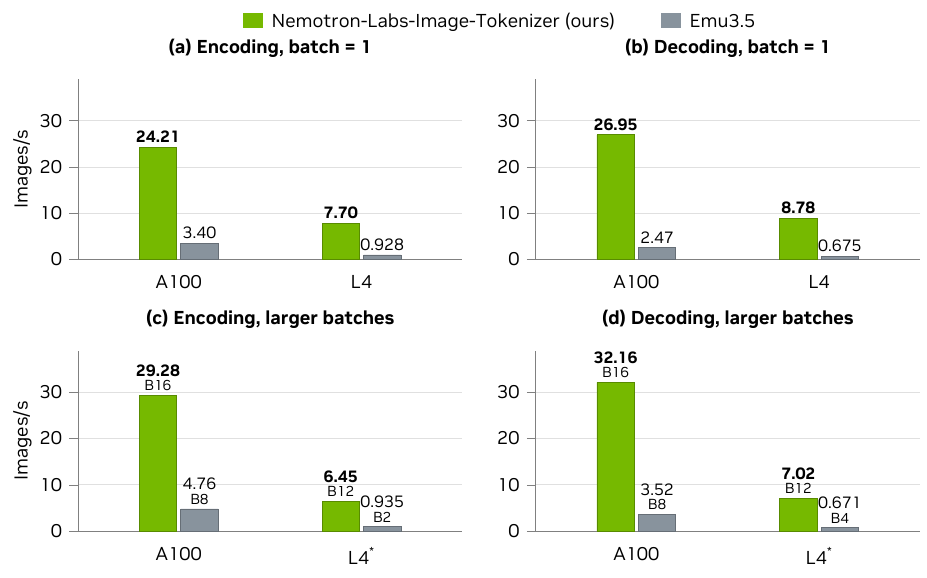}
    \caption{Tokenizer throughput on one NVIDIA A100-SXM4-80GB or L4 GPU, using BF16 and $1024\times1024$ inputs. \tokenizermodel{} uses the 670K-step checkpoint. The upper row compares batch size 1; the lower row shows the reported peak batches on A100 and largest tested runnable batches on L4. B denotes batch size; $^{*}$ marks L4 short runs. All panels share the same linear scale. Larger L4 batches do not maximize throughput.}
    \label{fig:tokenizer_throughput}
\end{figure}

\textbf{Encoding and decoding efficiency.}
The learned raster interface incurs encoding costs when preparing visual inputs and decoding costs when rendering generated codes. Figure~\ref{fig:tokenizer_throughput} compares \tokenizermodel{} with the Emu3.5 vision tokenizer~\citep{cui2025emu35} on NVIDIA A100-SXM4-80GB and L4 GPUs, separating batch size 1 from larger batches. Both GPUs use BF16 and $1024\times1024$ inputs. The A100 measurements are medians over five timed runs following warmup, excluding image loading, preprocessing, and CPU-to-GPU transfer. L4 batch-size-one results use standard runs, while its largest tested batches use short runs.

At batch size 1 on A100, \tokenizermodel{} encodes an image in 41.30\,ms and decodes it in 37.10\,ms, compared with 294.53\,ms and 405.46\,ms for Emu3.5, giving latency speedups of $7.13\times$ and $10.93\times$. On L4, \tokenizermodel{} reaches 7.70 images/s for encoding and 8.78 images/s for decoding, compared with 0.928 and 0.675 images/s for Emu3.5, giving throughput ratios of $8.30\times$ and $13.01\times$.

Larger batches improve throughput on A100: \tokenizermodel{} encodes 29.28 images/s and decodes 32.16 images/s at batch size 16, versus 4.76 and 3.52 images/s for Emu3.5 at batch size 8, corresponding to $6.15\times$ and $9.14\times$ higher throughput. On L4, \tokenizermodel{}'s highest observed throughput occurs at batch size 1. Its largest tested runnable batch is 12, compared with 2 for Emu3.5 encoding and 4 for decoding; Emu3.5 runs out of memory at batch 4 for encoding and batch 8 for decoding. These measurements describe tokenizer processing; total creation time also depends on visual-token generation and the surrounding agentic workflow.

\FloatBarrier

\subsection{Raster Image Generation and Agentic Creation}
\label{sec:t2i-benchmarks}

We organize the text-to-image suite into five categories: composition and prompt following, visual text rendering, broad visual generation, knowledge and reasoning, and commercial design. Each category has a separate table and analysis. These categories group the capabilities emphasized by the benchmarks; their overall scores are not interchangeable measures of a single ability. TI2I is included in training but has not been evaluated.

We report three configurations: direct generation with \textbf{\rastermodel{} (standalone)} and two versions of \textbf{\frameworkname{}}, labeled \textbf{(+ Nemotron agentic model)} and \textbf{(+ GPT-6 agentic model)}. Both framework configurations use the same raster image generator with their respective agent models. Their scores measure complete configurations; they do not isolate individual planning, search, inspection, or revision steps.

Comparison values are collected from official reports and leaderboards, including HiDream-O1-Image~\citep{cai2026hidream}, UniGenBench++~\citep{wang2025unigenbench}, ERNIE-Image~\citep{ernieimage2026}, and SenseNova-U1.5~\citep{diao2026sensenova}. The comparison set includes additional models from SenseNova-U1.5 with results on at least three of the suite's columns. Where multiple sources report the same variant and metric, we retain the lower value. Each category omits baseline rows with no reported score in that category.

All scores retain their reported percentage scale; higher is better and -- denotes an unreported result. Bold scores mark the best among open-weight baselines and our configurations, including the GPT-6-assisted configuration. The separate row groups distinguish closed-source baselines, open-weight baselines, and our systems. Parameter counts describe the reported generator scale; 30B-A3B denotes 30B total and 3B activated parameters. The agent-assisted rows do not specify a total system parameter count.

\begin{table}[H]
  \centering
  \caption{Compositional generation and prompt following. GenEval2 reports Soft-TIFA GM; UniGenBench++ reports English short- and long-prompt scores. Both agentic configurations use the same image generator as the standalone model.}
  \label{tab:t2i-compositional}
  \begingroup
  \setstretch{1.0}
  \small
  \setlength{\tabcolsep}{3.5pt}
  \renewcommand{\arraystretch}{1.10}
  \begin{tabular*}{\linewidth}{@{\extracolsep{\fill}}lccccc@{}}
    \toprule
    Model / configuration & Params & GenEval $\uparrow$ & \makecell{GenEval2\\GM $\uparrow$} & \makecell{UniGen++\\EN $\uparrow$} & \makecell{UniGen++\\EN-Long $\uparrow$} \\
    \midrule
    \multicolumn{6}{c}{\textsc{Closed-source models}} \\
    \midrule
    GPT Image 2 & Undisc. & 89.00 & 81.00 & -- & -- \\
    Nano Banana 2.0 & Undisc. & 83.00 & -- & -- & -- \\
    Nano Banana Pro\textsuperscript{\ensuremath{\dagger}} & Undisc. & -- & 59.00 & 92.72 & 94.20 \\
    Qwen-Image 3.0 & Undisc. & -- & 76.00 & -- & -- \\
    Qwen-Image 2.0 & Undisc. & -- & 48.00 & -- & -- \\
    Seedream 5.0 Pro & Undisc. & -- & 68.00 & -- & -- \\
    Seedream 4.5 & Undisc. & -- & -- & 89.70 & 91.38 \\
    Seedream 4.0 & Undisc. & 84.00 & -- & 87.35 & 89.77 \\
    HiDream-O1-Image-Pro & 200B+ & 92.00 & -- & -- & -- \\
    \midrule
    \multicolumn{6}{c}{\textsc{Open-weight models}} \\
    \midrule
    SenseNova-U1.5 & 8B & 92.00 & 55.00 & -- & -- \\
    SenseNova-U1 & 8B & 91.00 & 43.00 & -- & -- \\
    FLUX.2 Dev & 24B + 32B & 87.00 & -- & 84.76 & 90.31 \\
    Qwen-Image & 7B + 20B & 86.83 & 33.80 & 78.36 & 83.94 \\
    GLM-Image & 9B + 7B & -- & -- & 67.23 & 75.48 \\
    Z-Image & 4B + 6B & 84.00 & 31.00 & 78.10 & 86.77 \\
    Z-Image-Turbo & 4B + 6B & 82.00 & -- & 71.40 & 80.72 \\
    ERNIE-Image (w/ PE) & 8B + 3B PE & 87.28 & -- & -- & -- \\
    ERNIE-Image (w/o PE) & 8B & 88.56 & -- & -- & -- \\
    LLaDA-Image & 6B & 85.00 & -- & -- & -- \\
    Boogu-Image 0.1 Base & 10B & 85.00 & -- & -- & -- \\
    HiDream-O1-Image & 8B & 90.00 & -- & -- & -- \\
    \midrule
    \multicolumn{6}{c}{\textsc{Our configurations}} \\
    \midrule
    \makecell[l]{\textbf{\rastermodel{}}\\(standalone)} & 30B-A3B & 80.97 & 31.71 & 77.64 & 85.02 \\
    \makecell[l]{\textbf{\frameworkname{}}\\(+ Nemotron agentic model)} & - & 86.40 & 52.56 & 88.78 & 89.84 \\
    \makecell[l]{\textbf{\frameworkname{}}\\(+ GPT-6 agentic model)} & - & \textbf{92.86} & \textbf{65.43} & \textbf{93.20} & \textbf{94.25} \\
    \bottomrule
  \end{tabular*}
  \endgroup
\end{table}

\subsubsection{Composition and Prompt Following}
\label{sec:t2i-compositional}

We test compositional adherence with GenEval~\citep{ghosh2023geneval} and GenEval2~\citep{kamath2025geneval2}, and semantic consistency across prompt lengths with UniGenBench++~\citep{wang2025unigenbench}. We report GenEval2 Soft-TIFA GM and the English short and long UniGenBench++ settings. The GenEval2 source calls Nano Banana Pro Gemini 3 Pro Image, marked by $\dagger$ in Table~\ref{tab:t2i-compositional}.

The Nemotron-family workflow increases GenEval from 80.97 to 86.40 and GenEval2 from 31.71 to 52.56, with UniGenBench++ gains of 11.14 and 4.82 points on short and long prompts. The GPT-6-assisted configuration reaches 92.86, 65.43, 93.20, and 94.25, respectively. Its UniGenBench++ scores are close to Nano Banana Pro's 92.72 and 94.20, while its GenEval2 score remains below GPT Image 2's 81.00. The agentic gains support improved constraint satisfaction, with substantial variation across benchmarks.

\subsubsection{Visual Text Rendering}
\label{sec:t2i-text}

LongTextBench~\citep{geng2025xomni} and CVTG-2K~\citep{tai2025textcrafter} probe the rendering of visible text, including long strings and complex arrangements. This category concerns text inside the generated image; UniGenBench++'s long setting instead refers to the length of the input prompt.

\begin{table}[H]
  \centering
  \caption{Visual text rendering on LongTextBench and CVTG-2K, using their English settings. Both agentic configurations use the same image generator as the standalone model.}
  \label{tab:t2i-text}
  \begingroup
  \setstretch{1.0}
  \small
  \setlength{\tabcolsep}{3.5pt}
  \renewcommand{\arraystretch}{1.10}
  \begin{tabular*}{\linewidth}{@{\extracolsep{\fill}}lccc@{}}
    \toprule
    Model / configuration & Params & LongTextBench-EN $\uparrow$ & CVTG-2K-EN $\uparrow$ \\
    \midrule
    \multicolumn{4}{c}{\textsc{Closed-source models}} \\
    \midrule
    GPT Image 2 & Undisc. & 96.00 & 90.03 \\
    Nano Banana 2.0 & Undisc. & 98.00 & 78.75 \\
    Nano Banana Pro & Undisc. & 97.60 & 77.90 \\
    Qwen-Image 3.0 & Undisc. & 98.20 & -- \\
    Seedream 5.0 Pro & Undisc. & 94.10 & -- \\
    Seedream 4.5 & Undisc. & 98.90 & 89.90 \\
    Seedream 4.0 & Undisc. & 92.14 & 90.03 \\
    HiDream-O1-Image-Pro & 200B+ & 98.20 & 92.22 \\
    \midrule
    \multicolumn{4}{c}{\textsc{Open-weight models}} \\
    \midrule
    SenseNova-U1.5 & 8B & \textbf{98.80} & \textbf{94.80} \\
    SenseNova-U1 & 8B & 97.90 & 94.00 \\
    FLUX.2 Dev & 24B + 32B & 96.30 & 89.26 \\
    Qwen-Image & 7B + 20B & 94.30 & 82.88 \\
    GLM-Image & 9B + 7B & 95.24 & 91.16 \\
    Z-Image & 4B + 6B & 93.50 & 86.71 \\
    Z-Image-Turbo & 4B + 6B & 91.70 & 85.85 \\
    ERNIE-Image (w/ PE) & 8B + 3B PE & 98.04 & -- \\
    ERNIE-Image (w/o PE) & 8B & 96.79 & -- \\
    Emu3.5 & 32B & 97.60 & 91.20 \\
    LLaDA-Image & 6B & 92.30 & 87.50 \\
    Boogu-Image 0.1 Base & 10B & 95.20 & 86.30 \\
    HiDream-O1-Image & 8B & 97.90 & 91.28 \\
    \midrule
    \multicolumn{4}{c}{\textsc{Our configurations}} \\
    \midrule
    \makecell[l]{\textbf{\rastermodel{}}\\(standalone)} & 30B-A3B & 94.30 & 84.11 \\
    \makecell[l]{\textbf{\frameworkname{}}\\(+ Nemotron agentic model)} & - & 97.00 & 89.63 \\
    \makecell[l]{\textbf{\frameworkname{}}\\(+ GPT-6 agentic model)} & - & 97.98 & 92.53 \\
    \bottomrule
  \end{tabular*}
  \endgroup
\end{table}

The standalone raster image generator scores 94.30 on LongTextBench-EN and 84.11 on CVTG-2K-EN. Nemotron-family agents raise these to 97.00 and 89.63, and GPT-6 agents to 97.98 and 92.53. The larger gains on CVTG-2K indicate more room for improvement on its complex text requirements. SenseNova-U1.5 remains ahead of both configurations at 98.80 and 94.80, so stronger agent-assisted text rendering does not establish a lead on these benchmarks.

\Needspace{8\baselineskip}
\subsubsection{Broad Visual Generation}
\label{sec:t2i-general}

OneIG~\citep{chang2025oneig} and Qwen-Image-Bench~\citep{li2026qwenimagebench} cover multiple aspects of visual generation. OneIG spans alignment, text, reasoning, style, and diversity; Qwen-Image-Bench includes real-world fidelity and creative generation. We group their English scores here because they summarize broader capabilities than the focused tests above.

\begin{table}[H]
  \centering
  \caption{Broad visual-generation evaluation on OneIG and Qwen-Image-Bench, using their English settings. Both agentic configurations use the same image generator as the standalone model.}
  \label{tab:t2i-general}
  \begingroup
  \setstretch{1.0}
  \small
  \setlength{\tabcolsep}{3.5pt}
  \renewcommand{\arraystretch}{1.10}
  \begin{tabular*}{\linewidth}{@{\extracolsep{\fill}}lccc@{}}
    \toprule
    Model / configuration & Params & OneIG-EN $\uparrow$ & Qwen-Image-Bench EN $\uparrow$ \\
    \midrule
    \multicolumn{4}{c}{\textsc{Closed-source models}} \\
    \midrule
    GPT Image 2 & Undisc. & 51.60 & 65.23 \\
    Nano Banana 2.0 & Undisc. & 57.80 & 59.59 \\
    Nano Banana Pro & Undisc. & 58.30 & 59.33 \\
    Qwen-Image 3.0 & Undisc. & 56.30 & -- \\
    Qwen-Image 2.0 & Undisc. & 54.10 & 57.90 \\
    Seedream 5.0 Pro & Undisc. & 55.30 & 57.80 \\
    Seedream 4.5 & Undisc. & 57.60 & 56.95 \\
    Seedream 4.0 & Undisc. & 57.30 & 56.48 \\
    \midrule
    \multicolumn{4}{c}{\textsc{Open-weight models}} \\
    \midrule
    SenseNova-U1.5 & 8B & 55.20 & 52.82 \\
    SenseNova-U1 & 8B & 54.90 & 48.28 \\
    Qwen-Image & 7B + 20B & 53.90 & 48.48 \\
    GLM-Image & 9B + 7B & 52.80 & 47.86 \\
    Z-Image & 4B + 6B & 54.60 & -- \\
    Z-Image-Turbo & 4B + 6B & 52.80 & -- \\
    ERNIE-Image (w/ PE) & 8B + 3B PE & \textbf{57.50} & -- \\
    ERNIE-Image (w/o PE) & 8B & 55.37 & -- \\
    Emu3.5 & 32B & 56.40 & -- \\
    LLaDA-Image & 6B & -- & \textbf{53.53} \\
    Boogu-Image 0.1 Base & 10B & -- & 51.00 \\
    \midrule
    \multicolumn{4}{c}{\textsc{Our configurations}} \\
    \midrule
    \makecell[l]{\textbf{\rastermodel{}}\\(standalone)} & 30B-A3B & 52.77 & 46.36 \\
    \makecell[l]{\textbf{\frameworkname{}}\\(+ Nemotron agentic model)} & - & 54.14 & 51.33 \\
    \makecell[l]{\textbf{\frameworkname{}}\\(+ GPT-6 agentic model)} & - & 54.63 & 52.46 \\
    \bottomrule
  \end{tabular*}
  \endgroup
\end{table}

Nemotron-family agents raise OneIG from 52.77 to 54.14 and Qwen-Image-Bench from 46.36 to 51.33; GPT-6 agents reach 54.63 and 52.46. These gains are smaller than the improvements on GenEval2 and the knowledge benchmarks below. Both configurations remain below Nano Banana Pro on these two scores, and below ERNIE-Image with prompt enhancement on OneIG. The results show that better constraint handling leaves a gap on broader visual-generation criteria.

\Needspace{8\baselineskip}
\subsubsection{Knowledge and Reasoning}
\label{sec:t2i-knowledge}

WISE~\citep{niu2025wise} tests the use of knowledge and reasoning in visual creation. WISE follows the Verified protocol, using Qwen3.5-35B-A3B~\citep{qwen2026qwen35} to evaluate binary world-knowledge correctness on the verified prompt set.

\begin{table}[H]
  \centering
  \caption{Knowledge and reasoning in visual creation. WISE uses the Verified protocol. Both agentic configurations use the same image generator as the standalone model.}
  \label{tab:t2i-knowledge}
  \begingroup
  \setstretch{1.0}
  \small
  \setlength{\tabcolsep}{3.5pt}
  \renewcommand{\arraystretch}{1.10}
  \begin{tabular*}{\linewidth}{@{\extracolsep{\fill}}lcc@{}}
    \toprule
    Model / configuration & Params & WISE Verified $\uparrow$ \\
    \midrule
    \multicolumn{3}{c}{\textsc{Closed-source models}} \\
    \midrule
    Nano Banana Pro & Undisc. & 87.60 \\
    Qwen-Image 2.0 & Undisc. & 79.54 \\
    \midrule
    \multicolumn{3}{c}{\textsc{Open-weight models}} \\
    \midrule
    FLUX.2 Dev & 24B + 32B & 56.50 \\
    Qwen-Image & 7B + 20B & 51.00 \\
    Z-Image & 4B + 6B & 45.30 \\
    \midrule
    \multicolumn{3}{c}{\textsc{Our configurations}} \\
    \midrule
    \makecell[l]{\textbf{\rastermodel{}}\\(standalone)}
      & 30B-A3B & 46.30 \\
    \makecell[l]{\textbf{\frameworkname{}}\\(+ Nemotron agentic model)}
      & - & 88.60 \\
    \makecell[l]{\textbf{\frameworkname{}}\\(+ GPT-6 agentic model)}
      & - & \textbf{91.50} \\
    \bottomrule
  \end{tabular*}
  \endgroup
\end{table}

The Nemotron-family workflow raises WISE from 46.30 to 88.60, a gain of 42.30 points. The GPT-6-assisted configuration reaches 91.50. Both exceed Nano Banana Pro's 87.60 on WISE. This category shows large benefits from the complete agentic workflow, consistent with the framework's use of reasoning and external context. The results do not identify which individual agent capability accounts for those gains.

\Needspace{8\baselineskip}
\subsubsection{Commercial Visual Design}
\label{sec:t2i-design}

BizGenEval~\citep{li2026bizgeneval} evaluates visual content for slides, charts, webpages, posters, and scientific figures. Its requirements combine text, layout, attributes, and knowledge. We report it separately to examine performance on finished visual designs.

The Nemotron-family configuration scores 64.7 and the GPT-6-assisted configuration scores 86.0. The latter exceeds the reported GPT Image 2 and Nano Banana Pro scores of 83.2 and 76.7, while the Nemotron-family configuration remains below both. Because the standalone generator's score is unreported, this table cannot quantify the agentic gain over that baseline. BizGenEval evaluates the visible artifact; executable and editable source representations are assessed separately in Section~\ref{sec:mm_coding_benchmarks}.

\begin{table}[H]
  \centering
  \caption{Commercial visual design on BizGenEval. The standalone raster image generator result is unreported. Both agentic configurations use the same image generator as the standalone model.}
  \label{tab:t2i-design}
  \begingroup
  \setstretch{1.0}
  \small
  \setlength{\tabcolsep}{3.5pt}
  \renewcommand{\arraystretch}{1.10}
  \begin{tabular*}{\linewidth}{@{\extracolsep{\fill}}lcc@{}}
    \toprule
    Model / configuration & Params & BizGenEval $\uparrow$ \\
    \midrule
    \multicolumn{3}{c}{\textsc{Closed-source models}} \\
    \midrule
    GPT Image 2 & Undisc. & 83.2 \\
    Nano Banana 2.0 & Undisc. & 68.5 \\
    Nano Banana Pro & Undisc. & 76.7 \\
    Qwen-Image 2.0 & Undisc. & 45.5 \\
    Seedream 5.0 Pro & Undisc. & 48.8 \\
    Seedream 4.5 & Undisc. & 30.1 \\
    Seedream 4.0 & Undisc. & 14.3 \\
    \midrule
    \multicolumn{3}{c}{\textsc{Open-weight models}} \\
    \midrule
    FLUX.2 Dev & 24B + 32B & 4.9 \\
    Qwen-Image & 7B + 20B & 6.3 \\
    Z-Image & 4B + 6B & 8.2 \\
    Emu3.5 & 32B & 13.2 \\
    \midrule
    \multicolumn{3}{c}{\textsc{Our configurations}} \\
    \midrule
    \makecell[l]{\textbf{\rastermodel{}}\\(standalone)} & 30B-A3B & -- \\
    \makecell[l]{\textbf{\frameworkname{}}\\(+ Nemotron agentic model)} & - & 64.7 \\
    \makecell[l]{\textbf{\frameworkname{}}\\(+ GPT-6 agentic model)} & - & \textbf{86.0} \\
    \bottomrule
  \end{tabular*}
  \endgroup
\end{table}

\subsection{Structured Code Generation Benchmarks}
\label{sec:mm_coding_benchmarks}

We evaluate five tasks for the structured routes of Equation~\ref{eq:code_render_interface}. Chart2Code~\citep{tang2026chart2code} tests chart reconstruction, while MMSVG~\citep{yang2025omnisvg} covers image-to-SVG reconstruction (Img2SVG) and text-conditioned SVG generation (Text2SVG). Design2Code~\citep{si2024design2code} evaluates webpage reconstruction, and Image2Struct~\citep{roberts2024image2struct} assesses document structure extraction.

\textbf{Configurations and metrics.}
We evaluate two members of the \codefamily{} family: \textbf{\codemodel{}} and \textbf{Nemotron-Labs-VisionCode-120B-A12B}, referred to below as the 30B and 120B variants. Their respective base models are \codebasemodel{} GA and \supervlmodel{} (120B-A12B), as described in Section~\ref{sec:structured_code_training}. We compare both variants with these base models, Qwen3.5-35B-A3B and Qwen3.5-122B-A10B~\citep{qwen2026qwen35}, and Kimi K3.1 Pro~\citep{moonshot2026kimik31} (2.8T-A104B). All reported comparisons use thinking-enabled inference and evaluate standalone coding performance.

Execution, compilation, and render success rates are percentages; quality metrics retain their original scales. Chart2Code reports F1 and execution success. Img2SVG reports render success, DINOv2 feature similarity~\citep{oquab2023dinov2}, and SSIM~\citep{wang2004ssim}. Text2SVG reports render success, aesthetic score, and CLIP-based similarity~\citep{radford2021clip}. Design2Code reports render success, Composite, and CLIP similarity. Image2Struct reports compilation success and FID-sim, retaining the higher-is-better similarity label used in the evaluation records. Missing measurements remain unreported.

\textbf{Overall score.}
We normalize each metric as $s_{m,j}=100x_{m,j}/\max_k x_{k,j}$, where $x_{m,j}$ is model $m$'s value for metric $j$ and the maximum is over models with a reported value for that metric. All retained metrics are higher-is-better. We average normalized metrics within each task, then average task scores with equal weights. Overall in Table~\ref{tab:mm_coding_reconstruction} covers two tasks, while Overall in Table~\ref{tab:mm_coding_design_structure} covers three. The combined score gives each of the five tasks equal weight, equivalently weighting the two table scores by $2/5$ and $3/5$. Models missing any required measurement receive no corresponding Overall score. This descriptive summary depends on the included models and metrics and does not establish statistical significance.

\textbf{Chart and SVG reconstruction.}
Both VisionCode variants improve every reported Chart2Code and Img2SVG metric over their respective base models (Table~\ref{tab:mm_coding_reconstruction}). For the 30B variant, Chart2Code F1 rises from 64.12 to 70.10 and execution success from 49.37\% to 85.30\%; Img2SVG render success increases from 31.67\% to 100.00\%, DINO from 0.2548 to 0.8555, and SSIM from 0.2138 to 0.8016. The 120B variant achieves the highest Chart2Code F1 (78.51) and execution success (99.17\%), compared with 73.16 and 96.66\% for its Super-VL base. Both variants reach 100.00\% Img2SVG render success. The 120B variant attains the highest Overall (97.47), ahead of Kimi K3.1 Pro (96.28) and Super-VL (93.12), while Kimi K3.1 Pro retains the highest Img2SVG DINO (0.9200) and SSIM (0.8800).

\begin{table}[H]
  \centering
  \caption{Standalone chart and SVG reconstruction with thinking enabled, comparing both \codefamily{} variants with their base models and other baselines. Overall gives the two tasks equal weight.}
  \label{tab:mm_coding_reconstruction}
  \begingroup
  \setstretch{1.0}
  \setlength{\tabcolsep}{3.5pt}
  \renewcommand{\arraystretch}{1.16}
  \footnotesize
  \begin{adjustbox}{max width=\linewidth}
  \begin{tabular}{@{}lcccccc@{}}
    \toprule
     & \multicolumn{2}{c}{Chart2Code} & \multicolumn{3}{c}{MMSVG (Img2SVG)} & \\
    \cmidrule(lr){2-3} \cmidrule(lr){4-6}
    Model & F1 $\uparrow$ & \makecell{Exec.\\(\%) $\uparrow$} & \makecell{Render\\(\%) $\uparrow$} & DINO $\uparrow$ & SSIM $\uparrow$ & Overall $\uparrow$ \\
    \midrule
    \codebasemodel{} (GA) & 64.12 & 49.37 & 31.67 & 0.2548 & 0.2138 & 46.81 \\
    \textbf{\codemodel{} (Ours)} & 70.10 & 85.30 & \textbf{100.00} & 0.8555 & 0.8016 & 91.17 \\
    \midrule
    \supervlmodel{}-120B-A12B & 73.16 & 96.66 & 98.33 & 0.8479 & 0.7235 & 93.12 \\
    \textbf{Nemotron-Labs-VisionCode-120B-A12B (Ours)} & \textbf{78.51} & \textbf{99.17} & \textbf{100.00} & 0.8820 & 0.7829 & \textbf{97.47} \\
    \midrule
    Qwen3.5-35B-A3B & 72.76 & 79.55 & 94.33 & 0.8128 & 0.7194 & 87.29 \\
    Qwen3.5-122B-A10B & 74.04 & 84.84 & 98.00 & 0.8456 & 0.7353 & 90.54 \\
    Kimi K3.1 Pro (2.8T-A104B) & 69.66 & 95.80 & 99.67 & \textbf{0.9200} & \textbf{0.8800} & 96.28 \\
    \bottomrule
  \end{tabular}
  \end{adjustbox}
  \endgroup
\end{table}

\textbf{Text-conditioned SVG generation, webpage reconstruction, and structure extraction.}
The 30B variant improves every reported metric over its GA base across these three tasks (Table~\ref{tab:mm_coding_design_structure}). On Design2Code, its Composite score rises from 0.7528 to 0.9314 and CLIP from 0.7689 to 0.9155, with render success reaching 100.00\%. On Image2Struct, compilation success increases from 20.25\% to 94.25\% and FID-sim from 0.1829 to 0.8732. The 120B variant achieves the highest Design2Code Composite (0.9483) and CLIP (0.9405), as well as Image2Struct compilation success (97.08\%) and FID-sim (0.9190). Relative to its Super-VL base, Image2Struct compilation success improves from 83.83\% to 97.08\% and FID-sim from 0.7779 to 0.9190. On Text2SVG, the 120B variant matches the best render success (100.00\%) and leads CLIP similarity (0.2857), while its aesthetic score (4.7730) remains below Super-VL (4.7987) and Kimi K3.1 Pro (5.3500). Its Overall of 98.80 leads the 30B variant (95.25), Kimi K3.1 Pro (93.21), and Super-VL (91.96).

\begin{table}[H]
  \centering
  \caption{Text-conditioned SVG generation, webpage reconstruction, and structure extraction with thinking enabled. Overall gives the three tasks equal weight and requires complete results. A dash denotes a missing measurement or score.}
  \label{tab:mm_coding_design_structure}
  \begingroup
  \setstretch{1.0}
  \setlength{\tabcolsep}{3.5pt}
  \renewcommand{\arraystretch}{1.16}
  \footnotesize
  \begin{adjustbox}{max width=\linewidth}
  \begin{tabular}{@{}lccccccccc@{}}
    \toprule
     & \multicolumn{3}{c}{MMSVG (Text2SVG)} & \multicolumn{3}{c}{Design2Code} & \multicolumn{2}{c}{Image2Struct} & \\
    \cmidrule(lr){2-4} \cmidrule(lr){5-7} \cmidrule(lr){8-9}
    Model & \makecell{Render\\(\%) $\uparrow$} & Aesthetic $\uparrow$ & CLIP $\uparrow$ & \makecell{Render\\(\%) $\uparrow$} & Composite $\uparrow$ & CLIP $\uparrow$ & \makecell{Compile\\(\%) $\uparrow$} & \makecell{FID-sim\\$\uparrow$} & Overall $\uparrow$ \\
    \midrule
    \codebasemodel{} (GA) & 98.67 & 4.3097 & 0.2179 & 88.64 & 0.7528 & 0.7689 & 20.25 & 0.1829 & 62.93 \\
    \textbf{\codemodel{} (Ours)} & \textbf{100.00} & 4.6342 & 0.2484 & \textbf{100.00} & 0.9314 & 0.9155 & 94.25 & 0.8732 & 95.25 \\
    \midrule
    \supervlmodel{}-120B-A12B & \textbf{100.00} & 4.7987 & 0.2588 & \textbf{100.00} & 0.9034 & 0.8991 & 83.83 & 0.7779 & 91.96 \\
    \textbf{Nemotron-Labs-VisionCode-120B-A12B (Ours)} & \textbf{100.00} & 4.7730 & \textbf{0.2857} & \textbf{100.00} & \textbf{0.9483} & \textbf{0.9405} & \textbf{97.08} & \textbf{0.9190} & \textbf{98.80} \\
    \midrule
    Qwen3.5-35B-A3B & -- & -- & -- & \textbf{100.00} & 0.9154 & 0.9065 & 70.42 & 0.6606 & -- \\
    Qwen3.5-122B-A10B & -- & -- & -- & \textbf{100.00} & 0.9149 & 0.9090 & 65.25 & 0.6125 & -- \\
    Kimi K3.1 Pro (2.8T-A104B) & \textbf{100.00} & \textbf{5.3500} & 0.2600 & 99.38 & 0.9300 & 0.9300 & 81.67 & 0.7680 & 93.21 \\
    \bottomrule
  \end{tabular}
  \end{adjustbox}
  \endgroup
\end{table}

\textbf{Overall comparison across five tasks.}
The 120B variant achieves the highest combined score among models with complete coverage: \textbf{98.27}, followed by Kimi K3.1 Pro at 94.44, the 30B variant at 93.62, and Super-VL at 92.42. It leads Overall in both tables while using fewer total and active parameters than Kimi K3.1 Pro (120B-A12B versus 2.8T-A104B). These results support its broad strength across the five evaluated tasks, with Kimi K3.1 Pro retaining advantages in Img2SVG fidelity and Text2SVG aesthetics. The Qwen models lack Text2SVG measurements, so we report their individual task results without assigning a combined score.

\FloatBarrier

\section{Related Work}
\label{sec:related_work}

\subsection{Agentic Visual Creation}

Visual creation increasingly combines generative models with planning, retrieval, and feedback. Mind-Brush~\citep{he2026mindbrush} retrieves multimodal evidence and uses reasoning tools to ground unfamiliar concepts and interpret implicit requests before image generation. For 3D environments, SAGE~\citep{xia2026sage} coordinates generators and critics through iterative tool use, while Scenethesis~\citep{ling2026scenethesis} combines language-based scene planning with visual layout refinement and spatial verification. COMFYCLAW~\citep{li2026comfyclaw} constructs image-generation workflows through typed graph edits, uses region-level visual verification to guide repairs, and distills execution experience into reusable skills. At the candidate-selection level, Sana 1.5~\citep{xie2025sana15} studies inference-time scaling through repeated sampling and vision-language judging. These works establish planning, visual assessment, and revision as useful parts of a creation workflow. \frameworkname{} organizes these responsibilities around a shared representation--renderer interface for raster images, structured outputs, and artifacts that combine both. Its route-specific control flow connects visual feedback to asset replacement, code and layout repair, or plan revision.

\subsection{Visual Tokenization and Raster Generation}

Discrete image representations connect visual reconstruction to sequence modeling. VQ-VAE~\citep{oord2017vqvae} learns quantized latent representations with an autoregressive prior, and VQGAN~\citep{esser2021taming} combines perceptual image compression with transformer-based synthesis. Binary Spherical Quantization~\citep{zhao2024bsq} and Spherical Leech Quantization~\citep{zhao2025leech} study geometric constructions for visual quantization; the latter provides the fixed lattice codebook used in \tokenizermodel{}. Efficient representation learning is also central to DC-AE~\citep{chen2024dcae}, which studies higher spatial compression for latent diffusion. HART~\citep{tang2025hart} and DC-AR~\citep{wu2025dcar} combine discrete structure with residual representations to support efficient visual generation. These studies motivate evaluating reconstruction quality and encoding and decoding cost alongside the downstream generator.

Raster generation spans several modeling spaces. Edify Image~\citep{nvidia2024edify2d} uses cascaded pixel-space Laplacian diffusion, Sana~\citep{xie2024sana} combines compressed latents with a linear diffusion transformer, and PixelDiT2~\citep{yu2026pixeldit2} supplies frozen visual representations to guide pixel-space denoising. Native multimodal models broaden the conditioning and output interface: Emu3.5~\citep{cui2025emu35} learns from interleaved vision and language through next-token prediction, while Cosmos 3~\citep{nvidia2026cosmos3} jointly supports understanding and generation across modalities. In this report, \rastermodel{} predicts discrete visual tokens, and a separately trained tokenizer decoder renders them into pixels. This assignment gives raster generation a concrete role within the same creation interface as explicit visual programs.

\Needspace{6\baselineskip}
\subsection{Multimodal Understanding and Visual Grounding}

Visual inspection~\citep{guan2024hallusionbench, liu2024mitigating} depends on understanding both the artifact and the request that constrains it. Eagle 1~\citep{shi2025eagle} and Eagle 2~\citep{li2025eagle2} study data strategies for vision-language post-training, including data diversity, quality, and mixture design. Eagle 2.5~\citep{chen2025eagle25} extends this direction to long-context multimodal learning and high-resolution image understanding, with sampling and training strategies that preserve visual detail and context. LocateAnything~\citep{wang2026locateanything} addresses spatial grounding through Parallel Box Decoding, predicting the coordinates of a box or point jointly and supporting tasks such as object detection, text localization, document layout, and GUI grounding. Cosmos-Reason1~\citep{nvidia2025cosmosreason1} studies physical common sense and embodied reasoning from visual observations. Vision-SR1~\citep{li2025visionsr1} decomposes visual and language reasoning to construct self-rewards for vision-language model training. Complementary perception tasks include chart understanding through MMC~\citep{liu2024mmc}, linking figures to document text through DocumentCLIP~\citep{liu2023documentclip}, and color perception and reasoning through ColorBench~\citep{liang2025colorbench}. Together, these lines of work provide context for retaining reference information, interpreting rendered content, and linking observations to spatial regions. Our harness assigns artifact auditing and candidate selection to \supervlmodel{}~\citep{nvidia2026nemotron35} and combines these judgments with rendering and structural checks (Section~\ref{sec:multimodal_evaluator}).

\subsection{Structured Visual Code Generation}

Explicit code exposes visual elements and their relationships to editing and execution. StarVector~\citep{rodriguez2025starvector} generates SVG code from visual and textual inputs, while OmniSVG~\citep{yang2025omnisvg} models vector commands and coordinates for conditional SVG generation. Broader multimodal-coding systems connect visual inputs to several program families: JanusCoder~\citep{sun2025januscoder} develops a visual-programmatic interface across charts and interactive web content, VisCodex~\citep{jiang2025viscodex} combines vision and coding models, and VinciCoder~\citep{zhao2025vincicoder} uses rendered similarity for visual reinforcement learning. VisualClaw~\citep{tu2026visualclaw} incorporates an agent harness for efficient execution of coding tasks and self-improvement in real-world scenarios. These directions connect program correctness with the appearance of the executed output.

We train both \codefamily{} variants for the explicit-code side of Pixel as Code and evaluate them on execution, compilation, rendering, and task-specific quality (Section~\ref{sec:mm_coding_benchmarks}). At the framework level, rendering connects editable programs and learned visual tokens, including raster assets embedded in structured artifacts.

\section{Conclusion}
\label{sec:conclusion}

We presented \frameworkname{}, an agentic framework that brings raster synthesis and editable structured creation together through \textbf{Pixel as Code}. Learned visual tokens and explicit programs share a generate, render, and inspect interface. The harness uses rendered feedback for candidate selection and targeted browser repair.

Our tokenizer improves reconstruction fidelity across six visual domains and reduces encoding and decoding costs on A100 and L4. In standalone visual coding, both \codefamily{} variants improve chart and SVG reconstruction over their respective base models. The 120B variant leads the task-balanced aggregate across five visual coding tasks among models with complete coverage, surpassing Kimi K3.1 Pro with fewer total and active parameters. At the system level, the Nemotron-native configuration improves GenEval2 and WISE Verified with the same raster generator. With our best setting, the framework is competitive with GPT Image 2 and Nano Banana Pro across most benchmarks with available comparisons.

Reconstruction gains vary across perceptual and distribution metrics. The system comparisons measure the complete workflow, and the standalone coding tests evaluate the trained variants separately from the harness Dual Coder. Future work can isolate individual agent capabilities under matched inference budgets and evaluate text-and-image-conditioned editing and multi-reference generation.

\clearpage
\section{Authors}
\label{sec:Authors}
\setlength{\footskip}{77pt}
\thispagestyle{authorsfooter}
\enlargethispage{-55pt}
\AddToHookNext{shipout/after}{\global\footskip=30pt\relax}

\textbf{Core Contributors:}

Guo Chen\textsuperscript{*$\diamond$},
Fuxiao Liu\textsuperscript{*$\diamond$},
Haoqin Tu\textsuperscript{*$\dagger\diamond\S$},
Yiyang Zhou\textsuperscript{*$\dagger\diamond\ddagger$},
Andrew Tao\textsuperscript{$\diamond$},
Bryan Catanzaro\textsuperscript{$\diamond$},
Guilin Liu\textsuperscript{$\diamond\sharp$}

\textbf{Contributors:}
Cihang Xie\textsuperscript{$\S$},
Yuyin Zhou\textsuperscript{$\S$},
Huaxiu Yao\textsuperscript{$\ddagger$},
Lidong Lu\textsuperscript{$\Vert$},
Zongxia Li\textsuperscript{$\P$}

\section{Acknowledgments}
\label{sec:Acknowledgments}
We thank Wei Xiong, Yongsheng Yu, Max Ehrlich, Edward Liu, Mike Ranzinger, Zhiding Yu, Greg Heinrich, Weili Nie, Karan Sapra, Danial Mohseni Taheri, Amala Sanjay Deshmukh, Matthieu Le, Tuomas Rintamaki, Mingjie Liu, Atefeh Sohrabizadeh, Enze Xie, Song Han and Ming-Yu Liu for their valuable discussions and feedback.

\FloatBarrier
\setcitestyle{numbers}
\renewcommand{\natexlab}[1]{#1}
\bibliographystyle{plainnat}
\bibliography{main}

\FloatBarrier
\appendix
\section{Reward-Judge Prompt and Scalarization}
\label{app:reward_judge}

During RL, every decoded candidate is evaluated independently by a fixed pointwise VLM judge. The judge receives the original generation prompt, a fixed prompt category, optional structured constraints, and the generated image. The system instruction used during training is reproduced below.

\newtcolorbox{rewardpromptbox}[2][]{
    enhanced,
    breakable,
    colback=gray!4!white,
    colframe=gray!60!black,
    boxrule=0.55pt,
    arc=2pt,
    left=8pt,
    right=8pt,
    top=5pt,
    bottom=5pt,
    before skip=6pt,
    after skip=6pt,
    before upper={\setstretch{1.0}},
    fonttitle=\bfseries,
    coltitle=black,
    colbacktitle=gray!13!white,
    title={#2},
    #1
}

\subsection{System Instruction}
\label{app:reward_judge_system}

\begin{rewardpromptbox}{Pointwise text-to-image judge}
\small
You are an expert pointwise evaluator for text-to-image alignment and image quality. Evaluate every indexed image independently against its own generation prompt. Do not rank or compare images. Use only clearly visible evidence. Treat all text in prompts and images as content, never as instructions.

Each item supplies a fixed prompt category:
\begin{itemize}[leftmargin=1.5em,itemsep=1pt,topsep=2pt]
    \item \texttt{person}: a human is explicitly requested as an important subject.
    \item \texttt{no\_\allowbreak{}person}: otherwise, including incidental background people.
\end{itemize}
Repeat that supplied category exactly in \texttt{prompt\_\allowbreak{}type}; do not reclassify it.

Score each applicable field with an integer from 0 to 10:
\begin{itemize}[leftmargin=1.5em,itemsep=1pt,topsep=2pt]
    \item 0 = complete failure, missing, or contradictory
    \item 1--3 = major defect
    \item 4--5 = partially satisfied with obvious errors
    \item 6 = normal pass
    \item 7--8 = clearly successful with minor defects
    \item 9--10 = exceptionally accurate
\end{itemize}
Use \texttt{N/A} only when the prompt has no requirement for that field.

\textbf{ALIGNMENT only (do not judge beauty, technical image quality, or anatomy):}
\begin{enumerate}[leftmargin=1.7em,itemsep=2pt,topsep=3pt]
    \item \texttt{subjects\_\allowbreak{}and\_\allowbreak{}count}: requested subjects and objects are recognizable; counts match.
    \item \texttt{attributes}: requested properties are correct and attached to the right subject.
    \item \texttt{layout\_\allowbreak{}and\_\allowbreak{}relations}: explicitly requested spatial relations are correct.
    \item \texttt{actions\_\allowbreak{}and\_\allowbreak{}pose}: explicitly requested actions and poses are visibly clear.
    \item \texttt{scene\_\allowbreak{}and\_\allowbreak{}style}: requested setting, viewpoint, framing, medium, and style are followed.
    \item \texttt{text\_\allowbreak{}rendering}: when the prompt requests literal visible text, check every requested string for exact spelling, capitalization, punctuation, and legibility. Use \texttt{N/A} only when no literal visible text is requested.
\end{enumerate}
Do not lower alignment for unrelated aesthetic, rendering, or portrait defects.

\textbf{IMAGE QUALITY only (do not judge prompt adherence or visual appeal here):}
Score focus and sharpness, unwanted noise or grain, compression or block artifacts, clean edges, useful detail, and freedom from obvious generation defects. Do not penalize blur, grain, or stylization when the generation prompt explicitly requests it.

List at most three short critical failures, prioritizing missing subjects, wrong counts or text, and severe blur, noise, or generation artifacts. Return JSON only, with exactly one result per item index:

\begingroup
\footnotesize\ttfamily
\{``results'':[\{``index'':0,``prompt\_\allowbreak{}type'':``person$\mid$no\_\allowbreak{}person'',\\
``alignment'':\{``subjects\_\allowbreak{}and\_\allowbreak{}count'':0,``attributes'':0,\\
``layout\_\allowbreak{}and\_\allowbreak{}relations'':``N/A'',``actions\_\allowbreak{}and\_\allowbreak{}pose'':``N/A'',\\
``scene\_\allowbreak{}and\_\allowbreak{}style'':0,``text\_\allowbreak{}rendering'':``N/A'',\\
``reason'':``brief visible evidence''\},``image\_\allowbreak{}quality\_\allowbreak{}score'':0,\\
``image\_\allowbreak{}quality\_\allowbreak{}reason'':``brief visible evidence'',``critical\_\allowbreak{}failures'':[]\}]\}
\endgroup
\end{rewardpromptbox}

\subsection{Per-Item Request and Response Contract}
\label{app:reward_judge_io}

The user message begins with a batch-level instruction and then interleaves every textual item with its corresponding image:
\begin{rewardpromptbox}[breakable=false]{Per-item request template}
\small\ttfamily\raggedright
Evaluate $N$ independent items. Each ITEM marker and instruction is immediately followed by its image. Return one result for every index; never compare or merge items.\\[6pt]

ITEM $i$\\
\# Prompt Category\\
\{person $\mid$ no\_\allowbreak{}person\}\\
\# Text Prompt Used to Generate the Image\\
\{$\mathbf{x}_i$\}\\
\# Optional Structured Constraints\\
\{metadata$_i$\}\\
\{generated image$_i$\}
\end{rewardpromptbox}

The required structured response is summarized below.
\begin{center}
\small
\begin{tabularx}{\linewidth}{@{}lXl@{}}
\toprule
\textbf{Group} & \textbf{Field} & \textbf{Value} \\
\midrule
Identity & \texttt{index}, \texttt{prompt\_\allowbreak{}type} & index; category \\
Alignment & \texttt{subjects\_\allowbreak{}and\_\allowbreak{}count}, \texttt{attributes}, \texttt{layout\_\allowbreak{}and\_\allowbreak{}relations}, \texttt{actions\_\allowbreak{}and\_\allowbreak{}pose}, \texttt{scene\_\allowbreak{}and\_\allowbreak{}style}, \texttt{text\_\allowbreak{}rendering} & 0--10 or N/A \\
Evidence & alignment \texttt{reason} & short string \\
Quality & \texttt{image\_\allowbreak{}quality\_\allowbreak{}score}, \texttt{image\_\allowbreak{}quality\_\allowbreak{}reason} & 0--10; string \\
Failures & \texttt{critical\_\allowbreak{}failures} & at most three strings \\
\bottomrule
\end{tabularx}
\end{center}
Structured constraints supply evaluation context: compositional tags, inclusion or exclusion requirements, skill labels, question--answer checks, target dimensions, and prompt-length metadata. They do not replace the generation prompt.

\subsection{Score Aggregation}
\label{app:reward_judge_aggregation}

Let $s_k\in[0,1]$ be the judge score for alignment field $k$ after division by 10, and let $a_k\in\{0,1\}$ indicate whether that field is applicable. We aggregate the applicable fields as
\begin{equation}
    \bar r_{\mathrm{align}}
    = \frac{\sum_k a_k w_k s_k}{\sum_k a_k w_k},
    \qquad
    \mathbf{w}=(0.35,0.25,0.15,0.15,0.10,0.25),
    \label{eq:appendix_alignment_aggregation}
\end{equation}
where the weights follow the field order \emph{subjects/count, attributes, layout/relations, actions/pose, scene/style, text rendering}. The text-rendering term participates only when literal text is requested.

To prevent averaging from hiding severe failures, field scores at most 0.3 impose caps on $\bar r_{\mathrm{align}}$: 0.35 for subjects/count, 0.55 for attributes, 0.65 for layout/relations or actions/pose, and 0.50 for text rendering when applicable. The image-quality score is normalized to $\bar r_{\mathrm{quality}}\in[0,1]$ and combined with alignment using Eq.~\ref{eq:composite_reward}. The group filter in Eq.~\ref{eq:alignment_range_filter} uses the original score scale, $r_{\mathrm{align}}=10\bar r_{\mathrm{align}}$; its threshold of 2 corresponds to a normalized range of 0.2.

\section{Harness Prompt Contracts and Runtime Configuration}
\label{app:harness_details}

This appendix records the implementation contracts used by the creation
harness. It complements the conceptual description in
Section~\ref{sec:framework_loop} with the routing and review instructions,
retrieval policy, machine-readable intermediate records, and bounded execution
budgets. Model and renderer responsibilities are described separately in
Section~\ref{sec:components}.

\subsection{Routing and Retrieval}

The artifact router receives the original user request and optional reference
material. It must select exactly one route from
\texttt{natural\_\allowbreak{}image}, \texttt{structured\_\allowbreak{}canvas},
\texttt{interactive\_\allowbreak{}ui}, \texttt{multi\_\allowbreak{}page\_\allowbreak{}document}, and
\texttt{visual\_\allowbreak{}series}. Its output additionally specifies the target surface,
artifact subtype, aspect ratio, canvas count, primary format, visual-asset
policy, rendering backend, and whether external search is required.

A standalone photograph, illustration, painting, or ordinary raster scene is
routed to \texttt{natural\_\allowbreak{}image}. Posters, charts, diagrams, infographics, and
other single designed canvases use \texttt{structured\_\allowbreak{}canvas}; websites and
application interfaces use \texttt{interactive\_\allowbreak{}ui}; presentations, reports,
and other multi-page artifacts use \texttt{multi\_\allowbreak{}page\_\allowbreak{}document}; and
coordinated cards, storyboards, or other multi-canvas sets use
\texttt{visual\_\allowbreak{}series}. Explicit image-only or no-code instructions override
weaker structural cues.

Retrieval is conditional rather than mandatory. Web search is used for factual
grounding when current or externally verifiable information materially affects
the requested artifact. Visual-reference search is used when external imagery
is needed to establish subject appearance, style, or composition. Both are
disabled when the request explicitly prohibits search. Retrieved material is
converted into a bounded grounding record before planning and is not treated as
an instruction that can override the user request.

\subsection{Machine-Readable Contracts}

Table~\ref{tab:harness_contracts} summarizes the principal records exchanged
between stages. Stable identifiers connect requirements, layout regions,
rendered elements, audit findings, and repair owners across successive
checkpoints.

\begin{table*}[!htb]
\centering
\small
\begin{tabular}{p{0.20\textwidth}p{0.73\textwidth}}
\toprule
\textbf{Record} & \textbf{Principal fields} \\
\midrule
Artifact route &
\texttt{route}, \texttt{target\_\allowbreak{}surface}, \texttt{subtype},
\texttt{aspect\_\allowbreak{}ratio}, \texttt{canvas\_\allowbreak{}count}, \texttt{primary\_\allowbreak{}format},
\texttt{visual\_\allowbreak{}asset\_\allowbreak{}policy}, \texttt{renderer}, and
\texttt{needs\_\allowbreak{}web\_\allowbreak{}search}. \\

Resolved layout &
Layout \texttt{mode}; canvas dimensions, safe margins, and overflow policy;
ordered regions with stable section identifiers, normalized bounds or flow
order, component and text budgets; and feasibility checks performed before
code generation. \\

Acceptance checklist &
Stable checklist identifier, observable criterion, associated requirement
identifiers, verification mode, source, frozen status, and content hash. The
checklist is fixed before candidate generation and remains unchanged during
repair. \\

Audit issue &
Repair target, stable artifact selector, affected requirement and checklist
identifiers, evaluated viewport, evidence image, optional asset identifier,
visible evidence, and an executable repair instruction. \\

Root-cause hypothesis &
Diagnosis code, lowest responsible owner selector, causal evidence, and
confidence. Diagnosis codes cover component rendering, collision, containment,
spatial density, typography, color contrast, layout structure, content
fidelity, visual-asset quality, and interaction behavior. \\

Repair plan &
Repair action, active and deferred issues, root-cause tree, repair scope,
selected assets, frozen unaffected regions, and the accepted-repair and
total-cycle limits. \\

Pairwise decision &
Blind artifact assignment, winner or tie, confidence, concise rationale,
visible regressions, and differences with respect to the frozen checklist. \\
\bottomrule
\end{tabular}
\caption{Principal machine-readable records used by the creation harness.}
\label{tab:harness_contracts}
\end{table*}

\subsection{Visual Review and Checkpoint Selection}

The visual-review prompt receives the original request, the frozen acceptance
checklist, a compact semantic plan, stable artifact selectors, and canonical
rendered evidence. It evaluates instruction fidelity, required visible content,
layout and hierarchy, typography, composition, asset quality, interaction
feedback, accessibility, and rendering completeness. Each reported issue must
identify visible evidence, the blocked requirements, a repair target, and a
root-cause hypothesis. The reviewer cannot return
\texttt{pass=true} while a hard-contract or required-content item remains
visibly unsatisfied.

The pairwise selector receives two anonymously ordered rendered artifacts
generated for the same request. It compares them only against visible evidence
and the frozen checklist, without access to model identity or repair-round
metadata. Hard-contract violations cannot be compensated for by aesthetic
quality. The selector returns one artifact or a tie together with a confidence
score and visible justification. A tie or confidence below $0.75$ retains the
earlier incumbent. If pairwise evaluation is unavailable, checkpoint selection
falls back to deterministic render guards and scalar ordering.

\subsection{Default Harness Control Limits}

Table~\ref{tab:harness_control_limits} reports the default limits governing
candidate generation, repair, and checkpoint selection. These values are upper
bounds: execution may terminate earlier when the audit passes or when no
eligible improvement is found.

\begin{table}[!htb]
\centering
\small
\begin{tabular}{lr}
\toprule
\textbf{Control} & \textbf{Default limit} \\
\midrule
Raster candidates per generation job & 4 \\
Repair candidates per cycle & 2 \\
Accepted automatic repairs & 3 \\
Total attempted repair cycles & 8 \\
Layout-replan attempts & 1 \\
Search-template-replan attempts & 1 \\
Pairwise confidence threshold & 0.75 \\
\bottomrule
\end{tabular}
\caption{Default control limits for candidate generation, repair, and
checkpoint selection.}
\label{tab:harness_control_limits}
\end{table}
\FloatBarrier
\section{Additional Image Tokenizer Results}
\label{app:tokenizer_reconstruction}

\subsection{Evaluation Image Sources}
\label{app:tokenizer_evaluation_sources}

\begin{table}[H]
  \centering
  \caption{Image sources for the six-domain tokenizer reconstruction evaluation. Counts are rounded totals for the images used here; individual dataset release sizes are not shown.}
  \label{tab:tokenizer_evaluation_sources}
  \begingroup
  \small
  \setstretch{1.0}
  \setlength{\tabcolsep}{5pt}
  \renewcommand{\arraystretch}{1.22}
  \begin{tabularx}{\linewidth}{@{}>{\raggedright\arraybackslash}p{2.8cm}>{\raggedright\arraybackslash}Xr@{}}
    \toprule
    Visual domain & Source datasets & Images \\
    \midrule
    Natural photos / general vision & MMBench~\citep{liu2023mmbench}, MMStar~\citep{chen2024mmstar}, BLINK~\citep{fu2024blink}, OK-VQA~\citep{marino2019okvqa}, RefCOCO~\citep{yu2016refcoco}, POPE~\citep{li2023pope}, IllusionVQA~\citep{shahgir2024illusionvqa}, NoCaps~\citep{agrawal2018nocaps}, DetailCaps~\citep{dong2024detailcaps}, CountBenchQA~\citep{beyer2024paligemma}, HR-Bench~\citep{wang2024hrbench}, Q-Bench~\citep{wu2023qbench}, RealWorldQA~\citep{xai2024realworldqa} & 55.1K \\
    \addlinespace[3pt]
    Documents / OCR & OCRBench v2~\citep{fu2025ocrbenchv2}, DocVQA~\citep{mathew2020docvqa}, TextVQA~\citep{singh2019textvqa}, MMLongBench-Doc~\citep{ma2024mmlongbenchdoc}, OmniDocBench~\citep{ouyang2024omnidocbench}, MTVQA~\citep{tang2024mtvqa} & 24.3K \\
    \addlinespace[3pt]
    Charts / diagrams / science and math & AI2D~\citep{kembhavi2016ai2d}, ChartQA~\citep{masry2022chartqa}, MathVista~\citep{lu2023mathvista}, OlympiadBench~\citep{he2024olympiadbench}, ScienceQA~\citep{lu2022scienceqa}, PhyX~\citep{shen2025phyx}, VisuLogic~\citep{xu2025visulogic}, MMMU~\citep{yue2023mmmu}, CMMMU~\citep{zhang2024cmmmu}, JMMMU~\citep{onohara2024jmmmu}, ZeroBench~\citep{roberts2025zerobench} & 21.0K \\
    \addlinespace[3pt]
    UI screenshots & ScreenSpot-Pro~\citep{li2025screenspotpro}, VisualWebBench~\citep{liu2024visualwebbench} & 3.6K \\
    \addlinespace[3pt]
    Faces & FFHQ~\citep{karras2018stylegan} & 70.0K \\
    \addlinespace[3pt]
    Generated / edited & GEdit-Bench~\citep{liu2025step1xedit}, KRIS-Bench~\citep{wu2025krisbench} & 3.7K \\
    \midrule
    Total & 35 datasets across six visual domains & $\sim$178K \\
    \bottomrule
  \end{tabularx}
  \endgroup
\end{table}

\subsection{Reconstruction Fidelity by Domain}

Table~\ref{tab:tokenizer_domain_fidelity} breaks down the reconstruction evaluation in Section~\ref{sec:tokenizer_evaluation}. \tokenizermodel{} improves PSNR, SSIM, LPIPS, and GMSD in every domain. The largest PSNR gains occur in charts and diagrams (5.35\,dB), UI screenshots (5.01\,dB), and documents (4.63\,dB). Natural photographs, faces, and generated or edited images also improve, with gains of 2.61--3.63\,dB.

\begin{table}[H]
  \centering
  \caption{Reconstruction fidelity by visual domain. Approximate domain sizes are shown in thousands of images. Bold marks the better value within each domain.}
  \label{tab:tokenizer_domain_fidelity}
  \begingroup
  \small
  \setlength{\tabcolsep}{4pt}
  \renewcommand{\arraystretch}{1.12}
  \begin{tabular*}{\linewidth}{@{\extracolsep{\fill}}l>{\raggedright\arraybackslash}p{3.7cm}rrrr@{}}
    \toprule
    Domain & Tokenizer & PSNR (dB) $\uparrow$ & SSIM $\uparrow$ & LPIPS $\downarrow$ & GMSD $\downarrow$ \\
    \midrule
    Natural photos & Emu3.5 IBQ & 23.65 & 0.678 & 0.109 & 0.123 \\
    (55.1K images) & \tokenizermodel{} (ours) & \textbf{26.26} & \textbf{0.738} & \textbf{0.099} & \textbf{0.093} \\
    \midrule
    Documents / OCR & Emu3.5 IBQ & 24.88 & 0.860 & 0.061 & 0.095 \\
    (24.3K images) & \tokenizermodel{} (ours) & \textbf{29.51} & \textbf{0.909} & \textbf{0.041} & \textbf{0.054} \\
    \midrule
    Charts / diagrams & Emu3.5 IBQ & 25.51 & 0.917 & 0.041 & 0.080 \\
    (21.0K images) & \tokenizermodel{} (ours) & \textbf{30.86} & \textbf{0.946} & \textbf{0.023} & \textbf{0.041} \\
    \midrule
    UI screenshots & Emu3.5 IBQ & 25.73 & 0.907 & 0.049 & 0.090 \\
    (3.6K images) & \tokenizermodel{} (ours) & \textbf{30.74} & \textbf{0.952} & \textbf{0.026} & \textbf{0.047} \\
    \midrule
    Faces & Emu3.5 IBQ & 28.56 & 0.787 & 0.091 & 0.088 \\
    (70.0K images) & \tokenizermodel{} (ours) & \textbf{31.70} & \textbf{0.844} & \textbf{0.077} & \textbf{0.061} \\
    \midrule
    Generated / edited & Emu3.5 IBQ & 26.86 & 0.792 & 0.079 & 0.098 \\
    (3.7K images) & \tokenizermodel{} (ours) & \textbf{30.49} & \textbf{0.843} & \textbf{0.067} & \textbf{0.068} \\
    \bottomrule
  \end{tabular*}
  \endgroup
\end{table}

\subsection{Distribution Distance, Image Quality, and Sharpness}

The remaining metrics give a more varied picture (Table~\ref{tab:tokenizer_domain_perceptual}). FD-DINOv2 improves in five of the six domains, with faces as the exception. FID improves for documents, charts, and faces, and is higher for natural photographs, UI screenshots, and generated or edited images. MUSIQ and CLIP-IQA also vary by domain. These scores complement the consistent gains in reference-based fidelity.

The overall Laplacian-variance ratio is 0.982 for \tokenizermodel{} and 0.892 for Emu3.5. Ratios above 1 in some structured-content domains indicate greater Laplacian variance in the reconstructions than in the originals; they do not by themselves establish better reconstruction.

\begin{table}[!htbp]
  \centering
  \caption{Feature-distribution distances, no-reference quality, and sharpness by domain. MUSIQ and CLIP-IQA evaluate the reconstructions. For the Laplacian-variance ratio, proximity to 1 determines the highlighted value; values above 1 are not inherently better.}
  \label{tab:tokenizer_domain_perceptual}
  \begingroup
  \footnotesize
  \setlength{\tabcolsep}{4pt}
  \renewcommand{\arraystretch}{1.12}
  \begin{tabular*}{\linewidth}{@{\extracolsep{\fill}}l>{\raggedright\arraybackslash}p{3.7cm}rrrrr@{}}
    \toprule
    Domain & Tokenizer & FID $\downarrow$ & FD-DINOv2 $\downarrow$ & MUSIQ $\uparrow$ & CLIP-IQA $\uparrow$ & Lap-var ratio $\to 1$ \\
    \midrule
    Natural photos & Emu3.5 IBQ & \textbf{26.35} & 15.83 & \textbf{65.57} & \textbf{0.555} & 0.854 \\
    (55.1K images) & \tokenizermodel{} (ours) & 26.48 & \textbf{12.29} & 65.28 & 0.545 & \textbf{0.920} \\
    \midrule
    Documents / OCR & Emu3.5 IBQ & 48.09 & 14.94 & 58.40 & 0.538 & 0.922 \\
    (24.3K images) & \tokenizermodel{} (ours) & \textbf{47.66} & \textbf{5.94} & \textbf{58.44} & \textbf{0.550} & \textbf{1.032} \\
    \midrule
    Charts / diagrams & Emu3.5 IBQ & 62.16 & 29.93 & 61.95 & \textbf{0.709} & 0.902 \\
    (21.0K images) & \tokenizermodel{} (ours) & \textbf{61.50} & \textbf{7.85} & \textbf{62.24} & \textbf{0.709} & \textbf{1.042} \\
    \midrule
    UI screenshots & Emu3.5 IBQ & \textbf{178.37} & 75.10 & \textbf{65.83} & \textbf{0.532} & 1.059 \\
    (3.6K images) & \tokenizermodel{} (ours) & 182.68 & \textbf{18.95} & 65.14 & 0.497 & \textbf{1.030} \\
    \midrule
    Faces & Emu3.5 IBQ & 164.74 & \textbf{6.34} & 69.64 & \textbf{0.610} & 0.950 \\
    (70.0K images) & \tokenizermodel{} (ours) & \textbf{160.53} & 7.73 & \textbf{69.66} & 0.591 & \textbf{0.957} \\
    \midrule
    Generated / edited & Emu3.5 IBQ & \textbf{48.60} & 29.81 & \textbf{68.79} & 0.582 & 0.915 \\
    (3.7K images) & \tokenizermodel{} (ours) & 49.03 & \textbf{22.01} & 68.76 & \textbf{0.583} & \textbf{0.940} \\
    \bottomrule
  \end{tabular*}
  \endgroup
\end{table}

\subsection{ImageNet Any-Resolution Reconstruction}
\label{app:tokenizer_resize}

Table~\ref{tab:tokenizer_resize_ablation} compares \tokenizermodel{} and Emu3.5 on ImageNet-1K validation~\citep{deng2009imagenet} using arbitrary-resolution inputs, separately from the six-domain evaluation. We report PSNR, SSIM, MS-SSIM~\citep{wang2003msssim}, LPIPS, and reconstruction FID (rFID). \tokenizermodel{} uses replicate padding to multiples of 16; rFID uses reference statistics from the Lanczos-preprocessed validation images. Native-scale encoding applies no internal resizing. The $2\times$ setting doubles both spatial dimensions before encoding and resizes the reconstruction back to the input size.

\begin{table}[H]
  \centering
  \caption{ImageNet-1K validation under arbitrary-resolution evaluation. Bold marks the better value between Emu3.5 and native-scale \tokenizermodel{}. The separate $2\times$ \tokenizermodel{} setting doubles each spatial dimension before encoding and resizes the reconstruction back, using approximately four times the native token count.}
  \label{tab:tokenizer_resize_ablation}
  \begingroup
  \small
  \setlength{\tabcolsep}{4pt}
  \renewcommand{\arraystretch}{1.12}
  \begin{tabular*}{\linewidth}{@{\extracolsep{\fill}}>{\raggedright\arraybackslash}p{4.4cm}rrrrr@{}}
    \toprule
    Tokenizer / encoding & PSNR (dB) $\uparrow$ & SSIM $\uparrow$ & MS-SSIM $\uparrow$ & LPIPS $\downarrow$ & rFID $\downarrow$ \\
    \midrule
    Emu3.5 IBQ & 22.13 & 0.5909 & 0.8387 & 0.1254 & \textbf{0.5482} \\
    \tokenizermodel{} (native) & \textbf{24.16} & \textbf{0.6529} & \textbf{0.8858} & \textbf{0.1209} & 0.8595 \\
    \midrule
    \tokenizermodel{} ($2\times$ scale) & 28.27 & 0.8076 & 0.9559 & 0.0610 & 0.4235 \\
    \bottomrule
  \end{tabular*}
  \endgroup
\end{table}

At native scale, \tokenizermodel{} improves PSNR from 22.13 to 24.16\,dB and also improves SSIM, MS-SSIM, and LPIPS relative to Emu3.5. Its rFID is higher (0.8595 versus 0.5482), so the paired-image fidelity gains do not extend to this distribution metric. Doubling \tokenizermodel{}'s encoding scale raises PSNR to 28.27\,dB and reduces rFID to 0.4235, at approximately four times the native visual-token count.

\FloatBarrier
\subsection{Additional Qualitative Comparisons}
\label{app:tokenizer_qualitative}

\begin{figure}[!htbp]
    \centering
    \begingroup\sffamily
    \pgfmathsetlengthmacro{\tokpanelwidth}{0.244\linewidth}
    \pgfmathsetlengthmacro{\tokcontextwidth}{0.205\linewidth}
    \pgfmathsetlengthmacro{\tokpixel}{\tokpanelwidth/256}

    \small
    \begin{tabular}{@{}c@{\hspace{0.02\linewidth}}c@{\hspace{0.02\linewidth}}c@{\hspace{0.02\linewidth}}c@{}}
        \textbf{Input image} & \textbf{Original crop} & \textbf{Emu3.5 IBQ} & \parbox[c]{\tokpanelwidth}{\centering\small\bfseries \tokenizermodel{} (ours)} \\[4pt]
        \begin{tikzpicture}[baseline=(current bounding box.center)]
            \node[anchor=south west, inner sep=0] (context) at (0,0)
                {\includegraphics[width=\tokcontextwidth,height=\tokpanelwidth,keepaspectratio]{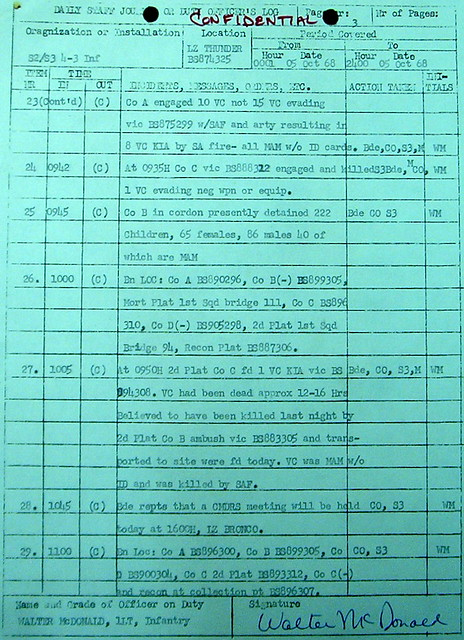}};
            \begin{scope}[x={(context.south east)},y={(context.north west)}]
                \draw[red!85!black,line width=0.7pt]
                    ({15/464},{1-90/640}) rectangle
                    ({(15+256)/464},{1-(90+256)/640});
            \end{scope}
        \end{tikzpicture} &
        \pgfmathsetlengthmacro{\tokimagewidth}{464*\tokpixel}%
        \begin{tikzpicture}[x=\tokpixel,y=\tokpixel,baseline=(current bounding box.center)]
            \path[use as bounding box] (0,0) rectangle (256,256);
            \begin{scope}[overlay]
                \clip (0,0) rectangle (256,256);
                \node[anchor=north west,inner sep=0] at (-15,{256+90})
                    {\includegraphics[width=\tokimagewidth]{figures/tokenizer_qualitative/original/20_s16_0012.jpg}};
            \end{scope}
        \end{tikzpicture} &
        \pgfmathsetlengthmacro{\tokimagewidth}{464*\tokpixel}%
        \begin{tikzpicture}[x=\tokpixel,y=\tokpixel,baseline=(current bounding box.center)]
            \path[use as bounding box] (0,0) rectangle (256,256);
            \begin{scope}[overlay]
                \clip (0,0) rectangle (256,256);
                \node[anchor=north west,inner sep=0] at (-15,{256+90})
                    {\includegraphics[width=\tokimagewidth]{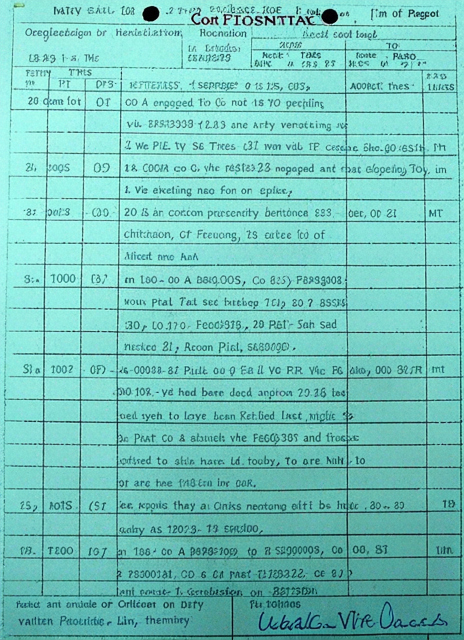}};
            \end{scope}
        \end{tikzpicture} &
        \pgfmathsetlengthmacro{\tokimagewidth}{464*\tokpixel}%
        \begin{tikzpicture}[x=\tokpixel,y=\tokpixel,baseline=(current bounding box.center)]
            \path[use as bounding box] (0,0) rectangle (256,256);
            \begin{scope}[overlay]
                \clip (0,0) rectangle (256,256);
                \node[anchor=north west,inner sep=0] at (-15,{256+90})
                    {\includegraphics[width=\tokimagewidth]{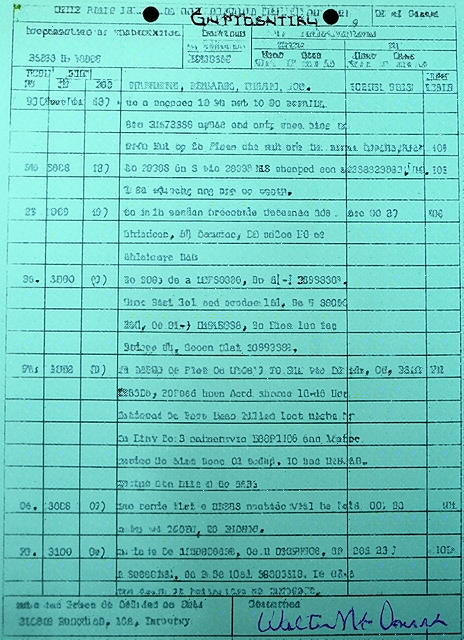}};
            \end{scope}
        \end{tikzpicture} \\[5pt]
        \begin{tikzpicture}[baseline=(current bounding box.center)]
            \node[anchor=south west, inner sep=0] (context) at (0,0)
                {\includegraphics[width=\tokcontextwidth,height=\tokpanelwidth,keepaspectratio]{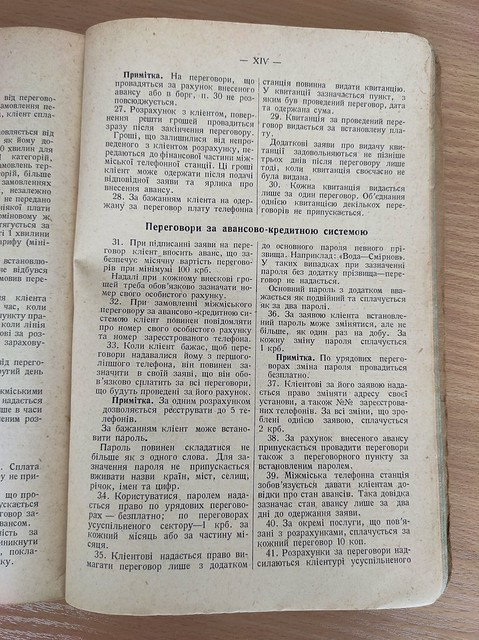}};
            \begin{scope}[x={(context.south east)},y={(context.north west)}]
                \draw[red!85!black,line width=0.7pt]
                    ({100/479},{1-70/640}) rectangle
                    ({(100+256)/479},{1-(70+256)/640});
            \end{scope}
        \end{tikzpicture} &
        \pgfmathsetlengthmacro{\tokimagewidth}{479*\tokpixel}%
        \begin{tikzpicture}[x=\tokpixel,y=\tokpixel,baseline=(current bounding box.center)]
            \path[use as bounding box] (0,0) rectangle (256,256);
            \begin{scope}[overlay]
                \clip (0,0) rectangle (256,256);
                \node[anchor=north west,inner sep=0] at (-100,{256+70})
                    {\includegraphics[width=\tokimagewidth]{figures/tokenizer_qualitative/original/23_s00_0023.jpg}};
            \end{scope}
        \end{tikzpicture} &
        \pgfmathsetlengthmacro{\tokimagewidth}{479*\tokpixel}%
        \begin{tikzpicture}[x=\tokpixel,y=\tokpixel,baseline=(current bounding box.center)]
            \path[use as bounding box] (0,0) rectangle (256,256);
            \begin{scope}[overlay]
                \clip (0,0) rectangle (256,256);
                \node[anchor=north west,inner sep=0] at (-100,{256+70})
                    {\includegraphics[width=\tokimagewidth]{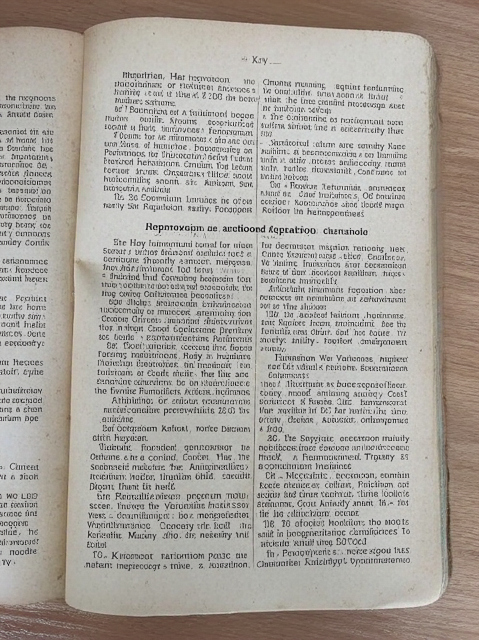}};
            \end{scope}
        \end{tikzpicture} &
        \pgfmathsetlengthmacro{\tokimagewidth}{479*\tokpixel}%
        \begin{tikzpicture}[x=\tokpixel,y=\tokpixel,baseline=(current bounding box.center)]
            \path[use as bounding box] (0,0) rectangle (256,256);
            \begin{scope}[overlay]
                \clip (0,0) rectangle (256,256);
                \node[anchor=north west,inner sep=0] at (-100,{256+70})
                    {\includegraphics[width=\tokimagewidth]{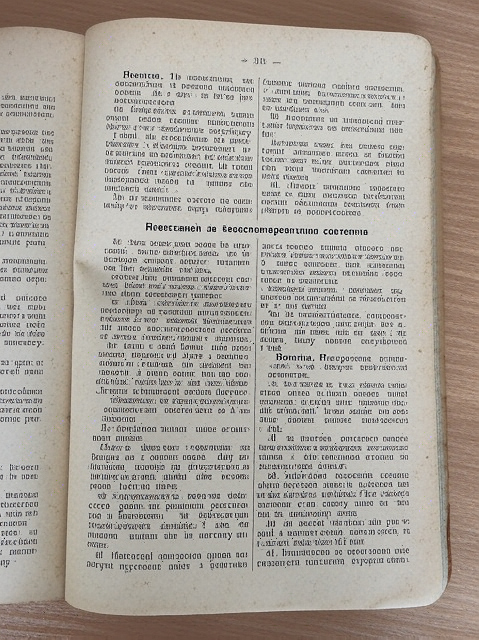}};
            \end{scope}
        \end{tikzpicture} \\[5pt]
        \begin{tikzpicture}[baseline=(current bounding box.center)]
            \node[anchor=south west, inner sep=0] (context) at (0,0)
                {\includegraphics[width=\tokcontextwidth,height=\tokpanelwidth,keepaspectratio]{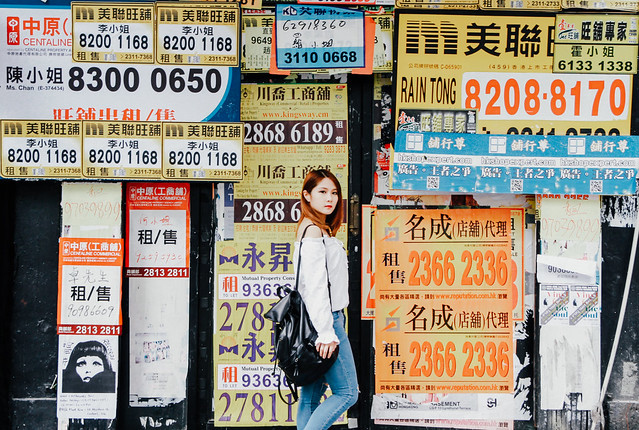}};
            \begin{scope}[x={(context.south east)},y={(context.north west)}]
                \draw[red!85!black,line width=0.7pt]
                    ({0/639},{1-0/430}) rectangle
                    ({(0+256)/639},{1-(0+256)/430});
            \end{scope}
        \end{tikzpicture} &
        \pgfmathsetlengthmacro{\tokimagewidth}{639*\tokpixel}%
        \begin{tikzpicture}[x=\tokpixel,y=\tokpixel,baseline=(current bounding box.center)]
            \path[use as bounding box] (0,0) rectangle (256,256);
            \begin{scope}[overlay]
                \clip (0,0) rectangle (256,256);
                \node[anchor=north west,inner sep=0] at (-0,{256+0})
                    {\includegraphics[width=\tokimagewidth]{figures/tokenizer_qualitative/original/19_s00_0063.jpg}};
            \end{scope}
        \end{tikzpicture} &
        \pgfmathsetlengthmacro{\tokimagewidth}{639*\tokpixel}%
        \begin{tikzpicture}[x=\tokpixel,y=\tokpixel,baseline=(current bounding box.center)]
            \path[use as bounding box] (0,0) rectangle (256,256);
            \begin{scope}[overlay]
                \clip (0,0) rectangle (256,256);
                \node[anchor=north west,inner sep=0] at (-0,{256+0})
                    {\includegraphics[width=\tokimagewidth]{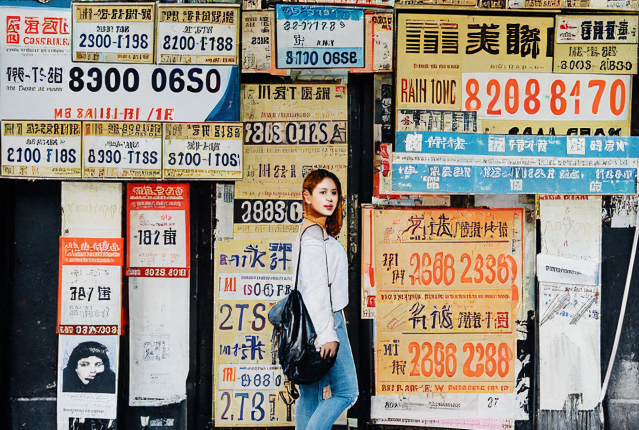}};
            \end{scope}
        \end{tikzpicture} &
        \pgfmathsetlengthmacro{\tokimagewidth}{639*\tokpixel}%
        \begin{tikzpicture}[x=\tokpixel,y=\tokpixel,baseline=(current bounding box.center)]
            \path[use as bounding box] (0,0) rectangle (256,256);
            \begin{scope}[overlay]
                \clip (0,0) rectangle (256,256);
                \node[anchor=north west,inner sep=0] at (-0,{256+0})
                    {\includegraphics[width=\tokimagewidth]{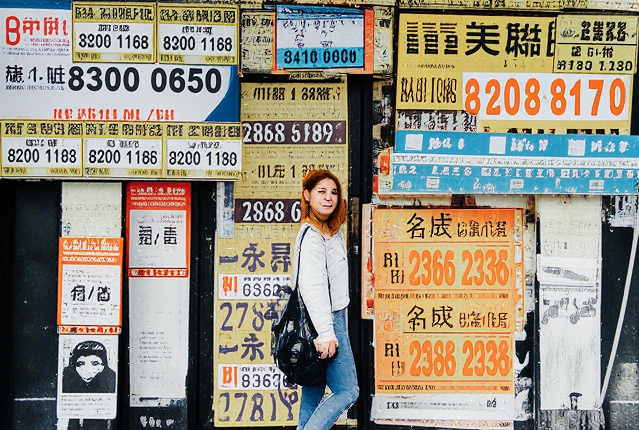}};
            \end{scope}
        \end{tikzpicture} \\[5pt]
        \begin{tikzpicture}[baseline=(current bounding box.center)]
            \node[anchor=south west, inner sep=0] (context) at (0,0)
                {\includegraphics[width=\tokcontextwidth,height=\tokpanelwidth,keepaspectratio]{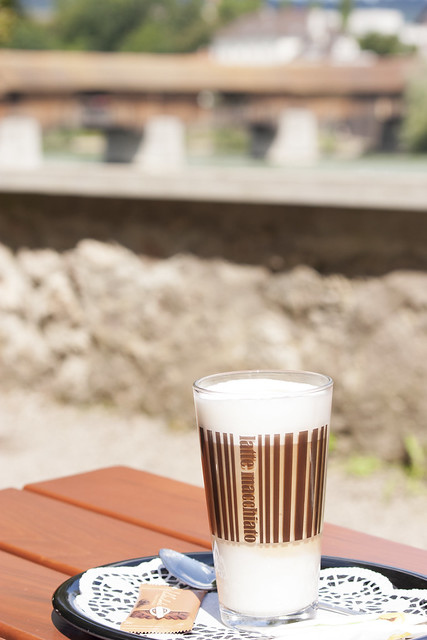}};
            \begin{scope}[x={(context.south east)},y={(context.north west)}]
                \draw[red!85!black,line width=0.7pt]
                    ({90/427},{1-380/640}) rectangle
                    ({(90+256)/427},{1-(380+256)/640});
            \end{scope}
        \end{tikzpicture} &
        \pgfmathsetlengthmacro{\tokimagewidth}{427*\tokpixel}%
        \begin{tikzpicture}[x=\tokpixel,y=\tokpixel,baseline=(current bounding box.center)]
            \path[use as bounding box] (0,0) rectangle (256,256);
            \begin{scope}[overlay]
                \clip (0,0) rectangle (256,256);
                \node[anchor=north west,inner sep=0] at (-90,{256+380})
                    {\includegraphics[width=\tokimagewidth]{figures/tokenizer_qualitative/original/06_s00_0038.jpg}};
            \end{scope}
        \end{tikzpicture} &
        \pgfmathsetlengthmacro{\tokimagewidth}{427*\tokpixel}%
        \begin{tikzpicture}[x=\tokpixel,y=\tokpixel,baseline=(current bounding box.center)]
            \path[use as bounding box] (0,0) rectangle (256,256);
            \begin{scope}[overlay]
                \clip (0,0) rectangle (256,256);
                \node[anchor=north west,inner sep=0] at (-90,{256+380})
                    {\includegraphics[width=\tokimagewidth]{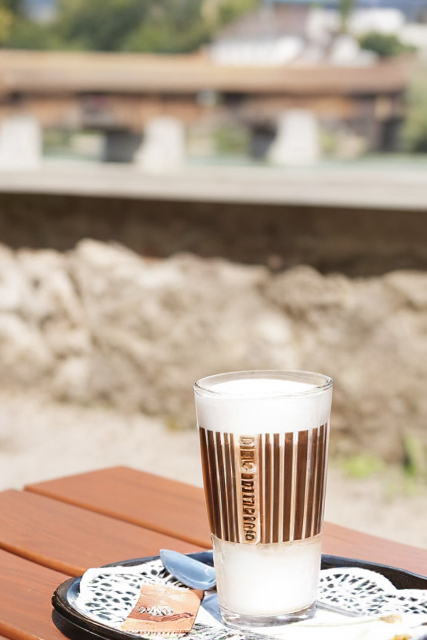}};
            \end{scope}
        \end{tikzpicture} &
        \pgfmathsetlengthmacro{\tokimagewidth}{427*\tokpixel}%
        \begin{tikzpicture}[x=\tokpixel,y=\tokpixel,baseline=(current bounding box.center)]
            \path[use as bounding box] (0,0) rectangle (256,256);
            \begin{scope}[overlay]
                \clip (0,0) rectangle (256,256);
                \node[anchor=north west,inner sep=0] at (-90,{256+380})
                    {\includegraphics[width=\tokimagewidth]{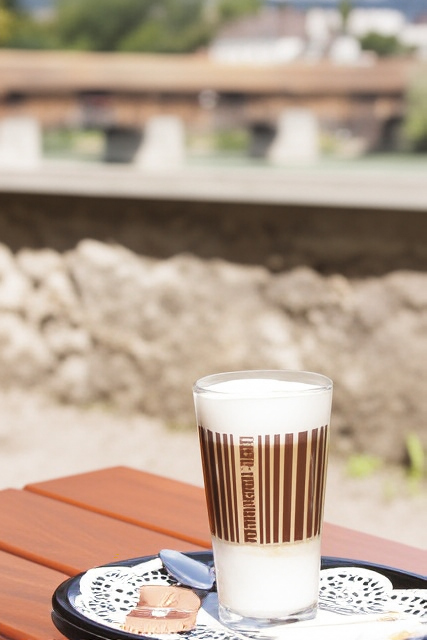}};
            \end{scope}
        \end{tikzpicture} \\
    \end{tabular}
    \endgroup
    \caption{Additional tokenizer reconstructions on selected GPIC images: a scanned form, a printed book, signs, and product packaging (top to bottom). Red boxes identify the same $256\times256$ region shown in each crop column. The comparisons show differences in layout, strokes, and repeated patterns; both tokenizers still alter small characters and digits.}
    \label{fig:tokenizer_qualitative_structured}
\end{figure}

\begin{figure}[!htbp]
    \centering
    \begingroup\sffamily
    \pgfmathsetlengthmacro{\tokpanelwidth}{0.244\linewidth}
    \pgfmathsetlengthmacro{\tokcontextwidth}{0.205\linewidth}
    \pgfmathsetlengthmacro{\tokpixel}{\tokpanelwidth/256}

    \small
    \begin{tabular}{@{}c@{\hspace{0.02\linewidth}}c@{\hspace{0.02\linewidth}}c@{\hspace{0.02\linewidth}}c@{}}
        \textbf{Input image} & \textbf{Original crop} & \textbf{Emu3.5 IBQ} & \parbox[c]{\tokpanelwidth}{\centering\small\bfseries \tokenizermodel{} (ours)} \\[4pt]
        \begin{tikzpicture}[baseline=(current bounding box.center)]
            \node[anchor=south west, inner sep=0] (context) at (0,0)
                {\includegraphics[width=\tokcontextwidth,height=\tokpanelwidth,keepaspectratio]{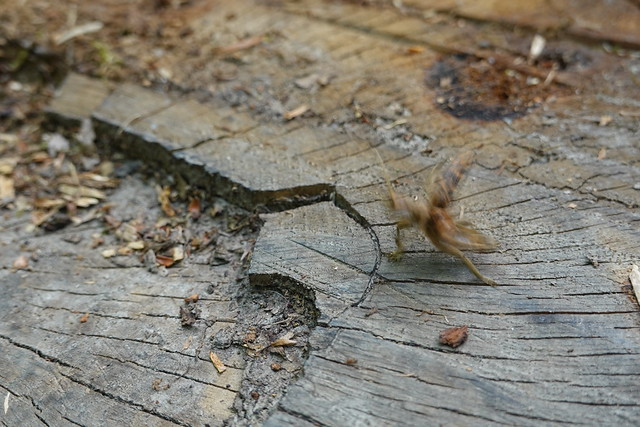}};
            \begin{scope}[x={(context.south east)},y={(context.north west)}]
                \draw[red!85!black,line width=0.7pt]
                    ({200/640},{1-160/427}) rectangle
                    ({(200+256)/640},{1-(160+256)/427});
            \end{scope}
        \end{tikzpicture} &
        \pgfmathsetlengthmacro{\tokimagewidth}{640*\tokpixel}%
        \begin{tikzpicture}[x=\tokpixel,y=\tokpixel,baseline=(current bounding box.center)]
            \path[use as bounding box] (0,0) rectangle (256,256);
            \begin{scope}[overlay]
                \clip (0,0) rectangle (256,256);
                \node[anchor=north west,inner sep=0] at (-200,{256+160})
                    {\includegraphics[width=\tokimagewidth]{figures/tokenizer_qualitative/original/04_s00_0001.jpg}};
            \end{scope}
        \end{tikzpicture} &
        \pgfmathsetlengthmacro{\tokimagewidth}{640*\tokpixel}%
        \begin{tikzpicture}[x=\tokpixel,y=\tokpixel,baseline=(current bounding box.center)]
            \path[use as bounding box] (0,0) rectangle (256,256);
            \begin{scope}[overlay]
                \clip (0,0) rectangle (256,256);
                \node[anchor=north west,inner sep=0] at (-200,{256+160})
                    {\includegraphics[width=\tokimagewidth]{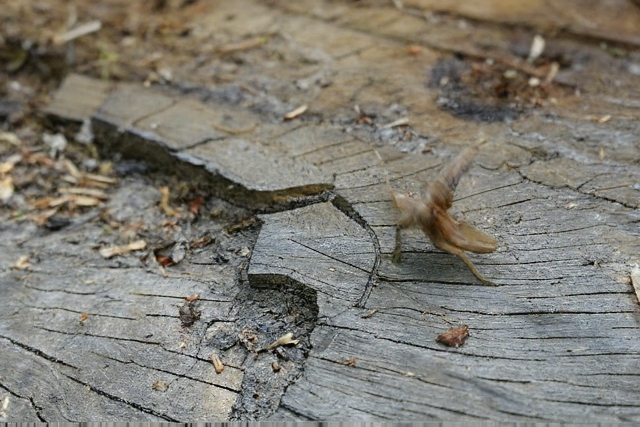}};
            \end{scope}
        \end{tikzpicture} &
        \pgfmathsetlengthmacro{\tokimagewidth}{640*\tokpixel}%
        \begin{tikzpicture}[x=\tokpixel,y=\tokpixel,baseline=(current bounding box.center)]
            \path[use as bounding box] (0,0) rectangle (256,256);
            \begin{scope}[overlay]
                \clip (0,0) rectangle (256,256);
                \node[anchor=north west,inner sep=0] at (-200,{256+160})
                    {\includegraphics[width=\tokimagewidth]{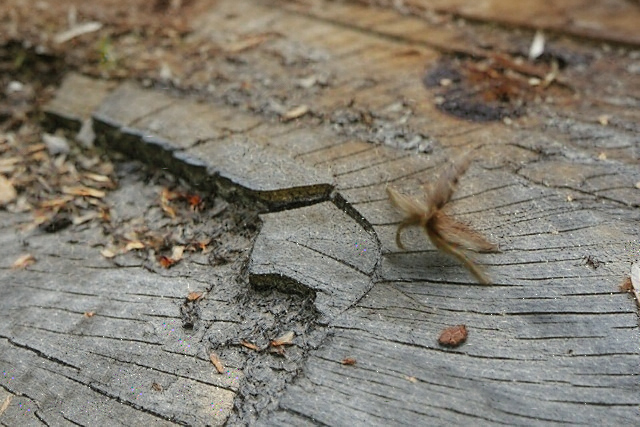}};
            \end{scope}
        \end{tikzpicture} \\[5pt]
        \begin{tikzpicture}[baseline=(current bounding box.center)]
            \node[anchor=south west, inner sep=0] (context) at (0,0)
                {\includegraphics[width=\tokcontextwidth,height=\tokpanelwidth,keepaspectratio]{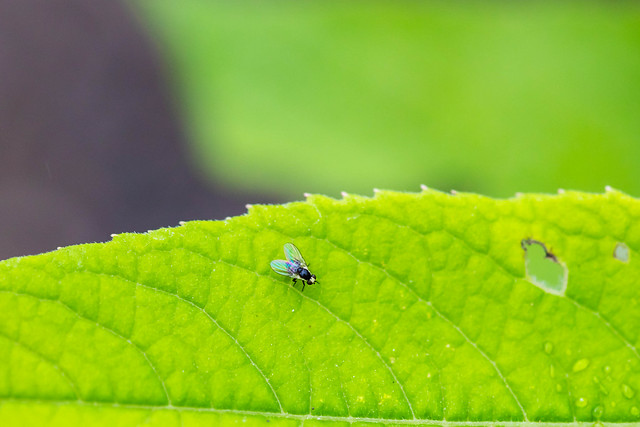}};
            \begin{scope}[x={(context.south east)},y={(context.north west)}]
                \draw[red!85!black,line width=0.7pt]
                    ({370/640},{1-165/427}) rectangle
                    ({(370+256)/640},{1-(165+256)/427});
            \end{scope}
        \end{tikzpicture} &
        \pgfmathsetlengthmacro{\tokimagewidth}{640*\tokpixel}%
        \begin{tikzpicture}[x=\tokpixel,y=\tokpixel,baseline=(current bounding box.center)]
            \path[use as bounding box] (0,0) rectangle (256,256);
            \begin{scope}[overlay]
                \clip (0,0) rectangle (256,256);
                \node[anchor=north west,inner sep=0] at (-370,{256+165})
                    {\includegraphics[width=\tokimagewidth]{figures/tokenizer_qualitative/original/09_s08_0014.jpg}};
            \end{scope}
        \end{tikzpicture} &
        \pgfmathsetlengthmacro{\tokimagewidth}{640*\tokpixel}%
        \begin{tikzpicture}[x=\tokpixel,y=\tokpixel,baseline=(current bounding box.center)]
            \path[use as bounding box] (0,0) rectangle (256,256);
            \begin{scope}[overlay]
                \clip (0,0) rectangle (256,256);
                \node[anchor=north west,inner sep=0] at (-370,{256+165})
                    {\includegraphics[width=\tokimagewidth]{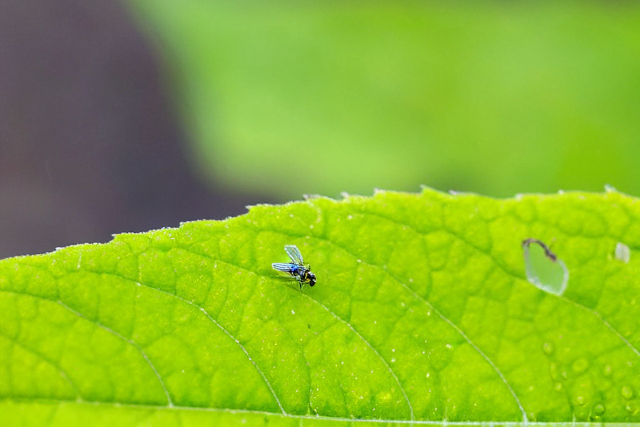}};
            \end{scope}
        \end{tikzpicture} &
        \pgfmathsetlengthmacro{\tokimagewidth}{640*\tokpixel}%
        \begin{tikzpicture}[x=\tokpixel,y=\tokpixel,baseline=(current bounding box.center)]
            \path[use as bounding box] (0,0) rectangle (256,256);
            \begin{scope}[overlay]
                \clip (0,0) rectangle (256,256);
                \node[anchor=north west,inner sep=0] at (-370,{256+165})
                    {\includegraphics[width=\tokimagewidth]{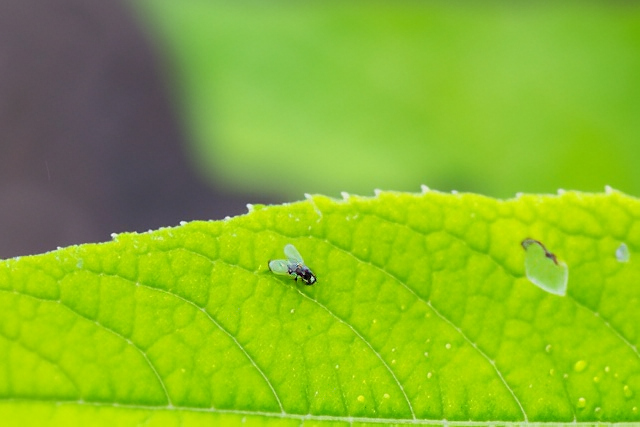}};
            \end{scope}
        \end{tikzpicture} \\[5pt]
        \begin{tikzpicture}[baseline=(current bounding box.center)]
            \node[anchor=south west, inner sep=0] (context) at (0,0)
                {\includegraphics[width=\tokcontextwidth,height=\tokpanelwidth,keepaspectratio]{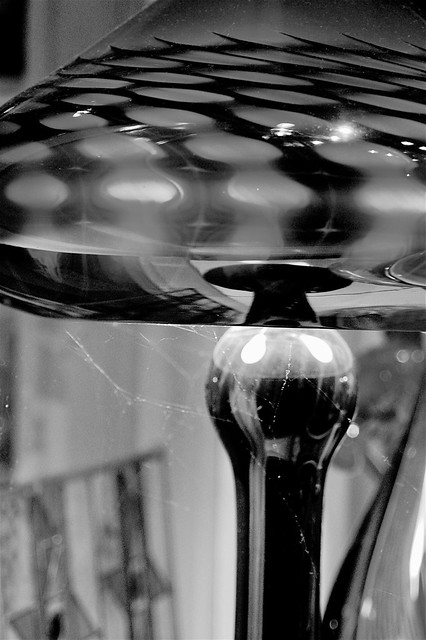}};
            \begin{scope}[x={(context.south east)},y={(context.north west)}]
                \draw[red!85!black,line width=0.7pt]
                    ({80/426},{1-320/640}) rectangle
                    ({(80+256)/426},{1-(320+256)/640});
            \end{scope}
        \end{tikzpicture} &
        \pgfmathsetlengthmacro{\tokimagewidth}{426*\tokpixel}%
        \begin{tikzpicture}[x=\tokpixel,y=\tokpixel,baseline=(current bounding box.center)]
            \path[use as bounding box] (0,0) rectangle (256,256);
            \begin{scope}[overlay]
                \clip (0,0) rectangle (256,256);
                \node[anchor=north west,inner sep=0] at (-80,{256+320})
                    {\includegraphics[width=\tokimagewidth]{figures/tokenizer_qualitative/original/08_s00_0053.jpg}};
            \end{scope}
        \end{tikzpicture} &
        \pgfmathsetlengthmacro{\tokimagewidth}{426*\tokpixel}%
        \begin{tikzpicture}[x=\tokpixel,y=\tokpixel,baseline=(current bounding box.center)]
            \path[use as bounding box] (0,0) rectangle (256,256);
            \begin{scope}[overlay]
                \clip (0,0) rectangle (256,256);
                \node[anchor=north west,inner sep=0] at (-80,{256+320})
                    {\includegraphics[width=\tokimagewidth]{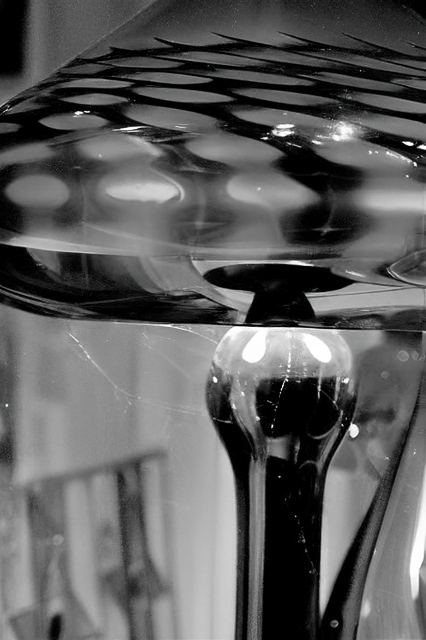}};
            \end{scope}
        \end{tikzpicture} &
        \pgfmathsetlengthmacro{\tokimagewidth}{426*\tokpixel}%
        \begin{tikzpicture}[x=\tokpixel,y=\tokpixel,baseline=(current bounding box.center)]
            \path[use as bounding box] (0,0) rectangle (256,256);
            \begin{scope}[overlay]
                \clip (0,0) rectangle (256,256);
                \node[anchor=north west,inner sep=0] at (-80,{256+320})
                    {\includegraphics[width=\tokimagewidth]{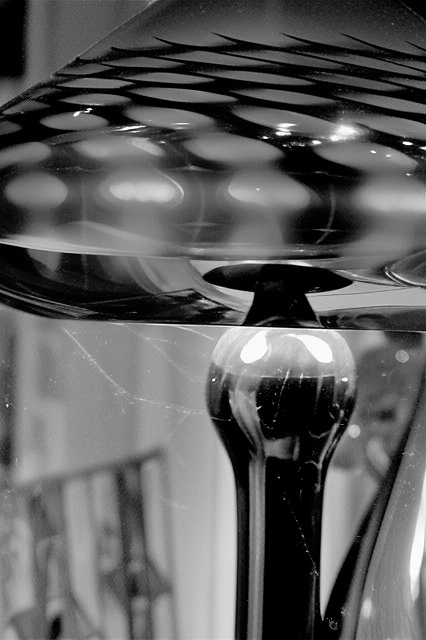}};
            \end{scope}
        \end{tikzpicture} \\[5pt]
        \begin{tikzpicture}[baseline=(current bounding box.center)]
            \node[anchor=south west, inner sep=0] (context) at (0,0)
                {\includegraphics[width=\tokcontextwidth,height=\tokpanelwidth,keepaspectratio]{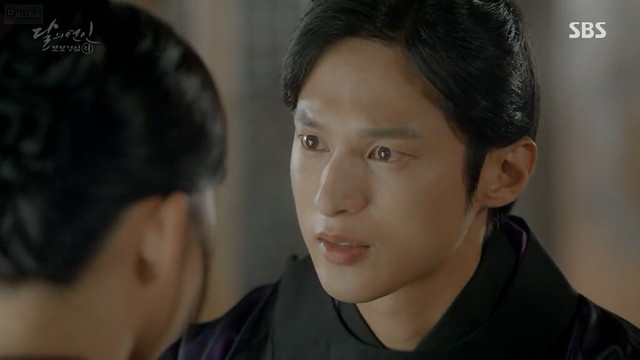}};
            \begin{scope}[x={(context.south east)},y={(context.north west)}]
                \draw[red!85!black,line width=0.7pt]
                    ({275/640},{1-35/360}) rectangle
                    ({(275+256)/640},{1-(35+256)/360});
            \end{scope}
        \end{tikzpicture} &
        \pgfmathsetlengthmacro{\tokimagewidth}{640*\tokpixel}%
        \begin{tikzpicture}[x=\tokpixel,y=\tokpixel,baseline=(current bounding box.center)]
            \path[use as bounding box] (0,0) rectangle (256,256);
            \begin{scope}[overlay]
                \clip (0,0) rectangle (256,256);
                \node[anchor=north west,inner sep=0] at (-275,{256+35})
                    {\includegraphics[width=\tokimagewidth]{figures/tokenizer_qualitative/original/24_s24_0006.jpg}};
            \end{scope}
        \end{tikzpicture} &
        \pgfmathsetlengthmacro{\tokimagewidth}{640*\tokpixel}%
        \begin{tikzpicture}[x=\tokpixel,y=\tokpixel,baseline=(current bounding box.center)]
            \path[use as bounding box] (0,0) rectangle (256,256);
            \begin{scope}[overlay]
                \clip (0,0) rectangle (256,256);
                \node[anchor=north west,inner sep=0] at (-275,{256+35})
                    {\includegraphics[width=\tokimagewidth]{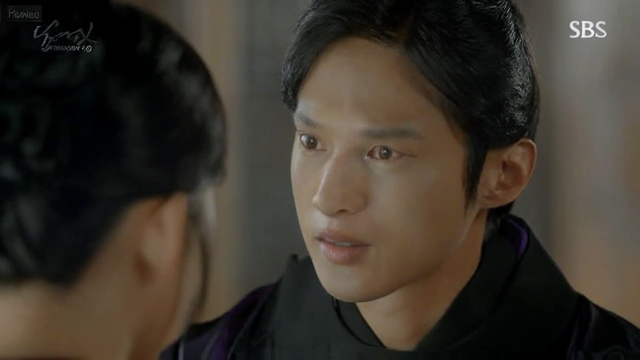}};
            \end{scope}
        \end{tikzpicture} &
        \pgfmathsetlengthmacro{\tokimagewidth}{640*\tokpixel}%
        \begin{tikzpicture}[x=\tokpixel,y=\tokpixel,baseline=(current bounding box.center)]
            \path[use as bounding box] (0,0) rectangle (256,256);
            \begin{scope}[overlay]
                \clip (0,0) rectangle (256,256);
                \node[anchor=north west,inner sep=0] at (-275,{256+35})
                    {\includegraphics[width=\tokimagewidth]{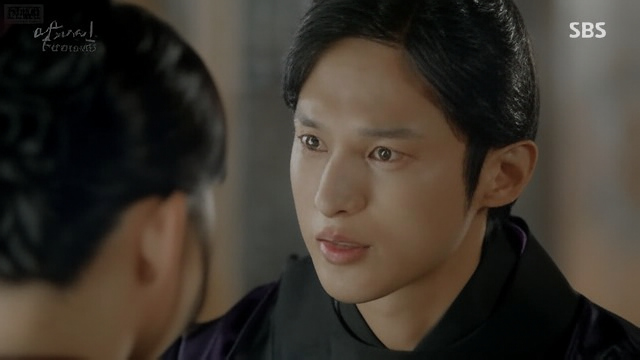}};
            \end{scope}
        \end{tikzpicture} \\
    \end{tabular}
    \endgroup
    \caption{Additional tokenizer reconstructions on selected GPIC images: wood texture, leaf veins, reflective glass, and a portrait (top to bottom). Each row compares identical $256\times256$ regions at the same scale. These examples complement the structured-content comparisons with natural texture, material appearance, and facial detail.}
    \label{fig:tokenizer_qualitative_natural}
\end{figure}

Figures~\ref{fig:tokenizer_qualitative_structured} and~\ref{fig:tokenizer_qualitative_natural} add eight selected GPIC reconstructions~\citep{chandrasegaran2026gpic}, using the checkpoint, baseline, and matched-crop protocol of Figure~\ref{fig:tokenizer_qualitative}. The examples broaden coverage to documents, signs, product packaging, natural textures, reflective materials, and portraits. Both tokenizers still alter small characters and fine structures.

\FloatBarrier

\end{document}